\documentclass[twoside,11pt]{article}

\usepackage{appendix}

\usepackage{jmlr2e}
\usepackage{lastpage}

\usepackage{mathtools}
\usepackage{bbm}
\usepackage{booktabs}
\usepackage{subcaption}
\usepackage{xparse}
\usepackage{sidecap}
\usepackage{float}
\usepackage{enumerate,mdwlist}
\usepackage{epsfig}
\usepackage[bbgreekl]{mathbbol}
\usepackage{amsfonts,amsmath,enumerate}
\usepackage{amscd}
\usepackage{amsmath,amsfonts,mathrsfs}\usepackage{bm}
\usepackage{latexsym}
\usepackage{mleftright}
\usepackage{color}
\usepackage{xifthen}
\usepackage{makeidx}
\usepackage{amsmath}
\usepackage{amsfonts}
\usepackage{amssymb}
\usepackage{mathrsfs}
\usepackage{tikz,tikz-cd}
\usepackage{tkz-euclide}
\usetikzlibrary{matrix}
\usepackage{smartdiagram}

\usepackage[linesnumbered,ruled,vlined]{algorithm2e}
\SetKwComment{Comment}{$\triangleright$\ }{}
\usepackage{etoolbox}
\usepackage[export]{adjustbox}
\renewcommand*{\algorithmcfname}{Meta-Algorithm}
\newcounter{algoline}

\AtBeginEnvironment{algorithm}{\setcounter{algoline}{0}}

\usepackage{hyperref} 
\hypersetup{
	colorlinks = true,
	linkcolor = blue,
	anchorcolor = blue,
	citecolor = blue,
	filecolor = blue,
	urlcolor = blue
}

\newcommand{\rr}{{\mathbb{R}}}
\newcommand{\pp}{{\mathbb{P}}}
\newcommand{\qq}{{\mathbb{Q}}}
\newcommand{\rrflex}[1]{{\ensuremath{\rr^{#1}
}}}
\newcommand{\rrD}{{\rrflex{D}}}
\newcommand{\rrd}{{\rrflex{d}}}

\newcommand{\rrF}{{\rrflex{F}}}

\newcommand{\nn}{{\mathbb{N}}}

\newcommand{\mmm}{{\mathscr{M}}}

\newcommand{\hhh}{{\mathscr{H}}}

\newcommand{\fff}{{\mathscr{F}}}	
	
\newcommand{\Skw}{{\operatorname{Skw}}}

\NewDocumentCommand\Gen{oo}{\operatorname{Gen}_M\left(\IfValueF{#1}{\hat{f}}\IfValueT{#1}{{#1}}\middle|\IfValueF{#1}{\{w_n\}_{n\leq N}}\IfValueT{#2}{{#2}}\right)}
\NewDocumentCommand\Dkl{oo}{{\operatorname{D}_{KL}\IfValueT{#1}{\left(\IfValueF{#2}{\cdot}\IfValueT{#2}{{#2}}\middle\|{#1}\right)}}}

\NewDocumentCommand{\NN}{oo}{{\mathcal{NN}^{\IfValueF{#2}{\sigma}\IfValueT{#2}{{#2}}}_{\IfValueT{#1}{{#1}}\IfValueF{#1}{d,d:J,K}}}}
\NewDocumentCommand{\RN}{oo}{{\Phi^{\IfValueT{#2}{{#2}}}_{\IfValueT{#1}{{#1}}\IfValueF{#1}{\star:d}}}}

\newcommand{\so}[1]{\ensuremath{{\mathfrak{so}_{{#1}}}}}

\newtheorem{defn}{Definition}[section]

\newtheorem{ass}[defn]{Assumption}

\usepackage{marginnote}
\definecolor{WowColor}{rgb}{.75,0,.75}
\definecolor{MildlyAlarming}{rgb}{0.85,0.25,0.1}
\definecolor{SubtleColor}{rgb}{0,0,.50}

\newcounter{margincounter}

\ShortHeadings{NEU}{Kratsios and Hyndman}

\jmlrheading{22}{2021}{1-\pageref{LastPage}}{11/18; Revised 2/21}{5/21}{18-803}{Anastasis Kratsios, Coy Hyndman}

\begin{document}
	\title{NEU: A Meta-Algorithm for Universal UAP-Invariant Feature Representation}

	\author{\name Anastasis Kratsios \email anastasis.kratsios@math.ethz.ch \\
		\addr Department of Mathematics\\
		Eidgen\"{o}ssische Technische Hochschule Z\"{u}rich\\ 
		R\"{a}mistrasse 101, 8092 Z\"{u}rich, ZH, Switzerland
		\AND
		\name Cody Hyndman \email cody.hyndman@concordia.ca \\
		\addr Department of Mathematics and Statistics\\
		Concordia University\\
		1455 boulevard de Maisonneuve Ouest, Montr\'{e}al, Qu\'{e}bec, H3G 1M8, Canada}
	
	 \editor{George Konidaris}

	\maketitle

	\begin{abstract}
		Effective feature representation is key to the predictive performance of any algorithm.  This paper introduces a meta-procedure, called Non-Euclidean Upgrading (NEU), which learns feature maps that are expressive enough to embed the universal approximation property (UAP) into most model classes while only outputting feature maps that preserve any model class's UAP.  We show that NEU can learn any feature map with these two properties if that feature map is asymptotically deformable into the identity.   We also find that the feature-representations learned by NEU are always submanifolds of the feature space.   NEU's properties are derived from a new deep neural model that is universal amongst all orientation-preserving homeomorphisms on the input space.  We derive qualitative and quantitative approximation guarantees for this architecture.  We quantify the number of parameters required for this new architecture to memorize any set of input-output pairs while simultaneously fixing every point of the input space lying outside some compact set, and we quantify the size of this set as a function of our model's depth.  Moreover, we show that deep feed-forward networks with most commonly used activation functions typically do not have all these properties.  NEU's performance is evaluated against competing machine learning methods on various regression and dimension reduction tasks both with financial and simulated data. 
	\end{abstract}

	\hfill\\

	\section{Introduction}
	The training phase of most learning problems seeks to identify a model $\hat{f}$ belonging to a model class $\fff$, which best approximates an unknown function $f$, as given by:
	\begin{equation}
	\min_{\hat{f} \in \fff} \,\sum_{n\leq N} \frac1{N} \,
	L(f(x_n),\hat{f}(x_n),x_n) + P(\hat{f})
	\label{eq_naive}
	\tag{L}
	,
	\end{equation}
	where $\{x_n\}_{n\leq N}$ is a given set of training data, $L$ is a loss-function, and $P$ is a penalty which encodes regularity into the model $\hat{f}$.  The effectiveness of the learning task~\eqref{eq_naive} often hinges on the appropriateness of the input data's representation.  Where by a \textit{representation} of the input space $\rrd$ is the subset $\phi(\rrd)\subseteq \rrF$ and where $\phi:\rrd\rightarrow \rrF$ is a \textit{feature map} mapping into the \textit{feature space} $\rrF$.  Following \cite{MicchelliUniversalFeature}, we define feature maps as being continuous, and by \cite{BrowerDomainInvarianceTheorem1911} we observe that the feature space's dimension must be at-least that of $\rrd$.  
	
	Two popular but differing approaches to representation learning are offered by kernel methods and by deep learning.  
	Introduced by \cite{boser1992training}, the former of the two implicitly embeds $\rrd$ into a high, and often infinite, dimensional linear space $H$ by using the correspondence between feature maps and kernels identified in \cite{AronszajnRKS,MichelliWhenRepresentor}.  However, the effectiveness of these methods hinges on the appropriateness of the specified kernel; see \cite{Misspecification2020ConvergenceAnalsysiskernelquadraturerules} for example.  
	In contrast, the deep learning paradigm offers a non-parametric approach to representation learning.  This is because any deep feed-forward network (DNN) $\hat{f}$ is necessarily of the form $\hat{f}(x)=W\circ \phi(x)$ where $W$ is an affine function on $\rrF$ and $\phi:\rrd\rightarrow \rrF$ is a feature map generated by iteratively applying feed-forward layers to the input space.

	This paper introduces Non-Euclidean Upgrading (NEU), a meta-algorithm that incorporates a linearizing preprocessing step into~\eqref{eq_naive} as summarized in Meta-Algorithm~\ref{algo_NEU_redux}.  During this preprocessing step, NEU generates a feature map that both increases the expressiveness of $\fff$ and preserves its approximation capabilities by learning a topological embedding of the input space $\rrd$ into a low-dimensional feature space $\rrF$.  NEU does this by training a new deep neural model type, called the \textit{reconfiguration network} and denoted by $\Phi_{\star:d}$, whose members form a universal class of regular feature maps.  
	
	NEU balances the newly found flexibility, which $\Phi_{\star:d}$ embeds into $\fff$ by optimally re-weighting the relative impact of each training data-point $\{x_n\}_{n\leq N}$ in~\eqref{eq_naive} so as to minimize the gap between the trained model's training and testing performance.  We denote these new, data-dependent, weights by $\{w_n^{\star,\lambda}\}_{n\leq N}$, where $\lambda>0$ is a hyper-parameter.  
	
	\begin{algorithm}[H]
		\SetAlgoLined
		\SetKwInOut{Input}{input}\SetKwInOut{Output}{output}
		\Input{Hypothesis class $\fff$, loss-function $L$, penalty function $P$, \\
			Training Data $\{x_n\}_{n\leq N}$\\
			Feature map's depth $J$\\
			Robustness Hyper-parameter $\lambda>0$\\
		}
		\Output{NEU-model $f^{NEU}\triangleq \hat{f}\circ \hat{\phi}_{\hat{I}}.$}
		\BlankLine
		$
		\hat{\phi} \in \underset{\phi \in \RN }{\operatorname{argmin}}
		\sum_{n\leq N} w_n^{\star,\lambda}\,
		L\left(f(x_n),A\phi(x_n)+b,x_n\right) + P(A\phi+b)
		$
		\Comment*[l]{Get Feature Map}	
		$
		\hat{f} \in 
		\underset{\hat{f} \in \mathcal{F}}{\operatorname{argmin}}
		\sum_{n\leq N} w_n^{\star,\lambda}\,
		L\left(f(x_n),\hat{f}\circ \hat{\phi}(x_n)+b,x_n\right) + P(\hat{f}\circ \hat{\phi})
		$
		\Comment*[l]{Get NEU-Model}
		\caption{Non-Euclidean Upgrading (NEU)}
		\label{algo_NEU_redux}
	\end{algorithm}

	We motivate NEU through its properties.  Many feature maps can impede the universal approximation property (UAP) of $\fff$.  Naturally, we require that any feature map generated by NEU, satisfies the following \textit{UAP-invariance property}, which is introduced and characterized in \cite{kratsios2020non}:\footnote{The authors find that (P-i) holds exactly when $\phi$ is injective.  Consequentially, \cite{BrowerDomainInvarianceTheorem1911} implies any UAP-invariant feature map must map into a feature space $\rrF$ of dimension at-least $d$.}
	\begin{enumerate}
		\item[  (P-i)] If $\fff$ is a universal in $C(\rrd,\rrD)$ then so is $\fff\circ \phi$.
	\end{enumerate}
	
	Next, we require NEU should be able to learn the identity continuously.  Thus, it should be capable of not imposing any additional unnecessary structure if, and once, the input space is sufficiently well-represented.   Mathematically, we require that the collection of feature maps that NEU can generate, denoted for the moment by $\Phi$, satisfy:
	\begin{enumerate}
		\item[  (P-ii)] Any $\phi \in \Phi$ can be parameterized to continuously learn the identity; i.e., there is a continuous map $\phi_{\alpha}:[0,1]\times \rrd \rightarrow \rrF$ such that
		$$
		\phi_{1}(x)= \phi(x)
		,\qquad
		\phi_0((x_1,\dots,x_n))=(x_1,\dots,x_d,0,\dots,0)
		,
		\qquad
		\phi_{\alpha} \mbox{ satisfies (P-i)}
		.
		$$
	\end{enumerate}
	Property (P-ii) is critical when members of $\Phi$ are built by repeatedly composing many layers since failing (P-i) forces all deeper layers to simultaneously learn the target function and compensate the mistakes of erroneously applied earlier layers.  As discussed in, \cite{hardt2016identity}, property (P-ii) is core to the success of the batch normalization algorithm of \cite{ioffe2015batch}, among other recent deep learning paradigms.  
	
	NEU is designed to exclusively generate feature maps satisfying both (P-i) and (P-ii).  Our main universal approximation results will show that the feature maps generated by NEU are universal amongst all those satisfying both properties (P-i) and (P-ii).  In contrast, we show that typically DNNs with the ReLU non-linearity of \cite{hahnloser2000digital} fail to satisfy (P-i).  
	
	Together, properties (P-i) and (P-ii) only guarantee that a class of feature maps $\Phi$ does not disrupt the representation of any input space.  However, we are most interested in identifying a feature map class $\Phi$, which additionally improves the expressiveness of any model class possessing a basic level of expressiveness.  By this, we mean that NEU should imbue most learning models with the universal approximation property:
	\begin{enumerate}
		\item[(P-iii)] If $\fff$ contains all linear maps, then $\{\fff \circ \phi\}_{\phi \in \Phi}$ is universal.
	\end{enumerate}
	Property (P-iii) is "asymptotic" since it guarantees that any function can eventually be approximated if a sufficiently complex feature map is used.  We complement it with the following, non-asymptotic, refined \textit{memorization} property:
	\begin{enumerate}
		\item[(P-iv)] If $n=d$ and $\fff$ contains all linear maps, then given any input-output pairs $\{x_n\}_{n\leq N}$ and $\{y_n\}_{n\leq N}$ and any tolerance $\delta>0$, some feature map $\phi \in \Phi$ satisfies
		$$
		\phi(x_n)=y_n \mbox{ and } \mu\left(\{x \in \rrd: \phi(x)\neq x\}\right)<\delta;
		$$
		for every $n\leq N$; where $\mu$ is the Lebesgue measure on $\rrd$.  
	\end{enumerate}
	Property (P-iv) is a refinement of the arbitrary \textit{memory capacity} of feed-forward networks studied in \cite{MemoryCapacilty2009}, which simultaneously asks that NEU be able to leave most of the unseen data unimpacted.  We show that NEU generates feature maps satisfying properties (P-iii) and (P-iv) and that DNNs with commonly-used non-ReLU activation functions, such as the Swish non-linearity of \cite{ramachandran2017searching}, the Gaussian Error Linear Unit of \cite{hendrycks2016gaussian}, the Soft-Plus activation of \cite{glorot2011deep}, and $\tanh$ activation functions all fail (P-iv).  Analogously to \cite{pmlrv75yarotsky18a}, \cite{GittaPetersenHelmutGrohs2019Optimal}, \cite{lu2020deep}, and \cite{kratsios2021quantitative} our approximation guarantees are \textit{quantitative} and analogously to \cite{MemoryCapacilty2009}, \cite{yun2019small}, and \cite{MemoryCapacityDNNs} our memorization guarantees are also quantitative.

	\subsection*{Outline of the Paper}
	Section~\ref{s_Prelim} begins by covering the topological background required for the framing of our main results and ends with the precise description of the deep neural model which NEU trains.  Section~\ref{s_Main_results} contains the paper's main theoretical contributions.  These include various universal approximation results, guarantees on the reconfiguration network's memory capacity, and guarantees that the reconfiguration network can approximate at the desired optimal rates.  The implications of these results are then unpacked in the context of NEU.  Section~\ref{s_Implementations} evaluates the predictive gain obtained by applying NEU across various regression and dimension reduction problems.  Our implementations focus on financial data analysis.  The performance of NEU regression methods is subsequently evaluated on simulated data to understand its implications in a fully controlled environment.  Specifically, NEU is stress-tested using various pathological regression challenges.  
	All proofs and any additional topological background is available in the supplementary material.  
	
	\section*{Notation}\label{s_notation}
	The following notation is maintained throughout this paper.  We denote the set of continuous functions from $\rrd$ to $\rrD$ by $C(\rrd,\rrD)$.  The set of DNNs from $\rrd$ to $\rrD$ with activation function $\sigma\in C(\rr)$ and at-least one hidden layer is denoted by $\NN[d,D]$.  
	
	\section{Preliminaries}\label{s_Prelim}
	This section covers the background and definition required in the remainder of this paper.  
	\subsection{Background}\label{ss_Background}
	\subsubsection{Continuous Functions}\label{sss_Background_CNT_Functions}
	We denote the Euclidean norm by $\|\cdot\|$ on $\rrd$ (resp $\rrD$).  Following \cite{hornik1989multilayer}, we view $C(\rrd,\rrD)$ as a metric space, with  metric $d_{ucc}$ defined for $f,g \in C(\rrd,\rrD)$ by
	\begin{equation}
	d_{ucc}\left(f,g\right) \triangleq 
	\sum_{k\in \nn_+}
	\frac{
		\sup_{\|x\|\leq k} \|f(x)- g(x)\|
	}{
		2^k\left(1 + \sup_{\|x\|\leq k} \|f(x)- g(x)\|\right)
	}
	\label{eq_definition_ucc_metric}
	.
	\end{equation}
	This metric describes the uniform convergence on compacts topology standard in the universal approximation literature such as \cite{leshno1993multilayer} and \cite{kidger2020universal}.
	
	Analogously to \cite{pmlrv75yarotsky18a} our quantitative approximation results depend on the regularity of the unknown target function.  The regularity of any $f \in C(\rrd,\rrD)$ is quantified by its (optimal) \textit{modulus of continuity}, denoted by $\omega_{f}$,\footnote{
		By the Heine-Cantor Theorem \citep[Theorem 27.6]{munkres2014topology}, any continuous functions on a compact subset of its input space, such as $[-M,M]^d\subset \rrd$, has a well-defined modulus of continuity.} 
	measures the input's space's distortion upon applying $f$ and it is defined by
	$
	\omega_{f}(\delta)\triangleq \sup_{x,y\in \rrd,\,\|x-y\|\leq \delta}
	\|f(x)-f(y)\|
	.
	$
	
	\subsubsection{Orientation-Preserving Homeomorphisms}\label{sss_topoback}
	In \cite{kratsios2020non}, it was shown that a feature map has the UAP-invariant property if and only if it is injective.  Geometrically, this is because any injective feature map is by definition, injective and continuous, and thus, as discussed in \cite{kratsios2021quantitative}, it preserves all the topological information of any compact subset of $\rrd$.  
	
	This perfect preservation of topological information is formalized by \textit{topological embeddings}.  Topological embeddings are continuous bijections which have a continuous inverse defined on their image.  These are closely related to \textit{homeomorphisms}; where by a homeomorphism on $\rrd$ we mean a bijection $\phi \in C(\rrd,\rrd)$ having a continuous inverse.  
	
	Throughout this paper, we focus on the subset $\hhh(\rrd)\subseteq C(\rrd,\rrd)$ consisting of homeomorphisms $\phi$ which preserve any the orientation of any basis of $\rrd$.  For example, no reflection in $\rrflex{2}$ belongs to $\hhh(\rrflex{2})$ but the map $x \mapsto 2x$ is.  This class is key to our analysis as it is interconnected with property (P-ii).  This is because the central result of \cite{KirbyStableHomeos1969} characterizes $\hhh(\rrd)$ as exactly describing the homeomorphisms on $\rrd$ can be continuously deformed into the identity; i.e.: there is a $\phi_{\alpha}\in C([0,1]\times \rrd, \rrd)$ satisfying:
	\begin{equation}
	\begin{aligned}
	& \phi_1(x) = \phi(x),\, &\phi_0(x)=x, \qquad
	& \phi_{\alpha} \mbox{ is a homeomorphism for each $\alpha \in [0,1]$}
	.
	\end{aligned}
	\label{eq_isotopy_definition}
	\end{equation}
	Following \cite{AdamsKnowBook}, we refer to the function $\phi_{\alpha}$ as an \textit{ambient-isotopy}.  
	
	In \cite{KirbyEdwardsFragmentationPropertyofHomeo}, it is shown that the homeomorphisms in $\hhh(\rrd)$ are characterized by their \textit{fragmentation property}.	This means that, given any $\phi \in \hhh(\rrd)$, $\delta>0$, and any $\{x_n\}_{n\leq N}$ for which $[-M,M]^d\subseteq\cup_{n\leq N}B(x_n,\delta)$ then there necessarily exist some $\{\phi_n\}_{n\leq N}\subseteq \hhh([-M,M]^d)$ satisfying
	\begin{equation}
	\begin{aligned}
	& \phi =\phi_N\circ \dots \circ \phi_1  \mbox{ and }
	& \phi_n(x-x_n)=x-x_n &\qquad (\mbox{whenever } \|x\|\geq \delta)
	.
	\end{aligned}
	\label{eq_fragmentation_property_description}
	\end{equation}
	Our interest in the fragmentation property is that it allows us to quantify the complexity of a homeomorphisms; which we rely on for our quantitative results.

	\subsection{The Space of Rotation Matrices}\label{sss_background_mat_theory}
	Key to our analysis are the higher-dimensional rotation matrices, which have recently been connected to DNNs in \cite{bansal2018can}, \cite{jia2019orthogonal}, and \cite{lezcano2019cheap}.  
	These are matrices $R$ are precisely those for which the map $x\mapsto Rx$ does not flip any basis of $\rrd$ and it preserves the distances between any two vectors.  These matrices are characterized by:
	$$
	SO(d)\triangleq \left\{
	R \in \operatorname{Mat}_{d\times d}:\,
	R^{\top}R=RR^{\top}=I_d \mbox{ and }
	\det(R)=1
	\right\}
	;
	$$
	where $I_d$ is the $d\times d$ identity matrix on $\rrd$ and $\operatorname{Mat}_{d\times d}$ denotes the set of $d\times d$ matrices.  
	
	Following \cite{knapp2002lie}, every $R\in SO(d)$ can be expressed as the \textit{matrix exponential} $\exp(A)$ of a $d\times d$-skew symmetric matrix $A$; where
	$
	\exp(A)\triangleq \sum_{k=0}^{\infty} \frac1{k!} A^k
	.
	$  
	Analogously to \cite{lezcano2019cheap}, we identify the vector space of $d\times d$-skew-symmetric matrices, denoted by $\so{d}$, with the Euclidean space of the same dimension.  This identification is realized via the bijection $\Skw:\rrflex{d(d-1)/2}\rightarrow\so{d}$ defined by
	$$
	\begin{aligned}
	\left(
	x_{1,2},\dots,x_{1,d},\dots,x_{d-1,d}
	\right)
	& \to 
	\begin{pmatrix}
	0 & x_{1,2} & \dots & x_{1,d} \\
	-x_{1,2} &  &  &  \\
	\dots &  & \ddots & x_{d-1,d} \\
	-x_{1,d} &  & -x_{d-1,d} & 0
	\end{pmatrix}
	.
	\end{aligned}
	$$
	Next, we describe the deep neural models which NEU trains.  
	\subsection{Reconfiguration Networks}\label{ss_learning_homeos}
	Recall that the DNN architecture, originating in \cite{LogicalCaluculus1943ANNBeginnings}, is built by repeatedly composing the following type of elementary functions
	$
	x\mapsto \sigma\bullet\left(Ax +b\right)
	,
	$ 
	where $A$ is a $d_i\times d_{i+1}$ matrix ($d_i,d_{i+1}\in \mathbb{N}_+$), $b \in \rrflex{d_{i+1}}$, $\sigma:\rr\rightarrow \rr$ is a non-linear function which is fixed across each feed-forward layer, and $\bullet$ denotes component-wise composition.  
	
\begin{figure}[H]
		\centering
		\begin{subfigure}[t]{0.5\textwidth}
			\centering
			\includegraphics[width=.5\linewidth]{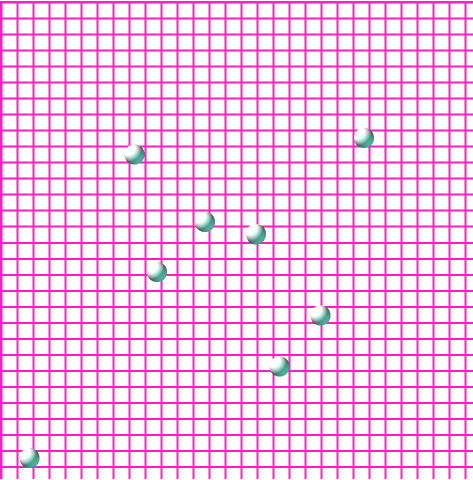}
			\caption{Input data}
		\end{subfigure}%
		\begin{subfigure}[t]{0.5\textwidth}
			\centering
			\includegraphics[width=.5\linewidth]{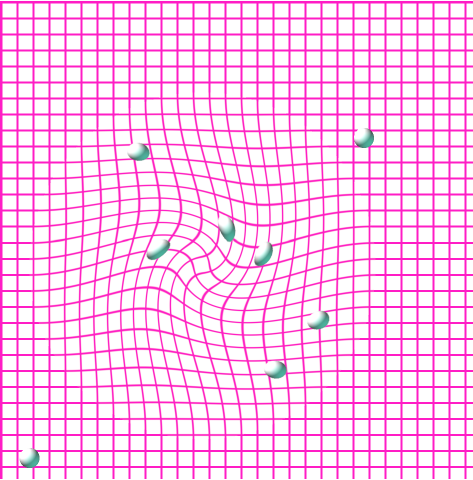}
			\caption{Reconfiguration unit's output.}
		\end{subfigure}
		\caption{Visualization of reconfiguration unit's effect.}
		\label{fig_deforms}
	\end{figure}

	NEU trains a variant of the DNN architecture whose layers are constrained between $\rrd$, but with the key difference being that the \textit{connection matrix} A is replaced by a specific $SO(d)$-valued function called a \textit{reconfiguration unit}.  These units allow the network's connection to depend, in a highly structured and non-constant way, on spatial data with the added flexibility of being able to only locally manipulate input data.  A visualization of reconfiguration units is found in Figure~\ref{fig_deforms}.
	\begin{defn}[Reconfiguration Unit]\label{defn_reconfig_unit}
		A \textit{reconfiguration unit} is a matrix-valued function $A:\rr\rightarrow \operatorname{Mat}_{d\times d}$ with representation
		\begin{equation}
		\begin{aligned}
		A(x) & \triangleq \exp\left(M_1(\|x-c\|^2) + M_2(\|x-c\|^2)
		\right)
		\\
		M_1(z)&\triangleq \Skw(W_{2,1}\circ\sigma_{\operatorname{ReLU}}\bullet W_{1,1})(z)\\
		M_2(z)&\triangleq \Skw(W_{2,2}\circ\sigma_{\operatorname{ReLU}}\bullet W_{1,2})(z)
		\sqrt[8]{ (z-\eta)(z+\eta)}I_{|z|<\eta}
		,
		\end{aligned}
		\label{eq_voila_unit_Description}
		\end{equation}
		where, for $i=0,1$, $W_{1,i}:\rr\rightarrow \rrflex{d(d-1)/2}$ and $W_{2,i}:\rrflex{d(d-1)/2}\rightarrow \rrflex{d(d-1)/2}$ are affine functions, 
		$c \in \rrd$, $\eta>0$, and where $\sigma_{\operatorname{ReLU}}(x)\triangleq \max\{0,x\}$.  
	\end{defn}
	\begin{remark}[Reconfiguration unit parameters]\label{rmk_unpacking_reconfiguration_unit}
		The map $M_2$ controls the local behaviour of $A$ and the map $M_1$ controls its global behaviour.  By setting $W_{2,2}=0$, the reconfiguration unit $A$ becomes the identity outside of the ball $\{z\in\rrd:\,\|z-c\|\leq \eta\}$. 
	\end{remark}
	We combine reconfiguration units, biases, and activation functions to build complex deep neural models.  However, again unlike DNNs, we use an activation function, which is always a homeomorphism on $\rr$.  Analogously to \cite{he2015delving} and \cite{ramachandran2017searching} we allow the activation function to depend on an additional parameter that can be used to turn the activation function into the identity map.  
	\begin{defn}[Reconfiguration Network]\label{defn_reconfig_nets}
		A \textit{reconfiguration network} is a function $\phi \in C(\rrd,\rrd)$ with representation $\phi(x)=\phi_J(x)$, where $\phi_J$ is defined iteratively via
		\begin{equation}
		\begin{aligned}
		& \phi_{n+1}(x)= \sigma_{\alpha_n}\left(A_n(\phi_{n}(x))(\phi_n(x)-c_n) \right) +b_n \qquad 
		\phi_0(x)=x,
		\end{aligned}
		\label{eq_representation_iterative_reconfiguration_network}
		\end{equation}
		for every $x \in\rrd$, some $J\in \nn_+$, some $\{c_n,b_n\}_{n=1}^N \in \rrd$, reconfiguration units $\{A_n\}_{n=1}^N$, and some $\{\alpha_n\}_{n=1}^N$ in $[0,\infty)$, where
		\begin{equation}
		\sigma_{\alpha}(x) \triangleq x + \tanh(\alpha x)
		\label{eq_swish_mod}
		.
		\end{equation}
		The set of all reconfiguration networks is denoted by $\RN$.  $J$ is called the \textit{depth} of $\phi$.  
	\end{defn}
	
	\section{Main Results}\label{s_Main_results}
	This section contains the paper's main theoretical contributions.  We begin by outlining the structured approximation capabilities of the reconfiguration networks, before describing their implications for NEU.  The section closes upon examining NEU's robustification of~\eqref{eq_naive}.  
	\subsection{Universal Orientation-Preserving Homeomorphisms}\label{ss_approx_homeo}
	We find that reconfiguration networks can approximate any homeomorphism in $\hhh(\rrd)$.  
	\begin{theorem}[{Reconfiguration Networks are Universal in $\hhh(\rrd)$}]\label{thrm_UAH}
		Let $d\in \nn_+$ with $d\geq 1$, $\phi \in \hhh(\rrd)$, $K$ be a non-empty compact subset of $\rrd$, and let $\epsilon>0$.  There exist $\phi^{\epsilon}\in \Phi_{\star:d}$ satisfying
		\begin{enumerate}[(i)]
			\item $
			\sup_{x \in K} \left\|
			\phi(x) - \phi^{\epsilon}(x)
			\right\|<\epsilon
			.
			$
			\item $\phi^{\epsilon}\in \hhh(\rrd)$.  
		\end{enumerate}
	\end{theorem}
	Theorem~\ref{thrm_UAH} is a \textit{qualitative} universal approximation result for DNN in $\hhh(\rrd)$.  However, analogously to \cite{barron1993universal} and \cite{siegel2020approximation}, by assuming some additional regularity of the homeomorphism being approximated, we may obtain a \textit{quantitative} approximation result describing the complexity of the reconfiguration network required to approximate a target homeomorphism.  
	
	Analogously to \cite{pmlrv75yarotsky18a,kratsios2021quantitative} the complexity of a reconfiguration network is quantified by its \textit{depth}.  Since homeomorphisms are more complex and structured objects than simple continuous functions, our rates depend both on the target homeomorphism's modulus of continuity, as in \cite{pmlrv75yarotsky18a}, and its best behaviour on a fragmentation; in the sense of~\eqref{eq_fragmentation_property_description}.

	Let $M,\delta>0$ and let $\omega:[0,\infty)\rightarrow [0,\infty)$ be a modulus of continuity.  We establish bounds for the class $\hhh^{\delta}_{M,\omega}(\rrd)\subseteq \hhh(\rrd)$ consisting of all homeomorphisms $\phi \in \hhh(\rrd)$ mapping $[-M,M]^d$ into itself satisfying a refinement of~\eqref{eq_fragmentation_property_description}.  Here, we additionally require that there is at-least one pair $\{x_n\}_{n\leq N}$ and $\{\varphi_n\}_{n\leq N}$ satisfying ~\eqref{eq_fragmentation_property_description} which also satisfies:
	\begin{align}
	\|\varphi_n(x)-x_n\|&=\|x-x_n\| \,
	\label{eq_definition_HMdeltaomega_1}\\
	\|\varphi_n(x-x_n) - \varphi_n(y-x_n)\|& =\|y-x\| \, & (\mbox{ if }\|x-x_n\|=\|y-x_n\|)
	\label{eq_definition_HMdeltaomega_2}\\
	\|x_n-x_m\| &>\frac{\delta}{2} \, & (\mbox{ if }n\neq m,\, n,m\leq N)
	\label{eq_definition_HMdeltaomega_3}\\
	\max_{n\leq N}\omega_{\phi_n}(u) &\leq \omega(u)
	\, &(\forall u\geq 0)
	.
	&
	\label{eq_definition_HMdeltaomega_4}
	\end{align}
	\begin{theorem}[Quantitative Approximation Rates]\label{thrm_uniform_bound}
		Fix $\delta,M>0$, $d\in\nn_+$, and a modulus of continuity $\omega$.  For any $\phi\in \hhh_{M,\omega}^{\delta}(\rrd)$, any $\nu \in \nn_+$ there is a $C\geq 0$, not depending on $d,D,\delta$ or $\omega$, and a reconfiguration network $\phi$ of depth at-most $\nu \left\lceil\frac{(8M)^dd^{d-1}}{\delta^d (d-1)}\right\rceil$, such that:
		\begin{equation}
		\sup_{\|x\|\leq M}
		\,
		\left\|
		\phi(x)-\hat{\phi}(x)
		\right\|
		\leq 
		\omega_{\nu,\delta,M,d,\omega}
		\label{eq_thrm_uniform_bound_formal_statement}
		;
		\end{equation}
		where $\omega_{\nu,\delta,M,d,\omega}\triangleq \omega_{\left\lceil\frac{(8dM)^{d}}{\delta^{d}}\right\rceil}$ and $ \omega_{\left\lceil\frac{(8dM)^{d}}{\delta^{d}}\right\rceil}$ is determined recursively by:
		\begin{equation}
		\omega_{n+1} = 2d MC \omega\left(
		\left(
		2^{-1}\nu\left\lceil
		\frac{\delta^d}{(8dM)^d}
		\right\rceil
		d(d-1)\right)^{\frac{-1}{d}}
		\right) + \omega\left(\omega_{n}\right),
		\qquad
		\omega_0=0
		.
		\label{eq_thrm_uniform_bound_formal_statement_recurrence_relation}
		\end{equation}
	\end{theorem} 
	Homeomorphisms depending on arbitrary moduli of continuity, as in Theorem~\ref{thrm_uniform_bound}, may have arbitrarily poor behaviour.  However, in many applications one has differentiability of the unknown homeomorphism $\phi$ and therefore $\phi$ is Lipschitz on $[-M,M]^d$ by the Mean-Value Theorem; i.e.: $\omega(t)=L|t|$ for some $L>0$.  In this case,~\eqref{eq_thrm_uniform_bound_formal_statement} implies the following.  
	\begin{example}[Simple Bounds for Lipschitz Homeomorphisms]\label{ex_smooth}
		In the case that $\phi \in \hhh_{M,L|\cdot|}^{\delta}$, then recursive upper-bound simplifies to 
		$
		\omega_{\nu,\delta,M,d,\omega} = \nu^{\frac{-1}{d}}\left(2 d M CL \frac{1 - L^{\frac{8d^2M}{\delta}}}{1-L}\right)
		.
		$  
	\end{example}
	
	Reconfiguration networks can memorize arbitrarily many input-output pairs without much guessing.  Analogously to \cite{MemoryCapacilty2009}, we provide an upper bound on the reconfiguration network's depth, trained on the memorization task.  
	\subsection{Memorization without Guessing}
	We quantify the size of the compact subset $K\subseteq \rrd$ on which the reconfiguration network guesses, by the number of $d$-dimensional balls of radius $\delta>0$ required to cover $K$.  This quantity is known as the $\delta$-external covering number of $K$, (see \citep[Chapter 3.5]{mohri2018foundations}), and it is denoted by $N_{\delta}(K)$.  Its advantage over the Lebesgue measure of $K$, is that it does not ignore sets of Lebesgue measure zero.\footnote{
		The two quantities are related, since the volume of a $d$-dimensional of radius $\delta>0$ is $\frac{\delta^2 \pi^{\frac{d}{2}}}{\Gamma\left(\frac{d}{2} + 1\right)}$; hence $N_{\delta}(K)\leq N$ implies $\mu(K)\leq \frac{N\delta^2 \pi^{\frac{d}{2}}}{\Gamma\left(\frac{d}{2} + 1\right)}$.  
	}
	\begin{theorem}[Quantitative Memory Capacity Bounds with Guessing Control]\label{thrm_memory_capacity_w_guessing}
		Let $d\geq 2$, $N\in \nn_+,$ $M\in \nn$, and $\{x_i\}_{i=1}^N$, $\{y_m\}_{m=1}^M$, and $\{z_i\}_{i=1}^N$ be sets of distinct points in $\rrd$ for which $x_i\neq z_i$, for $i=1,\dots,N$ and such that $y_m \not\in \{x_i,z_i\}_{i=1}^N$ for each $m\leq M$.  Define $
		\Delta \triangleq \frac1{2}\min\left\{ 2,
		\min{\|u-v\|:\, u\neq v,\, u,v \in \{x_i,z_i\}_{i=1}^N\cup\{y_m\}_{m=1}^M}
		\right\}.  
		$
		For any $0<\delta<\Delta$ there exists a reconfiguration network $\phi$ of depth $J$ and a compact subset $K\subseteq \rrd$ satisfying:
		\begin{enumerate}[(i)]
			\item $\phi(x_n)=z_n$ for each $n=1,\dots,N$.
			\item $\phi(y_m)=y_m$ for every $m=1,\dots,M$.
			\item $\phi(x)=x$ for every $x \in %
			\rrd-K%
			.$%
		\end{enumerate}
		Furthermore, the following upper-bounds on $J$ and $K$ hold:
		\begin{equation}
		\begin{aligned}
		N_{\delta}(K)\leq J
		\leq 
		\left\lceil
		\frac{N\pi}{2(\min\{2 \delta,1\})}
		\right\rceil
		.
		\end{aligned}
		\label{eq_dimensionless_interpolation_w_control_bounds}
		\end{equation}
	\end{theorem}
	In contrast, DNNs with analytic activation function always fail to have the memorization without guessing property (P-iv).  
	\begin{proposition}[No Memorization without Guessing]\label{prop_instability_theorem}
		Let $d\geq 1$.  If $\sigma$ is analytic then every $f \in \NN[d,d]$ fails (P-iv).  
	\end{proposition}
	DNNs with ReLU networks are not included in Proposition~\ref{prop_instability_theorem}.  However, the difference between these architectures and reconfiguration networks is addressed later in the paper.  
	
	\subsection{Universal Approximation via Topological Embeddings}\label{sss_UAP_cnt}
	By precomposing with an injective linear map, the universal approximation capabilities of reconfiguration networks can be extended to universal topological embeddings.  In turn, post-composing with a linear map we can approximate any continuous function.  
	\begin{theorem}[Topologically Regular Universal Approximation]\label{cor_GR_UAT}
		Fix $f \in C(\rrd,\rrD)$, a $(d+D)\times d$-matrix $A$, and $M>0$.  There exists a reconfiguration network $\hat{\phi}\in\hhh(\rrflex{d+D})$ and a $D\times (d+D)$-matrix $B$, such that
		$\mmm_{\hat{\phi},M,\epsilon}\triangleq \phi\circ(I\oplus A\cdot)[[-M,M]^d]\subseteq \rrflex{d+D}$ satisfies:
		\begin{enumerate}[(i)]
			\item \textbf{Embedding:} $\hat{\phi}\circ(I_d\oplus A\cdot)$ is a homeomorphism from $[-M,M]^d$ onto $\mmm_{\hat{\phi},M,\epsilon}$ and it is an isometry when $\mmm_{\hat{\phi},M,\epsilon}$ is equipped with the metric: $$
			d_{\hat{\phi},M,\epsilon}(z_1,z_2)\triangleq \|(I_d\oplus A)^{\dagger}\phi^{-1}(z_1) - (I_d\oplus A)^{\dagger}\phi^{-1}(z_2)\|,
			$$
			where $(I_d\oplus A)^{\dagger}\triangleq ((I_d\oplus A)^{\top}(I_d\oplus A))^{-1}(I_d\oplus A)$ and is a left-inverse of $I_d\oplus A$.
			\item \textbf{Regular Feature Space:} $\mmm_{\hat{\phi},M,\epsilon}$ is a topological submanifold of $\rrflex{d+D}$ with boundary,
			\item \textbf{Sparsity:} $L(x)=Bx$ and $B$ has exactly $D$ non-zero entries and,
			\item \textbf{Universal Approximation:} 
			$\sup_{\|x\|\leq M}
			\left\|
			f(x) - B\hat{\phi}(x,Ax)
			\right\|< \epsilon
			.
			$
		\end{enumerate}
	\end{theorem}

	Let us compare the topological regular approximation results of this section with the popular approach of generating $\phi$ with a DNN with ReLU activation function.  We frame our result in the setting of generalized-ReLU networks, as defined in \cite{gribonval2020approximation}.  
	\begin{proposition}\label{ex_ReLU_type_Example}
		Let $\sigma_{r-\operatorname{ReLU}}(x)=\max\{0,x^r\}$ and $0<d$.  If $\hat{f} \in \NN[d,D][\sigma_{r-\operatorname{ReLU}}]$ and $\hat{f}$ is UAP-preserving, then for every affine function $W:\rrd\rightarrow \rrd$ the deep ReLU network:
		$$
		\tilde{f} = \hat{f}\circ \sigma_{r-\operatorname{ReLU}}\bullet W
		,
		$$
		is not UAP-preserving.  In particular, it is not a topological embedding.  
	\end{proposition}
	Proposition~\ref{ex_ReLU_type_Example} highlights the main geometric difference between DNNs and reconfiguration networks. Namely, the latter always represents the input space as an embedded topological submanifold of the feature space $\rrF$ whereas the typically does not.  
	
	\begin{remark}[Discussion: Comparison with Kernel Methods]\label{rmk_first_comparison_kernel_methods}
		Kernel methods implicitly linearize functions in $C(\rrd,\rrD)$ by representing them within a high-dimension space.  In contrast, Theorem~\ref{cor_GR_UAT} guarantees that reconfiguration networks can perform this linearizing in a $d+D$-dimensional space.
		
		To see this, we consider a familiar example.  Consider the kernel $K_{\psi}(x,y)\triangleq \sum_{n=1}^{\infty} \frac1{2^{2n}} x^n y^n$ on $[0,1]$.  This Kernel is universal, in the sense that:
		$$
		\left\{
		\sum_{n=0}^F\beta_n K_{\psi}(\cdot,y_n):\, F \in \nn,\, y_1,\dots,y_F \in [0,1],\, \beta_1,\dots,\beta_F \in \rr
		\right\},
		$$ is a dense subset of $C([0,1],\rr)$\footnote{This is a direct consequence of \citep[Theorem 7]{MicchelliUniversalFeature} and the Weierstra{\ss} Approximation Theorem.}.    
		Moreover, as discussed in \citep[Section 3]{MicchelliUniversalFeature} the feature map $\tilde{\phi}$ associated to $K_{\psi}$ is 
		maps any $x \in [0,1]$ to the sequence $(\frac{x^n}{2^n})_{n=0}^{\infty} \in \ell^2(\nn)$.  Thus, for any $f \in C([0,1],\rr)$, there exists a linear map $\tilde{B}:\ell^2\rightarrow \rr$ such that
		\begin{equation}
		\max_{x \in [0,1]}\|\tilde{B}\circ \tilde{\phi}(x)-f(x)\|<\epsilon
		\label{eq_high_dimensional_representation}
		.
		\end{equation}
		The contrast between equation~\eqref{eq_high_dimensional_representation} and Theorem~\ref{cor_GR_UAT} (iv) is that, for every $\epsilon>0$, together the maps $B$ and $\hat{\phi}(I_d\oplus A \cdot)$ only require $2$ dimensions in order to approximately linearize the function $f$; whereas together $\tilde{B}$ and $\tilde{\phi}$ need infinitely many dimensions to do so.  Therefore, amongst other things, Theorem~\ref{cor_GR_UAT} can be interpreted as an explicit low-dimensional analogue of a kernel methods which are implicit and high-dimensional.  
	\end{remark}

	\subsection{Non-Euclidean Upgrading}\label{ss_NEU_properties_and_results}
	We close the approximation-theoretic portion of this paper by related our results on reconfiguration networks back to the learning problem~\eqref{eq_naive} and to NEU.  We present two results, each offering a different perspective on the improvement which can be gained by NEU.  Both qualitative and quantitative results are provided.  We operate under the following assumptions; typical in non-convex optimization (see \cite{DalMasoGamma1993}).  
	\begin{ass}\label{ass_optimizability}
		We assume the following regularity of $L$ and $P$ defining Problem~\eqref{eq_naive}:
		\begin{enumerate}[(i)]
			\item $L:\rrD\times \rrD\times \rrd\rightarrow \rr$ is continuous, and bounded-below.
			\item $P:C(\rrd,\rrD)\rightarrow \rr$ is continuous, bounded-below, and coercive; i.e.: for every $t \in \rr$ the sub-level set:
			$$
			\left\{
			P(f)\leq t:\, f \in C(\rrd,\rrD)
			\right\}
			,
			$$
			is compact in $C(\rrd,\rrD)$.  
		\end{enumerate}
	\end{ass}
	\begin{remark}[Why not lower semi-continuity of $L$ and of $P$?]\label{remark_lsc_vs_cnt}
		In general non-convex settings, lower semi-continuous (lsc) objective functions are typically considered instead of continuous ones.  However, \cite{LaszloSzilard2017MinimaxDenseSubsetLSC} shows that if our objective function is not continuous but is lsc and, in addition, if we do not optimize this objective function over the entire space but instead only optimize it over a proper dense subset (such as $\{f\circ \phi:f\in \fff,\, \phi\in \Phi_{\star:d+D}\}$ or $\fff$ if it is a universal model class), then the global optimum is typically unobtainable.  However, in \citep[Corollary 3.4]{LaszloSzilard2017MinimaxDenseSubsetLSC} the authors shows that this is never a theoretical issue for continuous objective functions.  
	\end{remark}
	
	The following result says that given a family of models $\fff$ which is at-least able to express linear functions, there must be a UAP-invariant feature map in $\Phi_{\star:d+D}$ which "upgrades" $\fff$ unit it approximately achieves the optimal value of the learning problem~\eqref{eq_naive}.  Moreover, the representation learned by the reconfiguration network never needs to in dimension above $d+D$.  Furthermore, the representation produced by the reconfiguration networks is an embedded topological submanifold of the feature space $\rrflex{F}=\rrflex{d+D}$ and the feature map is a topological embedding.  
	\begin{corollary}[Non-Euclidean Upgrading I]\label{cor_NEU_1}
		Let $\{x_n\}_{n=1}^N$ be a subset of $\rrd$ and suppose that $L$ and $P$ satisfy Assumption~\ref{ass_optimizability}.  Suppose also that $\fff\subseteq C(\rrflex{d+D},\rrD)$ contains all linear maps from $\rrflex{d+D}$ to $\rrD$; i.e.:
		$$
		\left\{
		x\mapsto Ax+b:\, b\in \rrD,\, A\in \operatorname{Mat}_{D\times d+D}
		\right\}
		.
		$$
		Then, for every $\epsilon,M>0$ and every full-rank matrix $A\in \operatorname{Mat}_{d,d+D}$, there exists some $\hat{f}^{\epsilon}\in \fff$ and some $\hat{\phi}^{\epsilon}\in \Phi_{\star:d+D}$ and some $f^{\epsilon}\in \fff$ such that: 
		\begin{enumerate}[(i)]
			\item The $\epsilon$-optimality criterion holds:
			\begin{equation}
			\begin{aligned}
			&\sum_{n=1}^N L
			\left(
			f(x_n),
			f^{\epsilon}\circ \hat{\phi}^{\epsilon}((I_d\oplus A)x_n)
			,x_n
			\right)+ P\left(
			f^{\epsilon}\circ \hat{\phi}^{\epsilon}((I_d\oplus A)\cdot)
			\right)
			\\
			< &
			\epsilon +
			\inf_{g \in C(X,\rrD)}
			\sum_{n=1}^N L
			\left(
			f(x_n),
			g(x_n)
			,x_n
			\right)+ P\left(
			g(x_n)
			\right)
			.
			\end{aligned}
			\label{cor_NEU_Formulation}    
			\end{equation}
			\item $\hat{\phi}^{\epsilon}\circ(I_d\oplus A\cdot)$ is a homeomorphism from $[-M,M]^d$ onto $\mmm_{\hat{\phi}^{\epsilon},M,\epsilon}$ and it is an isometry when $\mmm_{\hat{\phi}^{\epsilon},M,\epsilon}$ is equipped with the metric: $$
			d_{\hat{\phi}^{\epsilon},M,\epsilon}(z_1,z_2)\triangleq \|(I_d\oplus A)^{\dagger}\phi^{-1}(z_1) - (I_d\oplus A)^{\dagger}\phi^{-1}(z_2)\|.
			$$
			\item $\mmm_{\hat{\phi}^{\epsilon},M,\epsilon}$ is a topological submanifold of $\rrflex{d+D}$ with boundary.
		\end{enumerate}
	\end{corollary}
	\begin{remark}
		In particular, Corollary~\ref{cor_NEU_1} (i), implies that the \textit{"upgraded model"}: $f^{\epsilon}\hat{\phi}^{\epsilon}((I_d\oplus A)\cdot)$ must achieve a lower value of the objective function of the training problem~\eqref{eq_naive}.  
	\end{remark}
	
	The following result says that reconfiguration networks can modify models in $\fff$ to match the optimizer of Problem~\eqref{eq_naive} at finitely every observed data-point while leaving almost all of their other input-output pairs unaltered.  The result is quantitative in the number of points modified, and the external-covering number of the set of input-output pairs left unaltered.  
	\begin{corollary}[Non-Euclidean Upgrading II]\label{cor_NEU_2}
		Let $\{x_n\}_{n=1}^N$ belong to a compact subset $X$ of $\rrd$ and suppose that $L$ and $P$ satisfy Assumption~\ref{ass_optimizability}.  Suppose also that $\fff\subseteq C(\rrflex{d+D},\rrD)$ contains all linear maps from $\rrflex{d+D}$ to $\rrD$; i.e.:
		$$
		\left\{
		x\mapsto Ax+b:\, b\in \rrD,\, A\in \operatorname{Mat}_{D\times d+D}
		\right\}
		.
		$$
		Then, there exists an $f^{\star} \in C(X,\rrD)$ satisfying
		$$
		f^{\star}\in 
		\underset{g\in C(X,\rrD)}{\operatorname{argmin}}\,
		\sum_{n=1}^N L\left(
		f(x_n),g(x_n),x_n
		\right) + P(g)
		.
		$$
		Moreover, if $f^{\star}(x_n)\neq f^{\star}(x_m)$ whenever $n\neq m$ and $x_n\neq 0$ for $n\leq N$, then there exists some $\hat{f}\in \fff$ such that for every
		\begin{equation}
		0<\delta< 
		\frac1{2}
		\min\left\{
		\|u-v\|:\, u\neq v
		\mbox{ and }
		u,v \in \{f^{\star}(x_n),x_m\}_{n,m=1}^N
		\right\},
		\label{eq_condition_delta_NEU}
		\end{equation}
		there exists a reconfiguration network $\hat{\phi}^{\delta}\in \Phi_{\star:d+D}$ satisfying:
		\begin{enumerate}[(i)]
			\item $\hat{f}\circ \hat{\phi}^{\delta}(Ax_n)= f^{\star}(x_n)$ for every $n\leq N$,
			\item $
			N_{\delta}\left(
			\left\{
			(x,\hat{f}(Ax))\in \rrflex{d+D}:\, (x,\hat{f}(Ax))\neq \hat{\phi}^{\delta}(x,\hat{f}(Ax))
			\right\}
			\right)
			\leq 
			\left\lceil
			\frac{N\pi}{2(\min\{2 \delta,1\})}
			\right\rceil
			,	
			$
			\item $\hat{\phi}^{\delta}$ has depth at-most $\left\lceil
			\frac{N\pi}{2(\min\{2 \delta,1\})}
			\right\rceil$.
		\end{enumerate}
	\end{corollary}	
	\begin{remark}\label{remark_why_not_Lebesgue}
		Corollary~\ref{cor_NEU_2} (ii) highlights the need to quantify the smallness of the set $K$ in Theorem~\ref{thrm_memory_capacity_w_guessing} with its external covering number instead of its Lebesgue measure.  This is because, the set being described in Corollary~\ref{cor_NEU_2} (ii) is a $d$-dimensional subset of $\rrflex{d+D}$; therefore, it is of Lebesgue measure $0$.  However, within the context of Corollary~\ref{cor_NEU_2}, this set is not negligible as it describes the set of input-output pairs which are transformed by the feature map $\hat{\phi}^{\delta}$.  
	\end{remark}
	These two results show that NEU embeds a great deal of flexibility into any model class $\fff$ with a basic level of expressibility.  The next section describes how to counter-balance this flexibility by robustifying the learning problem~\eqref{eq_naive}.  
	\subsection{Robustification of Loss-Function for Improved Generalization}
	Fix $M>0$, a non-empty training set $\{x_n\}_{n\leq N}\subseteq [-M,M]^d$, and a model $\hat{f}$ for~\eqref{eq_naive}.  We can evaluate its generalizability on $[-M,M]^d$ outside the training data $\{x_n\}_{n\leq N}$ by the gap in the error on the training data $\sum_{n\leq N}\frac{
		L(f(x_n),\hat{f}(x_n),x_n)
	}
	{N}  + P(f)$ and the worst-case scenario error on $[-M,M]^d$.  We define this gap by
	\begin{equation}
	\begin{aligned}
	\sup_{\|x\|\leq M}\,
	L(f(x_n),\hat{f}(x_n),x_n)  
	+ P(f)
	-
	\left[
	\frac1{N}
	\sum_{n\leq N}\frac{
		L(f(x_n),\hat{f}(x_n),x_n)
	}
	{N}  + P(f)
	\right] 
	.
	\end{aligned}
	\label{eq_typical_representativness}
	\end{equation}
	A-priori it seems that, for any given $\hat{f}$, all the quantities in~\eqref{eq_typical_representativness} are fixed by the learning problem.  In fact, this is not the case here as we have implicitly made the assumption that the weight of each training data-point pulls equal weight on the left-hand side of~\eqref{eq_typical_representativness}.  
	
	Accordingly, we re-weight the training objective function to
	$\sum_{n\leq N}
	w_nL(f(x_n),\hat{f}(x_n),x_n)
	+ P(f)$ with new weights $\{w_n\}_{n\leq N}$ in $[0,1]$ summing to $1$.  Thus, we improve the generalizability of our model $\hat{f}$ by extending~\eqref{eq_naive} by coupling it with the following extension of~\eqref{eq_typical_representativness}
	\begin{equation}
	\hspace*{-1em}
	\begin{aligned}
	\min_{\hat{f} \in \fff}
	\,\sum_{n\leq N} w_n & L(f(x_n),\hat{f}(x_n),x_n) + P(\hat{f})\\
	\mbox{where:} 
	\underset{
		\underset{w_n\in [0,1] }{
			\sum_{n\leq N} w_n=1
		} 
	}{\operatorname{argmin}}
	\,
	\sup_{\|x\|\leq M}
	&
	\,
	L(f(x_n),\hat{f}(x_n),x_n)  
	+ P(f)\\
	-&
	\left[
	\sum_{n\leq N}
	w_n
	L(f(x_n),\hat{f}(x_n),x_n)  + P(f)
	\right] 
	.
	\end{aligned}
	\label{eq_typical_representativness_modded}
	\end{equation} 
	The multi-function $\operatorname{argmin}$ is invariant under addition.  Therefore, we may simplify the constraint in~\eqref{eq_typical_representativness_modded}.  Thus, we are interested in the following equivalent optimization problem:
	\begin{equation}
	\begin{aligned}
	\underset{
		\underset{w_n\in [0,1] }{
			\sum_{n\leq N} w_n=1
		} 
	}{\operatorname{argmax}}
	\,
	\sum_{n\leq N} w_n
	L(f(x_n),\hat{f}(x_n),x_n)
	.
	\end{aligned}
	\label{eq_typical_representativness_modded_redux}
	\end{equation} 
	The key advantage of~\eqref{eq_typical_representativness_modded_redux} over~\eqref{eq_typical_representativness_modded} is that it is completely independent of the behaviour of our model's test-set performance quantified by $\sup_{\|x\|\leq M}\,
	L(f(x_n),\hat{f}(x_n),x_n)  
	+ P(f)$.  Thus, any optimizer of~\eqref{eq_typical_representativness_modded_redux} can be computed independently of any test-set information.  
	
	Nevertheless, problem~\eqref{eq_typical_representativness_modded_redux} is generally ill-posed.  In order to identify a good set of weights $\{w_n\}_{n\leq N}$, we interpret any set of weights $\{w_n\}_{n\leq N}$ as describing a discrete probability measure on $\{1,\dots,N\}$.  Hence, by adding the following Kullback-Leibler divergence between the discrete probability measures implicitly defined by $\{w_n\}_{n\leq N}$ and the uniform probability measure on $\{1,\dots,N\}$ implicitly specified by the naive weighting scheme $\{\frac{1}{N}\}_{n\leq N}$ we obtain the following well-posed variant of~\eqref{eq_typical_representativness_modded_redux}, with hyper-parameter $\lambda>0$
	\begin{equation}
	\begin{aligned}
	\underset{
		\underset{w_n\in [0,1] }{
			\sum_{n\leq N} w_n=1
		} 
	}{\operatorname{argmax}}
	\,
	\sum_{n\leq N} w_n
	L(f(x_n),\hat{f}(x_n),x_n)
	- \lambda \sum_{n\leq N} w_n\log\left(\frac{w_n}{N}\right)
	.
	\end{aligned}
	\label{eq_typical_representativness_modded_redux_final}
	\end{equation}
	\begin{theorem}[{Optimal Robust Weights for~\eqref{eq_typical_representativness_modded_redux_final}}]\label{thrm_Robustification}
		Let $\{x_n\}_{n\leq N}$ be a non-empty training dataset in $\rrd$ and $L$ be continuous.  Then $\{w_{n}^{\lambda,\hat{f}}\}_{n=1}^N$ belongs to~\eqref{eq_typical_representativness_modded_redux_final}, where
		$$
		w_{n}^{\lambda,\hat{f}}
		\triangleq
		\frac{e^{\lambda^{-1}L(f(x_n),\hat{f}(x_n),x_n)}}{\sum_{n\leq N} e^{\lambda^{-1}L(f(x_n),\hat{f}(x_n),x_n)}}
		.
		$$
	\end{theorem}
	Moreover, the robust learning Problem~\eqref{eq_typical_representativness_modded} is equal to
	\begin{equation}
	\min_{\hat{f} \in \fff} \,\sum_{n\leq N} 
	\frac{e^{\lambda^{-1}L(f(x_n),\hat{f}(x_n),x_n)}
		L(f(x_n),\hat{f}(x_n),x_n)
	}{\sum_{n\leq N} e^{\lambda^{-1}L(f(x_n),\hat{f}(x_n),x_n)}}
	+ P(\hat{f})
	\label{eq_optima_train_loss}
	.
	\end{equation}
	By the representation~\eqref{eq_typical_representativness_modded}, the learning problem~\eqref{eq_optima_train_loss} necessarily yields better generalizability than its naive counterpart~\eqref{eq_naive}.  This generalization improvement is quantified by the gap between~\eqref{eq_typical_representativness} and the value of the constraint of~\eqref{eq_typical_representativness_modded}.  Given a model $\hat{f}\in \fff$ and weights $\{w_n\}_{n\leq N}$ in $[0,1]$ summing to $1$ we dfine
	$$
	\Gen \triangleq 
	\sup_{\|x\|\leq M}\,
	L(f(x_n),\hat{f}(x_n),x_n)  
	+ P(f)
	-
	\left[
	\sum_{n\leq N}
	w_n
	L(f(x_n),\hat{f}(x_n),x_n)  + P(f)
	\right] 
	.
	$$
	
	\begin{corollary}[NEU's Loss-Function Modification Improves in Generalizability]\label{cor_gain_quantification}
		Let $x_1,\dots,x_N \in\rrd$, $\hat{f}\in \fff$, and $M>0$.  Then $\Gen[\hat{f}][\left\{\frac1{N}\right\}_{n\leq N}] 
		- 
		\Gen[\hat{f}][\{w_{n}^{\lambda,\hat{f}}\}_{n\leq N}]$ equals to:
		$$
		\begin{aligned}
		\sum_{n\leq N} 
		\left(
		\frac{
			e^{-\lambda^{-1}L(f(x_n),\hat{f}(x_n),x_n)}
		}{
			\sum_{n\leq N} e^{-\lambda^{-1}L(f(x_n),\hat{f}(x_n),x_n)}
		}
		-
		\frac1{N}
		\right)
		L(f(x_n),\hat{f}(x_n),x_n)
		\geq 0.
		\end{aligned}
		$$
	\end{corollary}
	
	We end the this portion of the paper with the following observation.  
	\subsection*{Training Very Deep UAP-invariant Feature Maps}\label{sss_NEU_definition}
	From the computational standpoint, property (P-i) can be used to to subdivide step 1 of Meta-Algorithm~\ref{algo_NEU_redux} into an incremental procedure, analogously to \cite{bengio2007greedy,JMLRv10larochelle09a}, allowing for the handling of extremely deep feature maps without negatively impacting the model's UAP.  This incremental procedure, summarized by sub-routine~\ref{algo_NEU}, views the feature map $\hat{\phi}$ Meta-Algorithm~\ref{algo_NEU_redux} step 1 as a composition $\hat{\phi}=\hat{\phi}_{\hat{I}}\circ \dots \circ \hat{\phi}_{1}$ of deep reconfiguration networks $\{\hat{\phi}_i\}_{i\leq \hat{I}}$ trained in a loop.  
	
	\renewcommand*{\algorithmcfname}{Sub-Routine}
	\begin{algorithm}
		\SetAlgoLined
		\SetKwInOut{Input}{input}\SetKwInOut{Output}{output}
		\Input{Loss-function $L$, penalty function $P$, \\
			Training and Validation Data $\{x_n\}_{n\leq N}=\{x_{n,t}\}_{n\leq N_t}\cup	
			\{x_{n,v}\}_{n\leq N_v}$\\
			Number of blocks $I$ of Feature Map, $J$ depth per block\\
			Robustness Hyper-parameter $\lambda>0$, $F \in \nn_+$\\
		}
		\Output{NEU-Feature Map $\hat{\phi}\triangleq 
			\hat{\phi}_{\hat{I}}\circ \dots \circ \hat{\phi}_1.
			$}
		\BlankLine
		\For{$i\leq I$
		}{
			$
			\hat{\phi} \in \underset{\phi \in \RN[\star:F]}{\operatorname{argmin}}
			\sum_{n\leq N} w_n^{\star,\lambda}\,
			L\left(f(x_n),A\phi\circ \hat{\phi}_{i-1}(x_n)+b,x_n\right) + P(A\phi\circ \hat{\phi}_{i-1}+b)
			$
			\;
			$
			\phi_{i} = \hat{\phi}\circ \hat{\phi}_i
			$
			\;
			$
			\hat{A}_i,\hat{b}_i \in \underset{{A,b}}{\operatorname{argmin}}
			\sum_{n\leq N_t} w_n^{\star,\lambda}\,
			L\left(f(x_{n,t}),A\hat{\phi}_i(x_{n,t})+b,x_{n,t}\right) + P(A\hat{\phi}_i+b)
			$
		}
		$
		\hat{I} \in 
		\underset{i \leq I}{\operatorname{argmin}}
		\sum_{n\leq N_v} w_n^{\star,\lambda}\,
		L\left(f(x_{n,v}),\hat{A}_i\hat{\phi}_i(x_{n,v})+\hat{b}_i,x_{n,v}\right) + P(\hat{A}_i \hat{\phi}_i+\hat{b}_i)
		$ 
		\caption{Incremental training of very deep UAP preserving feature maps.}
		\label{algo_NEU}
	\end{algorithm}
	
	Sub-routine~\ref{algo_NEU} would typically yield sub-optimal $\hat{\phi}$. However, by replacing step 1 of Meta-Algorithm~\ref{algo_NEU_redux} with sub-routine~\ref{algo_NEU}, we can train feature maps that are too deep to train on a single machine while also guaranteeing that these maps have property (P-i).  In contrast, Proposition~\ref{ex_ReLU_type_Example} guarantees that DNNs with the ReLU activation function cannot be trained analogously without disrupting the DNN's UAP.  
	
	\section{Numerical Evaluation of NEU-OLS and NEU-PCA}\label{s_Implementations}
	Next, we evaluate the performance of NEU across various learning tasks.  First, we investigate the performance of NEU in the chaotic environment provided by real-world financial data.  Then, we stress test NEU's behaviour within the controlled environment provided by simulation studies.  The Tensorflow (v.2.4.1) code and data-sets for our implementations is available online at \cite{NEUGit}.  
	
	\subsection{Financial Data Analysis}\label{ss_dd_studs}
	The performance of the NEU meta-algorithm will be investigated both on regression and dimension reduction tasks using financial data.  We begin with the regression problem of constructing a stock return replication and then move to non-Euclidean yield-curve analysis.  
	
	\subsubsection{Regression Analysis: Apple Stock Tracker}\label{ex_regression_analysis}
	Predicting the relationship between the price of a set of assets is central to many trading strategies.  %
	For example, strategies that rely on illiquid assets may create a portfolio comprised entirely of liquid assets that tracks the illiquid asset's movements.  In this example the technique is demonstrated using liquid stocks for both the target and the tracking portfolio so we can better evaluate performance, which would be more difficult with illiquid assets due to missing contemporaneous prices for the illiquid target.   
	
	We consider Apple's stock as the target asset while the tracking portfolio is comprised of IBM, Google, Cisco Systems, Microsoft, Acacia Communications, NXP Semiconductors NV, Qualcomm, Analog Devices, Glu Mobile, Jabil, Micron, and STMicroelectronics NV.  Thus, the tracking portfolio is comprised of the stock of major companies in the same industry and  Apple's supply chain (see \cite{APPLsc} and \cite{TechGiantsAPPL}).  
	
	\begin{figure}
		\centering
		\includegraphics[width=.8\textwidth]{%
			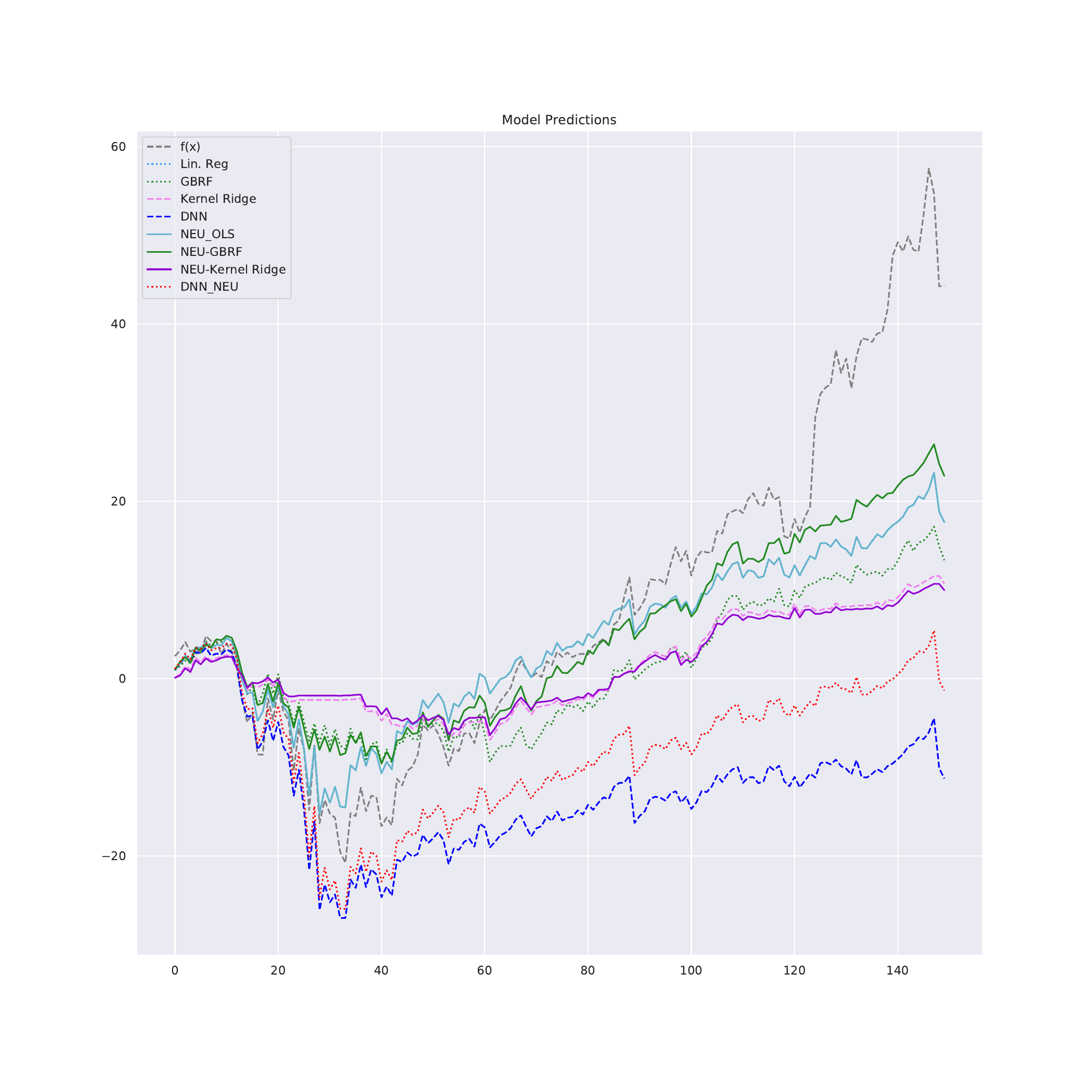}
		\caption{Regression models' 150 day-ahead out-of-sample predictions.}
		\label{fig_trackers_on_test_set}
	\end{figure} 
	
	We build a tracking portfolio using various linear, non-linear, and discontinuous regression model classes.  These include Elastic Net regularization of \cite{zou2005regularization}, generalizing the LASSO regressor of \cite{tibshirani1996regression} and Tychonov regularization of \cite{tikhonov1963} (Elastic Net), kernel ridge regression (Kernel), gradient boosted random forests (GBRF), and a DNN with ReLU activation function.  Each of the hyper-parameters is selected by cross-validations and randomized search from a large grid, hyper-parameters include the choice of kernel.  The NEU version of these models is also used considered as a general evaluation of the improvement capabilities of NEU.

	We consider $2$ years of closing stock prices, ending on September $25^{th}$ $2020$, to compute the regression weights.  The models are trained on the first $75 \%$ of the data and the remaining $25 \%$ is used to evaluate the out-of-sample predictive performance of the trained models, and is illustrated in Figure~\ref{fig_trackers_on_test_set}.  
	
	\begin{table}
		\centering
		\begin{tabular}{lrrrrr}
			\toprule
			{Train} &   Er. 95L &      Er. Mean &   Er. 95U &       MAE &       MSE  \\
			\midrule
			NEU-ENET           & -0.057708 &  6.552474e-09 &  0.058055 &  0.466568 &  0.455442  \\
			ENET        & -0.054082 &  3.295653e-18 &  0.054309 &  0.452092 &  0.413066 \\
			NEU-GBRF           & -0.069073 &  0.000000e+00 &  0.068067 &  0.525591 &  0.647727  \\
			GBRF               & -0.066513 &  2.966087e-17 &  0.066309 &  0.488983 &  0.622644  \\
			NEU-kRidge       & -0.005229 & -8.935916e-05 &  0.005009 &  0.044634 &  0.003772  \\
			kRidge             & -0.038941 & -5.949302e-04 &  0.037408 &  0.326579 &  0.200016  \\
			NEU-OLS            & -0.046744 &  5.180375e-03 &  0.055844 &  0.424678 &  0.361532  \\
			NEU-DNN           & -0.026579 &  2.685197e-02 &  0.081460 &  0.439256 &  0.411793  \\
			DNN               & -0.021030 &  3.271767e-02 &  0.088175 &  0.445274 &  0.427985  \\
			\bottomrule
			\label{tab_ins_perform_STOCK}
		\end{tabular}
		\caption{Train performance of fixed time-horizon problem.}
	\end{table} 
	
	\begin{table}[H]
		\centering
		\begin{tabular}{lrrrrr}
			\toprule
			{Test} &   Er. 95L &  Er. Mean &   Er. 95U &       MAE &       MSE  \\
			\midrule
			NEU-ENET           & -0.159483 &  0.407651 &  1.004434 &  1.448866 &  5.369902  \\
			ENET        & -0.166303 &  0.364576 &  0.929381 &  1.368901 &  4.903049 \\
			NEU-GBRF           & -0.096924 &  0.492442 &  1.096471 &  1.587017 &  5.989319  \\
			GBRF               & -0.168434 &  0.427108 &  1.052304 &  1.591232 &  5.971403  \\
			NEU-kRidge       & -0.150280 &  0.411124 &  1.026502 &  1.429903 &  5.515611  \\
			kRidge             & -0.169045 &  0.380833 &  0.958410 &  1.427682 &  5.110043  \\
			NEU-OLS            & -0.167247 &  0.344157 &  0.929869 &  1.309060 &  4.816255 \\
			NEU-DNN           & -0.062913 &  0.455043 &  1.034286 &  1.407882 &  4.948565  \\
			DNN               & -0.079580 &  0.462161 &  1.049870 &  1.444588 &  5.112289  \\
			\bottomrule
		\end{tabular}
		\caption{Test performance of fixed time-horizon problem.}
		\label{tab_oos_perform_STOCK}
	\end{table} 
	
	NEU-OLS and the DNN model both outperform each of the linear models.  However, NEU-OLS's out-performance of the DNN model is a joint effort between its representation properties and its robustness properties.  The in-sample advantage can be explained by NEU's memory capacity, as demonstrated by Theorem~\ref{thrm_memory_capacity_w_guessing}, and its expressibility improvement, as demonstrated by Theorem~\ref{cor_GR_UAT}.  The out-of-sample performance, described in Table~\ref{tab_oos_perform_STOCK}, has also benefited from the robustness of the NEU weights, described by Theorem~\ref{thrm_Robustification}.

	\subsubsection{Dimension Reduction: US-Bond Yield Curve}\label{ex_PCA_German}
	Principal component analysis (PCA) is commonly used in finance  to reduce the effective dimension of data and a classical application is for representing the yield curve corresponding to zero-coupon bond prices.  Denote by $B(t,T)$ the price at time $t$ of a zero-coupon bond that pays the face value, by assumption \$1, at maturity $T$.  The bond's yield, denoted $y(t,T)$, is the continuously compounded interest rate at which an investment of $B(t,T)$ would accumulate to the face value.  That is,
	$
	y(t,T)\triangleq -\frac{\ln\left(B(t,T)\right)}{T-t}
	.
	$  
	The yield curve is the map of a bond's yields as a function of time to maturity, $(T-t) \rightarrow y(t,T)$.  Since the bond prices for all maturity dates $T\geq t$ are not observed it is an important problem for a variety of financial applications to construct the curve using the available observed bond prices at a given time.
	%
	We benchmark NEU against auto-encoders (AE) with bottleneck dimension equal to the number of principal components (or factors) and against kernel-PCA (kPCA), two popular non-linear alternatives to classical PCA.  NEU variants of both these methods are also considered.

	
	The daily bond data considered in this example consists of 6385 consecutive instances of stripped US government bond prices between June $8^{th}$ 1990 to April 2$^{rst}$ 2017.  Each instance records the value of zero-coupon bonds with 1, 3, and 6 month, and 1, 2, 3, 5, 7, 10, 20, 30 year maturities.  
	
	We shall consider the performance of PCA, kPCA, NEU-PCA, NEU-kPCA, a deep auto-encoder (AE), and NEU-AE. The test set consists on instance ahead yield curves, and thus it measures the robustness to the dimension reduced yield-curves factor models to market movements.

	Figures~\ref{fig_PCA_Sums}-\ref{fig_PCA_Sums_2} show that NEU-AE's performs best from all the proposed models both in and out-of-sample when using one, two, and three factors.  NEU-PCA only becomes competitive, both on the training and testing sets, when three factors are utilised.  This observation highlights the importance of NEU's UAP-invariance property (P-i) as, in this case, NEU is able to maintain and improve the expressiveness of the auto-encoder model.  
	
	\begin{table}[H]
		\centering 
		\begin{tabular}{lrrrr}
			\toprule
			{} &       Test-MAE &        Test-MSE 
			&       Train-MAE &        Train-MSE \\
			\midrule
			NEU-PCA  &  3.214483 &  15.518402  & 3.406530 &  16.814745 \\
			PCA      &  3.218449 &  15.302093    &  2.972712 &  13.817181 \\
			NEU-AE   &  2.715108 &  11.043391  & 3.010322 &  13.117202 \\
			AE       &  3.172362 &  14.792091  & 2.950363 &  13.495282 \\
			NEU-kPCA &  3.554145 &  18.568798 & 3.687914 &  19.654074 \\
			kPCA     &  3.253660 &  15.626541 &  3.067545 &  14.513802 \\
			\bottomrule
		\end{tabular}
		\caption{Performance of reconstructed factor models - 1 Factor.} 
		\label{fig_PCA_Sums} 
	\end{table} 
	
	\begin{table}[H]
		\centering 
		\begin{tabular}{lrrrr}
			\toprule
			{} &       Test-MAE &        Test-MSE 
			&       Train-MAE &        Train-MSE \\
			\midrule
			NEU-PCA  &  3.189997 &  14.913756  & 3.227090 &  15.089537 \\
			PCA      &  3.195051 &  15.143381    &  2.955438 &  13.675315 \\
			NEU-AE   &  2.623852 &  10.425360  & 2.685428 &  10.647568 \\
			AE       &  2.827589 &  11.930328  & 2.703830 &  11.108966 \\
			NEU-kPCA &  3.842316 &  21.080980 & 3.697985 &  20.525725 \\
			kPCA     &  3.252147 &  15.614535 &  3.064285 &  14.491074 \\
			\bottomrule
		\end{tabular}
		\caption{Performance of reconstructed factor models - 2 Factors.} 
		\label{fig_PCA_Sums_!} 
	\end{table} 
	
	\begin{table}[H]
		\centering 
		\begin{tabular}{lrrrr}
			\toprule
			{} &       Test-MAE &        Test-MSE 
			&       Train-MAE &        Train-MSE \\
			\midrule
			NEU-PCA  &  3.241300 &  14.439484  & 2.755066 &  11.301304 \\
			PCA      &  3.197890 &  15.176116    &  2.933380 &  13.557278 \\
			NEU-AE   &  3.093146 &  13.579660  & 2.613067 &  10.845548 \\
			AE       &  3.162107 &  14.688698  & 2.940275 &  13.391631 \\
			NEU-kPCA &  3.883233 &  20.833542 & 3.464134 &  17.237757 \\
			kPCA     &  3.252588 &  15.178260   &  3.064387 &  14.493373 \\
			\bottomrule
		\end{tabular}
		\caption{Performance of reconstructed factor models - 3 Factors.} 
		\label{fig_PCA_Sums_2} 
	\end{table}

	In each case, NEU-PCA and NEU-AE reconstructs the yield curve more accurately from a small number of learned driving factors.  We find that the in and out of sample explanatory capabilities of NEU-PCA surpass even the auto-encoder.  As expected, NEU-AE offers the best performance amongst all the models, however, the advantages over NEU-PCA is nevertheless marginal.  As with the regression tasks, the kPCA's rigid feature map negatively interacts with NEU's feature map causing instability.  
	
	\begin{remark}[Clashing NEU Features and Kernel Features]\label{remark_clash_NEU_kernel}
		This last point is a recurrent theme throughout our experiments; namely that the kernel methods such as kPCA and kRidge's features tend to clash with the features learned by NEU.  At times, they harmonize and the Non-Euclidean Updgraded kernel model offers astounding performance, however, at other times the performance deteriorates.  This unstable behaviour is not observed in the other non-Euclidean upgraded methods and this is because the other methods either do not impose any additional features (such as OLS, PCA, or Elastic Net) or are flexible enough to blend their feature representation with NEU's (as for GBRF, AE, or DNN).  
	\end{remark}
	
	\subsection{Simulated Experiments}\label{ss_sim_IUP}
	Next, we unpack and understand the detailed behaviour of NEU in the controlled environment offered by simulation studies.  We consider a series of regression problems.  In each situation, the data is generated using to the non-linear regression model with additive and multiplicative noises
	\begin{equation}
	y = U_{\delta}m(x)+\sigma Z,
	\label{eq_lornius_regolius}
	\end{equation}
	where $Z\sim N(0,1)$ and $U_{\delta}\sim U(1-\delta,1+\delta)$, $0<\delta <1$, $\sigma>0$, and $m$ is a non-linear function.  The multiplicative noise $U_{\delta}$ encapsulates model misspecification as it discontinuously (in $x$) distorts the shape of the unknown function $m$, and the additive noise $\sigma_{\epsilon}Z$ quantifies the noise distorting the signal, as in classical regression problem formulations.  
	
	We consider four challenging non-linear functions, each exhibiting a distinct pathology.  The first, is comprised of several distinct local sub-patterns.  The second exhibits aperiodic oscillations.  The third, is split by a sharp jump discontinuity.  The last pattern is highly discontinuous and we use it to evaluate each model's ability to discern between a sharp irregular signal and varying levels of noise.  
	
	The NEU-OLS and NEU-DNN models will be benchmarked against three standard non-parametric regression algorithms, penalized smoothing splines regression ($p$-splines), locally weighted scatterplot smoothing (LOESS), kernel ridge regression (Ker-Ridge), and feed-forward artificial neural networks (DNN).  Other than the DNNs which were discussed thoroughly in the paper's introductory section, we review the benchmark models here.

\begin{figure}[H]
		\centering
		\includegraphics[width=0.5\textwidth]{%
			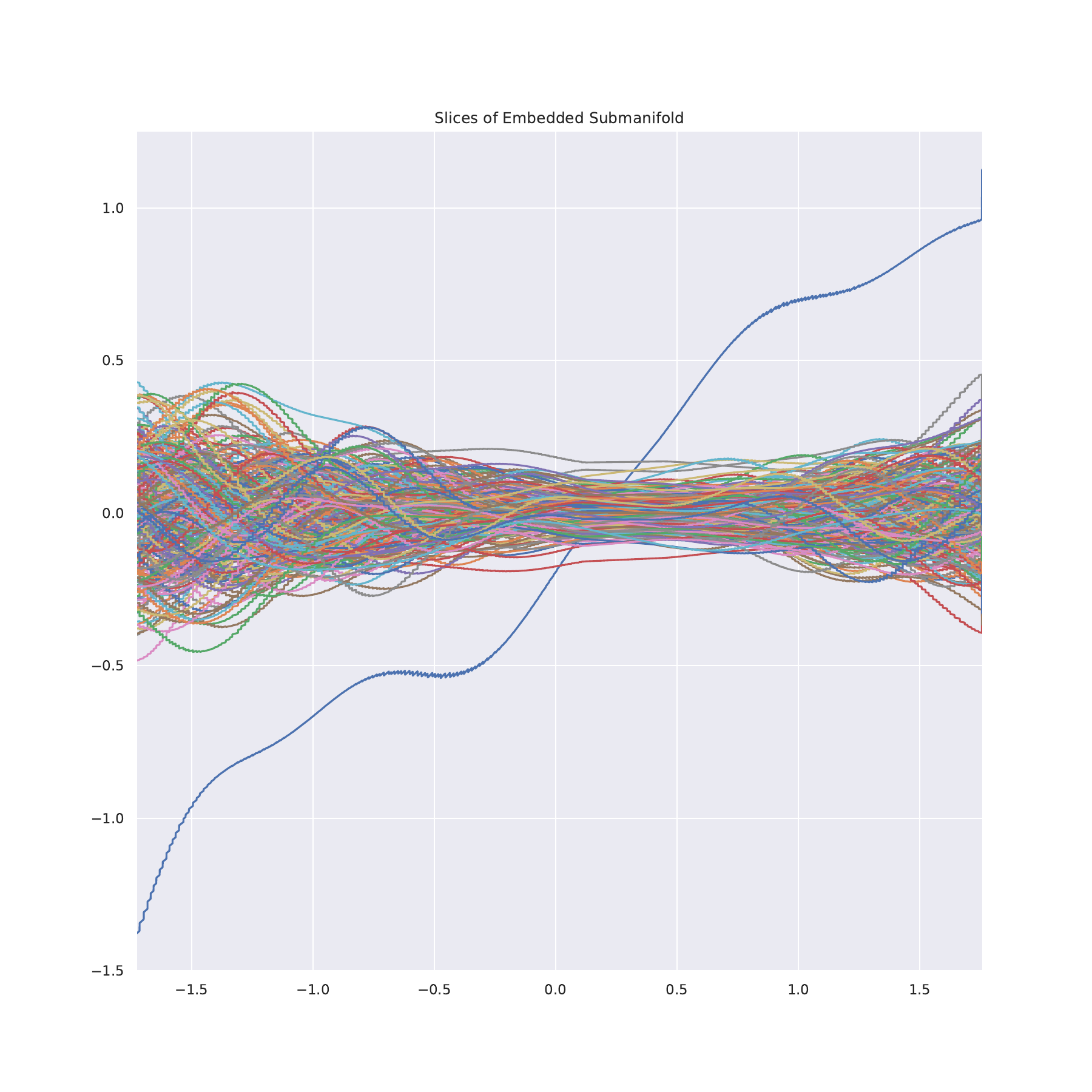}
		\caption{NEU's features for: $min(e^{\frac{-1}{(1+x)^2}},x+cos(x))$.  }
		\label{fig_plot_funny_function_osculatory}
	\end{figure}

\begin{figure}[H]
		\centering
		\includegraphics[width=.5\textwidth]{%
			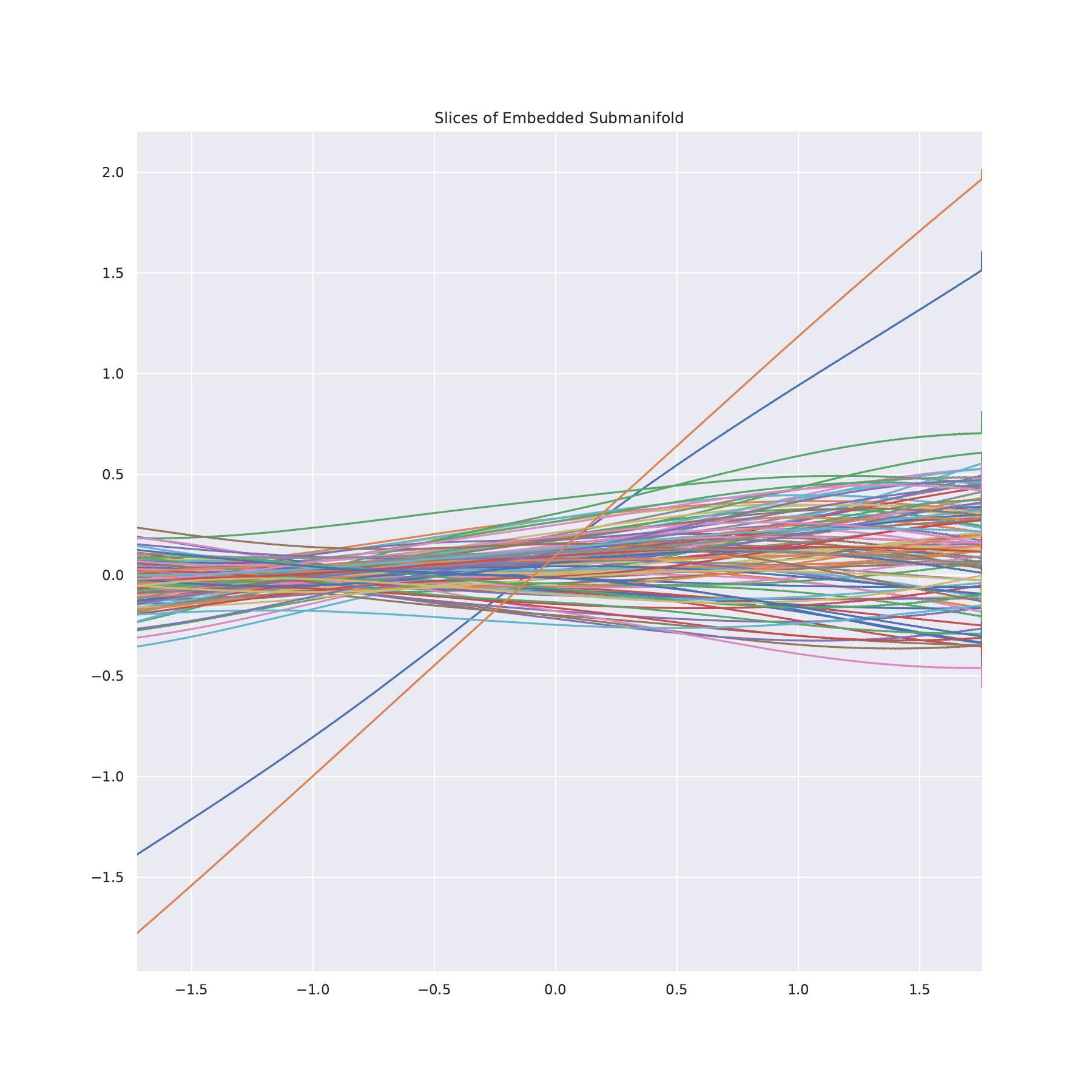}
		\caption{NEU's features for: $min(e^{\frac{-1}{(1+x)^2}},x+cos(x))$.}
		\label{fig_plot_funny_function_1}
	\end{figure}

	In each of our experiments, we visualize the feature representation learned by NEU by plotting each of the coordinates of $\phi(x)$.  	These plots are given in Figures~\ref{fig_plot_funny_function_1}, \ref{fig_plot_funny_function_osculatory}, and ~\ref{fig_plot_jump_disc}, respectively for each experiment.  	Essentially, these can be interpreted as the \textit{features} learned by NEU, which are then fed into the upgraded model.  In particular, when the model is linear, the target function is approximately expressible as a linear combination of these features.  
	
	We see that the target function is reflected by each of the feature maps learned by NEU.  For example, in the first implementation, NEU's feature representation illustrated in Figure~\ref{fig_plot_funny_function_1} has a dramatic change at the precise point where the two sub-patterns deviate from one another.  In the second experiment, NEU's produces a feature map, illustrated by Figure \ref{fig_plot_funny_function_osculatory}, whose coordinates represent osculations happening at different rates; these are then combined by the linear model being upgraded to produce the correct osculating pattern.  In the final experiment, NEU's features are illustrated in Figure~\ref{fig_plot_jump_disc}, and draw out two distinct and relatively flat heaps.  This reflects the sharp discontinuity separating the two otherwise constant parts of the target function.

	For each simulation, $10^4$ observations are generated on the interval $[-1,1]$; the data is then normalized to the unit square for uniformity between the three examples.  The models' tuning-parameters are then estimated by cross-validation.

	\subsubsection{Aperiodic Oscillations}\label{sss_Aperiodic_oscul}
	We begin by evaluating each model's ability to handle aperiodic oscillations.  To this end, we simulate from the unknown function $
	m_{2}(x)=cos(e^{2+x})
	.
	$  
	
	\begin{table}[H]
		\centering
		\begin{tabular}{lrrrrr}
			\toprule
			{Test} &   Er. 95L &      Er. Mean &   Er. 95U &       MAE &       MSE \\
			\midrule
			NEU-OLS            & -0.021313 & -2.570136e-09 &  0.020961 &  0.435468 &  0.292991 \\
			Smoothing Splines   & -0.006834 &  1.428117e-02 &  0.035403 &  0.435920 &  0.295227 \\
			LOESS              & -0.012301 &  1.862346e-02 &  0.048980 &  0.630788 &  0.613180 \\
			ENET        & -0.034831 & -1.136868e-17 &  0.035714 &  0.740855 &  0.805618 \\
			NEU-GBRF           & -0.023963 & -5.684342e-17 &  0.024021 &  0.494789 &  0.382412 \\
			GBRF               & -0.024555 & -5.684342e-18 &  0.024455 &  0.499496 &  0.387694 \\
			NEU-kRidge       & -0.020963 & -2.570866e-04 &  0.020747 &  0.429108 &  0.284365 \\
			kRidge             & -0.020745 & -3.927500e-05 &  0.020682 &  0.434514 &  0.291221 \\
			NEU-DNN &  0.004625 &  2.622699e-02 &  0.047207 &  0.435748 &  0.296543 \\
			DNN               & -0.021901 & -1.453060e-04 &  0.021675 &  0.441955 &  0.304498 \\
			\bottomrule
		\end{tabular}
		\caption{A-Periodic oscillations - $
			m_{2}(x)=cos(e^{2+x})
			$ : $\sigma=\delta=0.5$}
		\label{fig_test_oscul_noisy}
	\end{table}
	
	Figure~\ref{fig_Error_dists_OScul_2} highlights the clash between the rigid structure imposed by the Kernel regression's implicit feature map and NEU's feature map.  Since NEU's feature map is designed for models that are either linear or can efficiently approximate linear maps, then kernel regression's feature map can, and in this case, it does, interfere with the representation learned by NEU.  However, as is also reflected in Table~\ref{Test_non_local_train_noisy}, this only happens with the Kernel regression method and not with the GBRF, linear regression, or DNN methods.

	\begin{table}
		\centering
		\begin{tabular}{lrrrrr}
			\toprule
			{Train} &   Er. 95L &  Er. Mean &   Er. 95U &       MAE &       MSE \\
			\midrule
			NEU-OLS            & -0.007318 & -0.005953 & -0.004600 &  0.042497 &  0.003580 \\
			Smoothing Splines   & -0.016907 & -0.006644 &  0.002417 &  0.070279 &  0.180917 \\
			LOESS              &  0.016649 &  0.029347 &  0.042376 &  0.467625 &  0.321612 \\
			ENET        & -0.007673 &  0.009254 &  0.025702 &  0.656024 &  0.526628 \\
			NEU-GBRF           &  0.000899 &  0.008715 &  0.016585 &  0.308246 &  0.120689 \\
			GBRF               & -0.004417 &  0.003626 &  0.011627 &  0.313712 &  0.126322 \\
			NEU-kRidge       & -0.008430 & -0.006731 & -0.005058 &  0.056449 &  0.005649 \\
			kRidge             & -0.007045 & -0.005678 & -0.004326 &  0.046354 &  0.003695 \\
			DNN               & -0.011299 & -0.008462 & -0.005682 &  0.102038 &  0.015454 \\
			NEU-DNN &  0.016808 &  0.018696 &  0.020518 &  0.055287 &  0.007063 \\
			\bottomrule
		\end{tabular}
		\caption{A-Periodic Oscillations - $
			m_{2}(x)=cos(e^{2+x})
			$ : $\sigma=\delta=0.5$}
		\label{fig_trian_oscul_noisy}
	\end{table}

	\begin{figure}
		\centering
		\begin{subfigure}[b]{.5\textwidth}
			\includegraphics[width=1\textwidth]{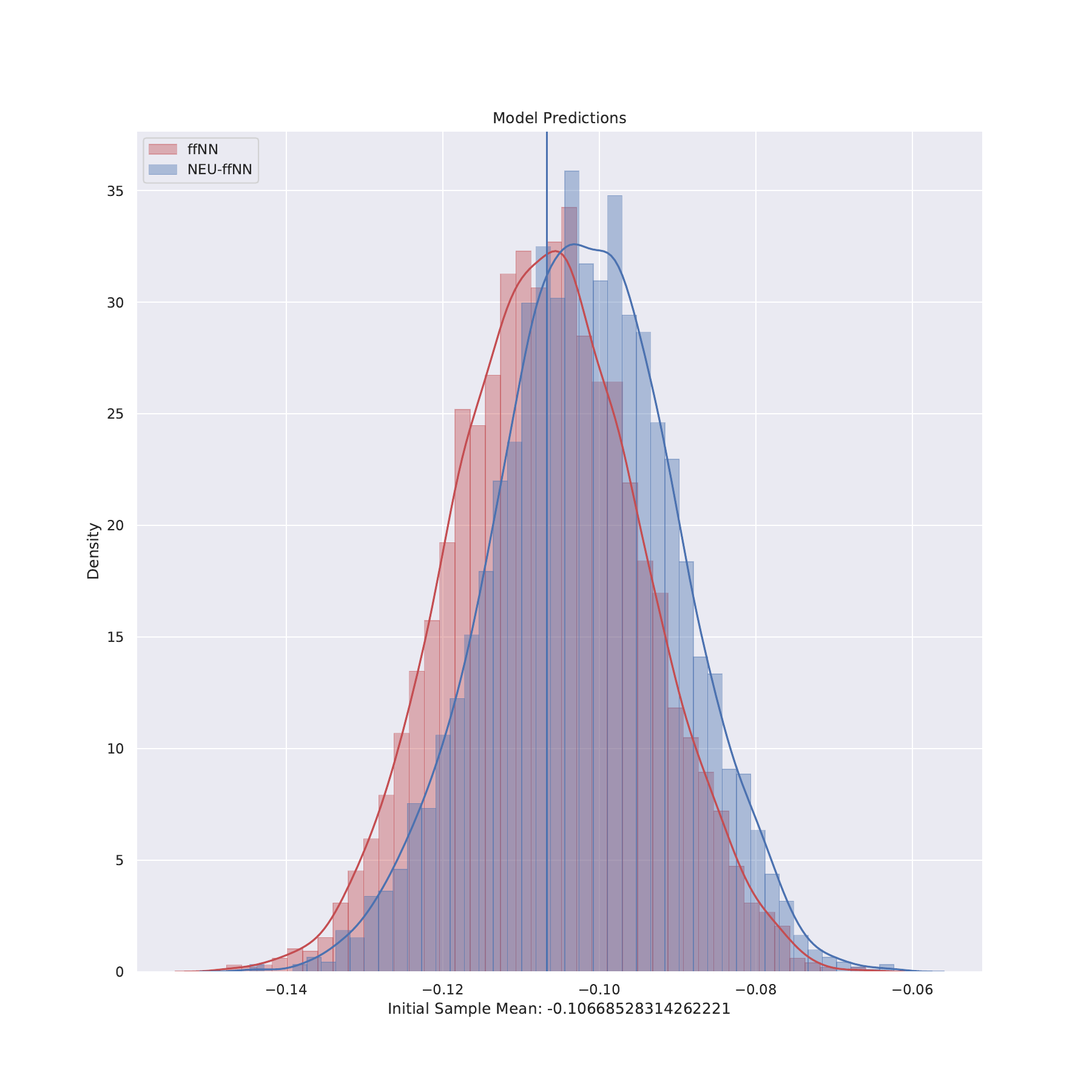}
			\caption{NEU-DNN vs. DNN}
		\end{subfigure}%
		\begin{subfigure}[b]{.5\textwidth}
			\includegraphics[width=1\textwidth]{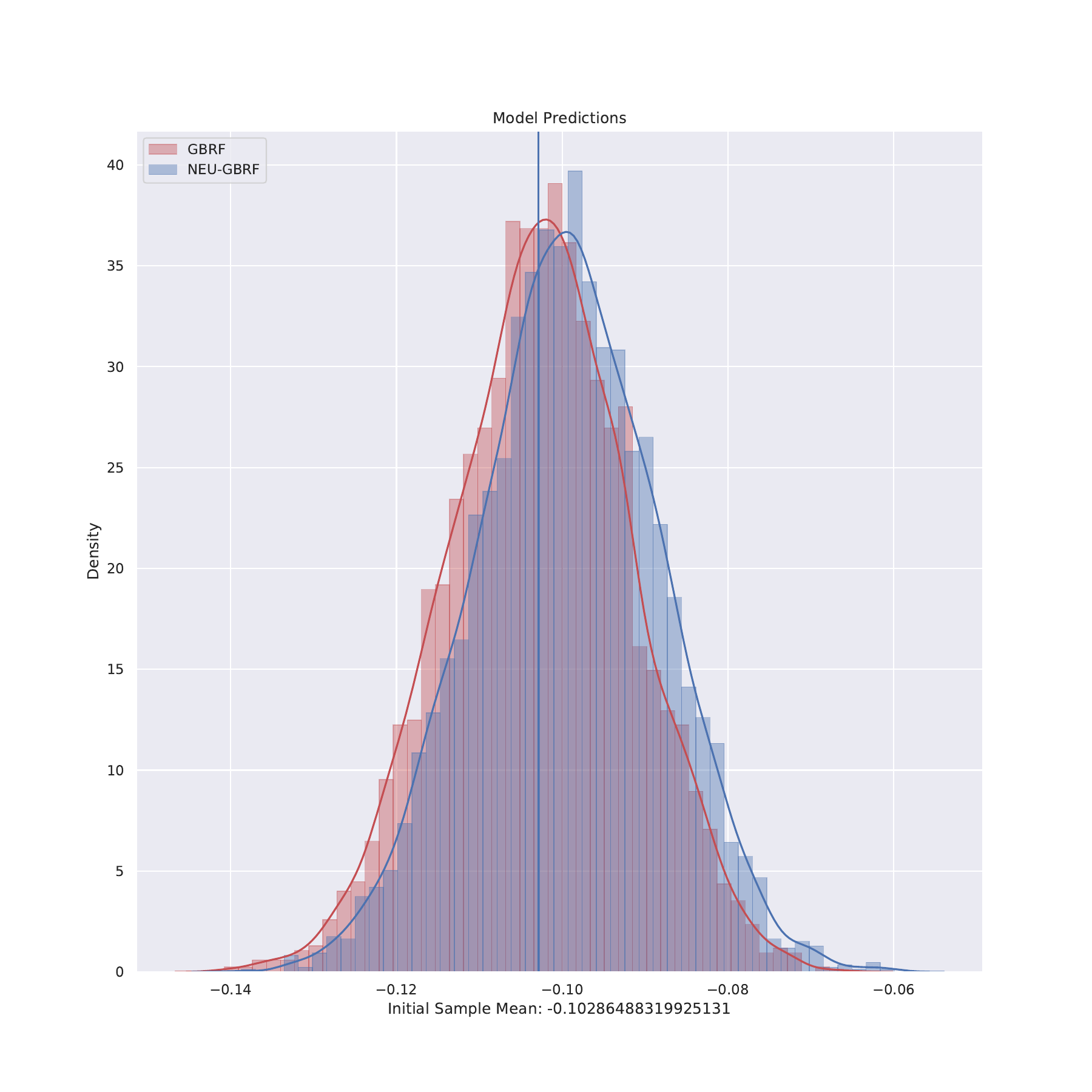}
			\caption{NEU-GBRF vs. GBRF}
		\end{subfigure}%
		
		\begin{subfigure}[b]{.5\textwidth}
			\includegraphics[width=1\textwidth]{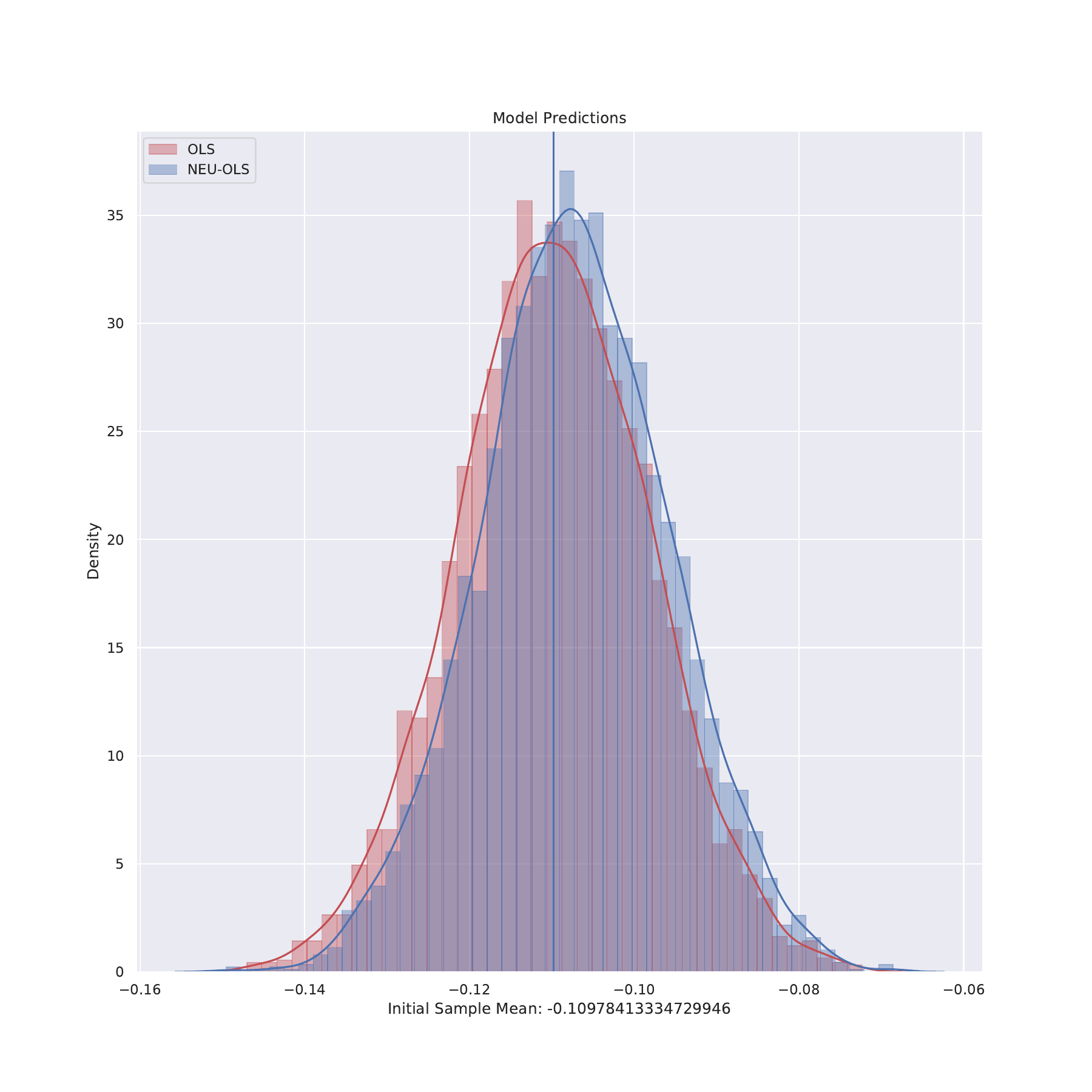}
			\caption{NEU-OLS vs. OLS}
		\end{subfigure}%
		\begin{subfigure}[b]{.5\textwidth}
			\includegraphics[width=1\textwidth]{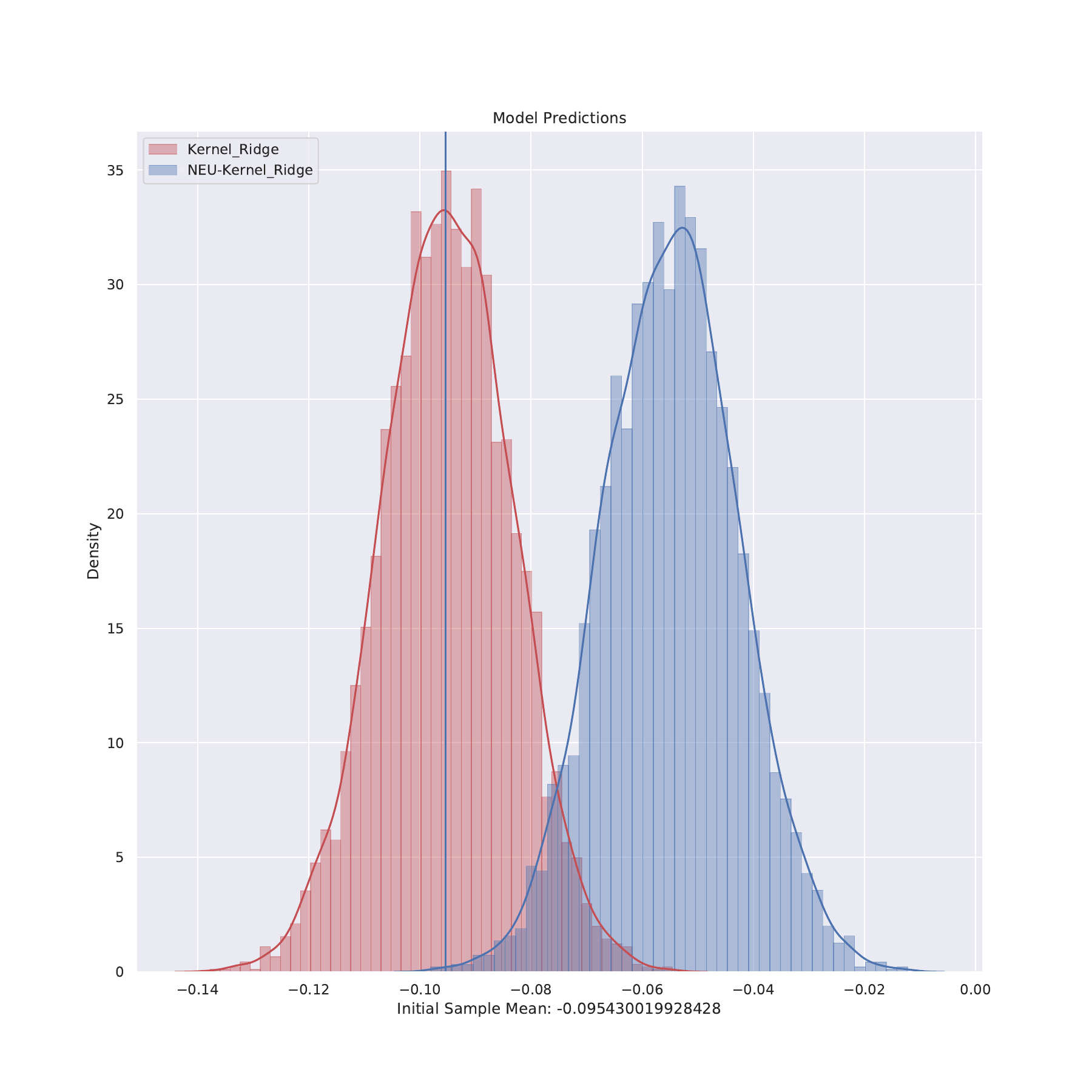}
			\caption{NEU-kRidge vs. Kernel}
		\end{subfigure}%
		\caption{Error Distribution Comparisons for:  $\cos(e^{2+x})$.  }
		\label{fig_Error_dists_Aperiodic_2}
	\end{figure}
	
	\subsubsection{Functions with Local Behaviour}\label{sss_Implementations_local}
	Next, we compare each model's abilities to learn from functions determined by several, exclusively local, sub-patterns.  Thus, the unknown function $m$ of~\eqref{eq_lornius_regolius} is taken to be $m(x)\triangleq \min(e^{-\frac1{(x+1)^2}},x+\cos(x))$.  The underlying pattern is therefore generated from two distinct sub-patterns $e^{-\frac{1}{(x+1)^2}}$ and $x+\cos(x)$, with the change between the two occurring every time the condition 
	$e^{-\frac{1}{(x+1)^2}}<x+\cos(x)$ either holds or fails.  
	
	\begin{table}[H]
		\centering
		\begin{tabular}{lrrrrr}
			\toprule
			{Test} &   Er. 95L &      Er. Mean &   Er. 95U &       MAE &       MSE \\
			\midrule
			NEU-OLS            & -0.019358 &  6.980434e-04 &  0.021225 &  0.418056 &  0.266197 \\
			Smoothing Splines   & -0.020310 &  4.915925e-11 &  0.019642 &  0.416499 &  0.265408 \\
			LOESS              & -0.017103 &  3.094336e-03 &  0.023561 &  0.418556 &  0.268401 \\
			ENET                 & -0.020520 &  4.760636e-17 &  0.020322 &  0.419717 &  0.269394 \\
			NEU-GBRF           & -0.020503 & -1.705303e-16 &  0.020621 &  0.430601 &  0.284261 \\
			GBRF               & -0.020795 &  7.958079e-17 &  0.020671 &  0.432727 &  0.286976 \\
			NEU-kRidge       & -0.020580 &  1.588384e-05 &  0.020460 &  0.418158 &  0.267961 \\
			kRidge             & -0.019606 &  1.823117e-06 &  0.020202 &  0.418136 &  0.267707 \\
			NEU-DNN          &  0.001613 &  2.200638e-02 &  0.041955 &  0.418093 &  0.269347 \\
			DNN               & -0.012107 &  8.756267e-03 &  0.029376 &  0.418196 &  0.268941 \\
			\bottomrule
		\end{tabular}
		\caption{Non-Local function - $min(e^{\frac{-1}{(1+x)^2}},x+cos(x))$ : $\delta=\sigma=0.5$.}
		\label{Test_non_local_train_noisy}
	\end{table}
	
	Figure~\ref{Test_non_local_train_noisy} show that NEU-OLS and NEU-DNN still offer the best out-of-sample performance amongst the DNN, LOESS, ENET, GBRF, kRidge, and DNN Models.  However, this implementation suggests that smoothing splines may are better suited to locally-determined target functions.  This is not surprising since NEU performs any localization after representing the pattern in a higher-dimensional space, whereas smoothing splines can locally approximate any function directly.

	\begin{table}
		\centering
		\begin{tabular}{lrrrrr}
			\toprule
			{Train} &   Er. 95L &  Er. Mean &   Er. 95U &       MAE &       MSE \\
			\midrule
			NEU-OLS            & -0.003219 & -0.001356 &  0.000763 &  0.029811 &  0.007868 \\
			Smoothing Splines   & -0.005625 & -0.004594 & -0.003564 &  0.035477 &  0.002062 \\
			LOESS              & -0.003280 & -0.001950 & -0.000587 &  0.037035 &  0.003622 \\
			ENET        & -0.007302 & -0.005885 & -0.004524 &  0.047878 &  0.003902 \\
			NEU-GBRF           & -0.008818 & -0.004893 & -0.000868 &  0.151281 &  0.031597 \\
			GBRF               & -0.008343 & -0.004443 & -0.000530 &  0.148070 &  0.029684 \\
			NEU-kRidge       & -0.005622 & -0.004717 & -0.003816 &  0.026037 &  0.001607 \\
			kRidge             & -0.004352 & -0.003566 & -0.002819 &  0.024096 &  0.001135 \\
			NEU-DNN             &  0.014463 &  0.015718 &  0.016956 &  0.040284 &  0.003335 \\
			DNN               &  0.001173 &  0.002462 &  0.003720 &  0.035320 &  0.003244 \\
			\bottomrule
		\end{tabular}
		\caption{Non-Local function - $min(e^{\frac{-1}{(1+x)^2}},x+cos(x))$ : $\delta=\sigma=0.5$.}
		\label{Train_non_local_train_noisy}
	\end{table}
	
	Figure~\ref{Train_non_local_train_noisy} shows that in-sample, NEU-OLS offers provides a better fit than the Smoothing Splines, LOESS, ENET, GBRF, and the DNN model.  However, kRidge seems best suited to the in-sample fitting of this type of pattern.  We also note that, though the UAP-invariance property (P-i) guarantees that NEU-DNN is universal, since DNN was, Figure~\ref{Train_non_local_train_noisy} shows that at-times NEU-OLS can fit finite data-sets better than NEU-DNN does.  This shows that though the DNN and the reconfiguration networks have arbitrarily large memory capacities, these two may at-times unfavorably interact since both memorize input-output pairs differently (which we see by comparing the proofs of Theorem~\ref{thrm_memory_capacity_w_guessing} and the central result of \cite{MemoryCapacityffNNs}).  This interaction effect is less likely with NEU-OLS than NEU-DNN, since the reconfiguration since the memorization only passes through one layer in the latter case.   
	
	\begin{figure}
		\centering
		\begin{subfigure}[b]{.5\textwidth}
			\includegraphics[width=1\textwidth]{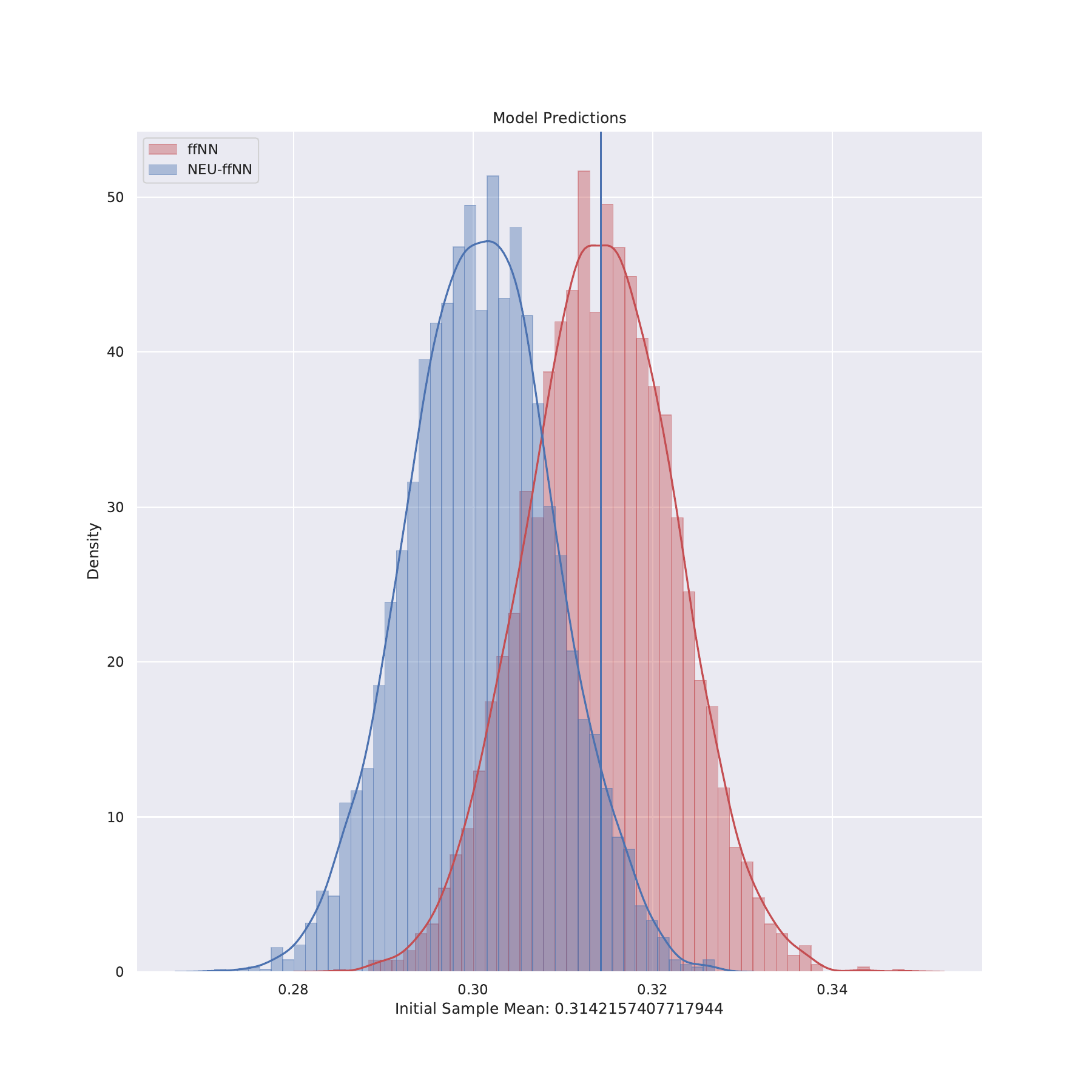}
			\caption{NEU-DNN vs. DNN}
		\end{subfigure}%
		\begin{subfigure}[b]{.5\textwidth}
			\includegraphics[width=1\textwidth]{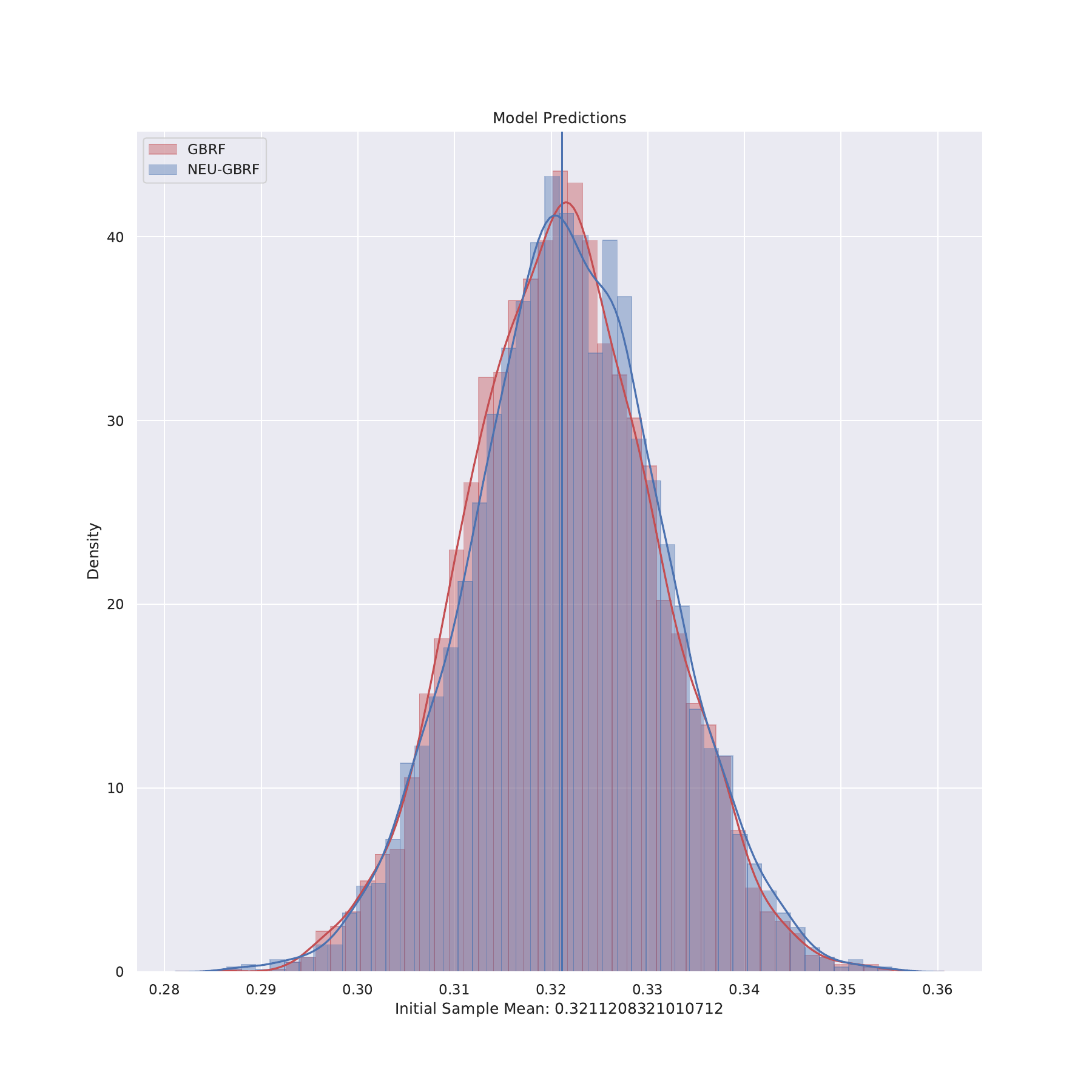}
			\caption{NEU-GBRF vs. GBRF}
		\end{subfigure}%
		
		\begin{subfigure}[b]{.5\textwidth}
			\includegraphics[width=1\textwidth]{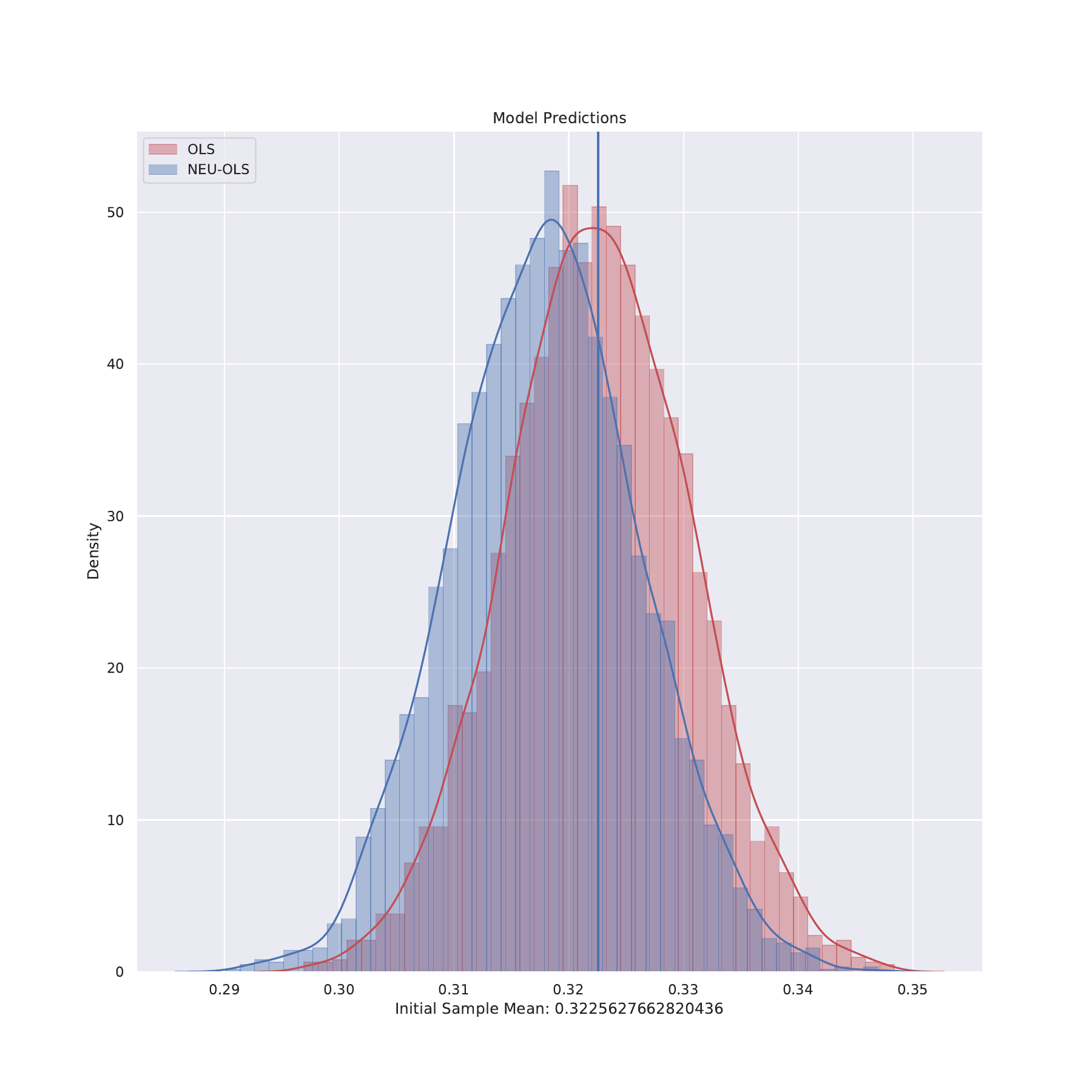}
			\caption{NEU-OLS vs. OLS}
		\end{subfigure}%
		\begin{subfigure}[b]{.5\textwidth}
			\includegraphics[width=1\textwidth]{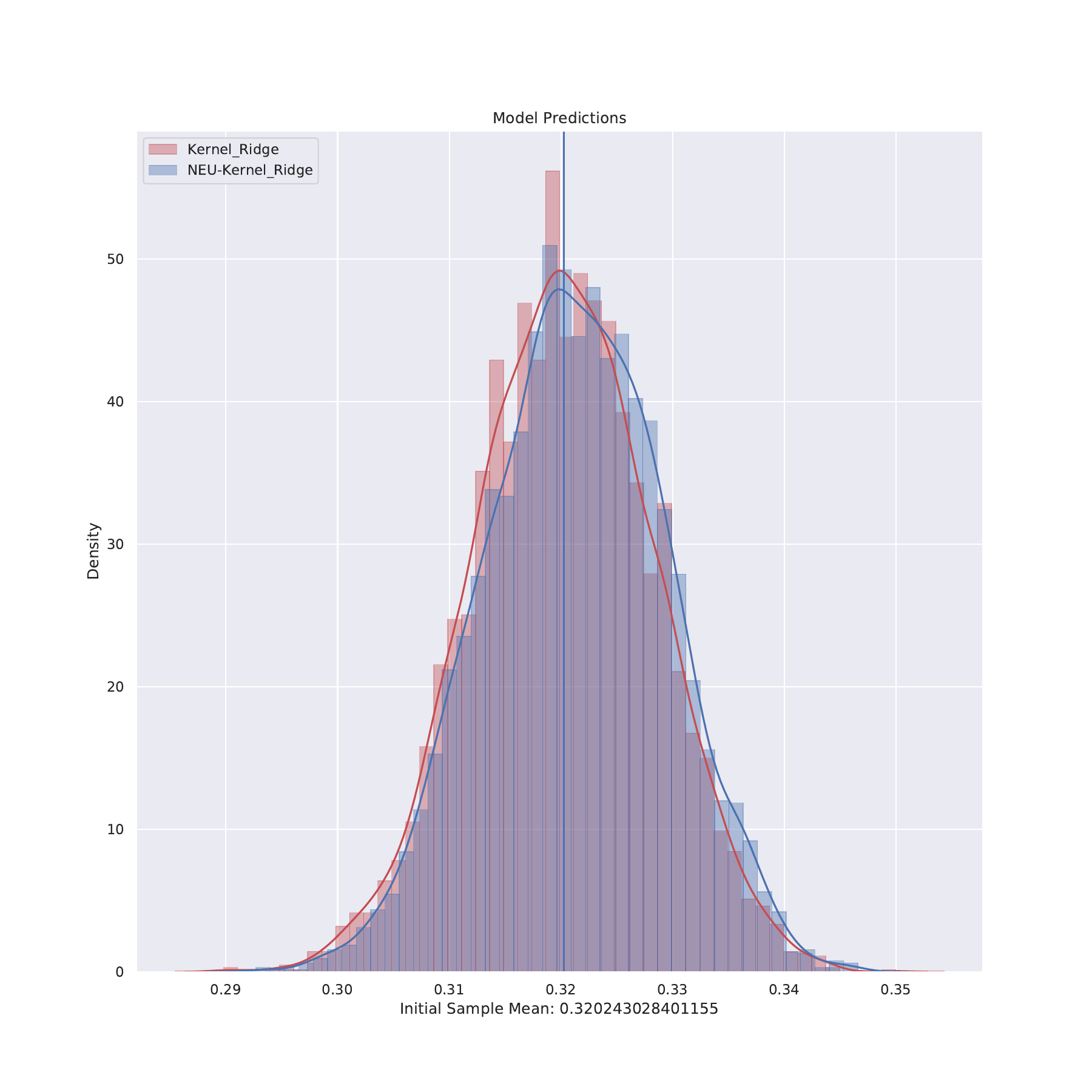}
			\caption{NEU-kRidge vs. Kernel}
		\end{subfigure}%
		\caption{Error Distribution Comparisons for:  $min(e^{\frac{-1}{(1+x)^2}},x+cos(x))$.  }
		\label{fig_Error_dists_OScul_2}
	\end{figure}
	
	Figure~\ref{fig_Error_dists_OScul_2} reflect the bias reduction obtained by NEU's feature map, which is most significant when applied to the OLS and DNN models.  Here, NEU showcases the benefit of it being able to only locally modify a pattern, which is especially important in this case since two unrelated local sub-patterns determine $ m$.

	Tables~\ref{Train_non_local_train_noisy} and~\ref{Test_non_local_train_noisy} show that the NEU models achieve an improved performance both in-sample and out-of sample over their classical variants.  The feature maps were only trained once for the OLS model and then used in the remaining models.  Thus, the non-linear feature presentation must be correct as it transfers its improvement to each of the benchmarked regression models.  
	We see that NEU-OLS offers the best accuracy, and most stable in-to-out of sample performance.  This is likely due to the learned linearizing feature map not having to conflict with any other assumed feature map, as is the case for NEU-kRidge, NEU-DNN, and NEU-GBRF.

	\subsubsection{Jump Discontinuities}\label{sss_implementations_jump_disc}
\begin{figure}[H]
		\centering
		\includegraphics[width=0.5\textwidth]{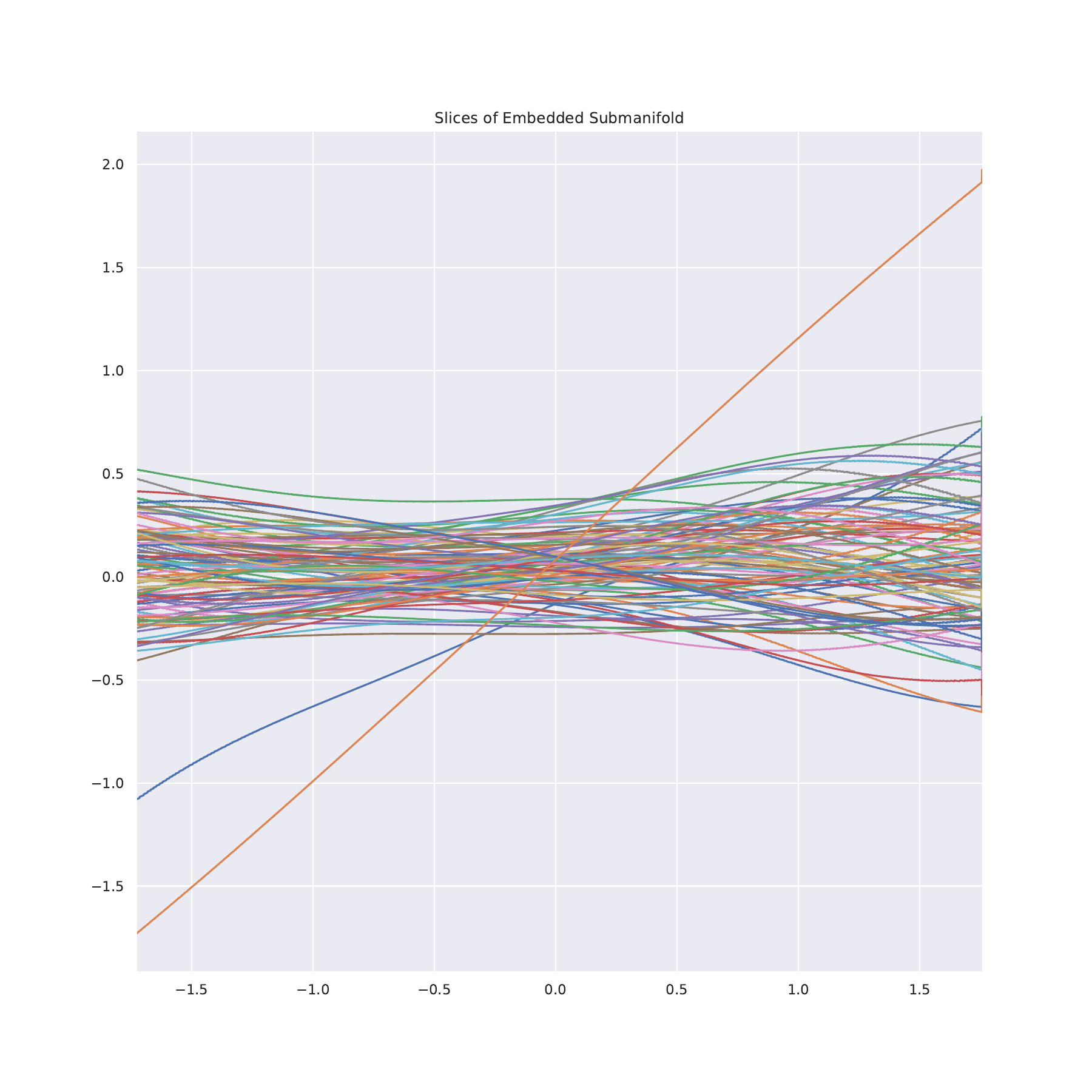}
		\caption{NEU's features for: $I_{(-\infty,.5]}$.  }
		\label{fig_plot_jump_disc}
	\end{figure}

	The last simulation experiment explores a situation with discontinuities that is outside the scope of the standard p-sline, LOESS, and DNN methods. However, the NEU-OLS is able to perform well when the data exhibits these jump discontinuities. The function $m$ in equation~\eqref{eq_lornius_regolius} is assumed to be
	$
	m(x)=I_{(-\infty,\frac1{2}]}(x).
	$
	As reflected by Figure~\ref{fig_plot_jump_disc}, NEU behaves the same when the unknown noisy function being approximated has a jump discontinuity as when distinct, locally-determined functions determined it; as in Figure~\ref{fig_Error_dists_OScul_2}.  In both situations, NEU can learn a feature map that is determined only by local data, so it implicitly separates the behavior of the model on each part of the jump discontinuity, just as it did on the different sub-patterns in Figure~\ref{fig_Error_dists_OScul_2}.

	\begin{table}[H]
		\centering
		\begin{tabular}{lrrrrr}
			\toprule
			{Test} &   Er. 95L &      Er. Mean &   Er. 95U &       MAE &       MSE \\
			\midrule
			NEU-OLS            & -0.007420 & -1.978159e-03 &  0.003309 &  0.219500 &  0.075624 \\
			Smoothing Splines   & -0.005347 & -1.902232e-12 &  0.005499 &  0.221883 &  0.077588 \\
			LOESS              & -0.003747 &  2.843149e-03 &  0.009276 &  0.261553 &  0.109594 \\
			ENET        & -0.007296 & -5.968559e-17 &  0.007283 &  0.293706 &  0.134424 \\
			NEU-GBRF           & -0.007103 &  3.467449e-16 &  0.007091 &  0.292915 &  0.128904 \\
			GBRF               & -0.006864 &  3.240075e-16 &  0.007085 &  0.292915 &  0.128904 \\
			NEU-kRidge       & -0.004780 &  4.746673e-04 &  0.005816 &  0.217793 &  0.074313 \\
			kRidge             & -0.005810 &  1.327000e-06 &  0.005720 &  0.237115 &  0.088674 \\
			NEU-DNN & -0.017917 & -1.253940e-02 & -0.007353 &  0.218370 &  0.074929 \\
			DNN               &  0.004379 &  9.808546e-03 &  0.015200 &  0.218675 &  0.075156 \\
			\bottomrule
		\end{tabular}
		\caption{Discontinuous Function - $
			m(x)=I_{(-\infty,\frac1{2}]}(x).
			$ : $\delta=\epsilon=0.5$}
	\end{table}
	
	The presented simulations studies, highlight the strengths and weaknesses of NEU.  In nearly every case, the Non-Euclidean Upgraded model outperforms is classical counterpart. However, we typically find that the more linear the original model, the more reliable NEU's performance will be.  This is likely due to conflicts between the assumed feature map, especially in NEU-kRidge, and the feature map being learned.  This is because, assuming a feature map forces NEU to simultaneously learn the target while undoing any feature map misspecification.  
	
	\begin{table}
		\centering
		\begin{tabular}{lrrrrr}
			\toprule
			{Train} &   Er. 95L &  Er. Mean &   Er. 95U &       MAE &       MSE \\
			\midrule
			NEU-OLS            &  0.002959 &  0.003741 &  0.004552 &  0.011879 &  0.001689 \\
			Smoothing Splines   &  0.001218 &  0.002504 &  0.003796 &  0.022726 &  0.004295 \\
			LOESS              &  0.005587 &  0.009368 &  0.013222 &  0.135893 &  0.038387 \\
			ENET        &  0.002787 &  0.007589 &  0.012474 &  0.210154 &  0.063623 \\
			NEU-kRidge       &  0.001912 &  0.002406 &  0.002911 &  0.011165 &  0.000659 \\
			kRidge             & -0.003193 & -0.000655 &  0.001772 &  0.098367 &  0.015523 \\
			NEU-GBRF           & -0.001760 &  0.002934 &  0.007482 &  0.238186 &  0.056756 \\
			GBRF               & -0.001773 &  0.002934 &  0.007643 &  0.238186 &  0.056756 \\
			NEU-DNN & -0.010775 & -0.010128 & -0.009464 &  0.021392 &  0.001207 \\
			DNN               &  0.011372 &  0.012157 &  0.012948 &  0.019651 &  0.001737 \\
			\bottomrule
		\end{tabular}
		\caption{Discontinuous Function - $
			m(x)=I_{(-\infty,\frac1{2}]}(x).
			$ : $\delta=\epsilon=0.5$.}
	\end{table}
	
	\begin{figure}
		\centering
		\begin{subfigure}[b]{.5\textwidth}
			\includegraphics[width=1\textwidth]{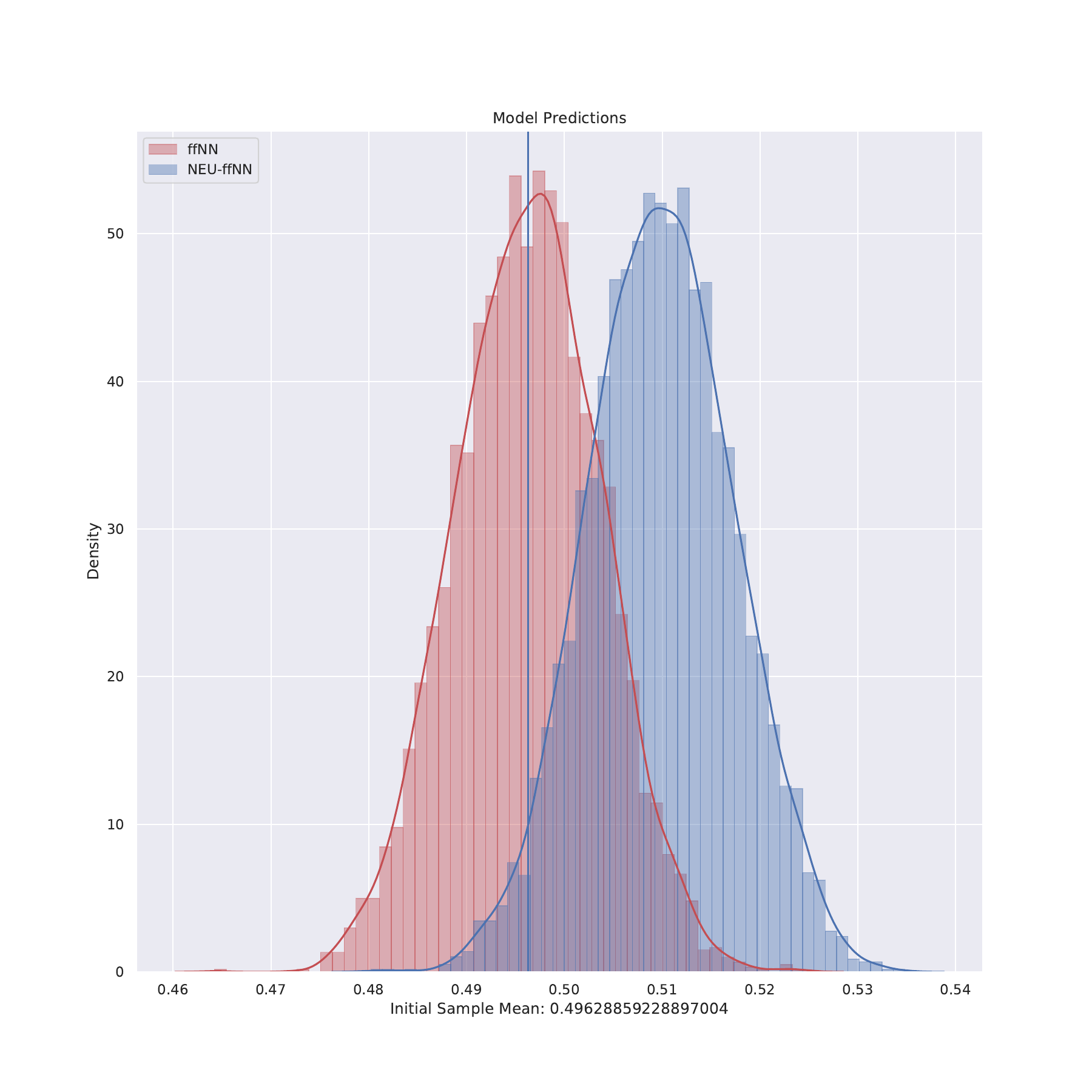}
			\caption{NEU-DNN vs. DNN}
		\end{subfigure}%
		\begin{subfigure}[b]{.5\textwidth}
			\includegraphics[width=1\textwidth]{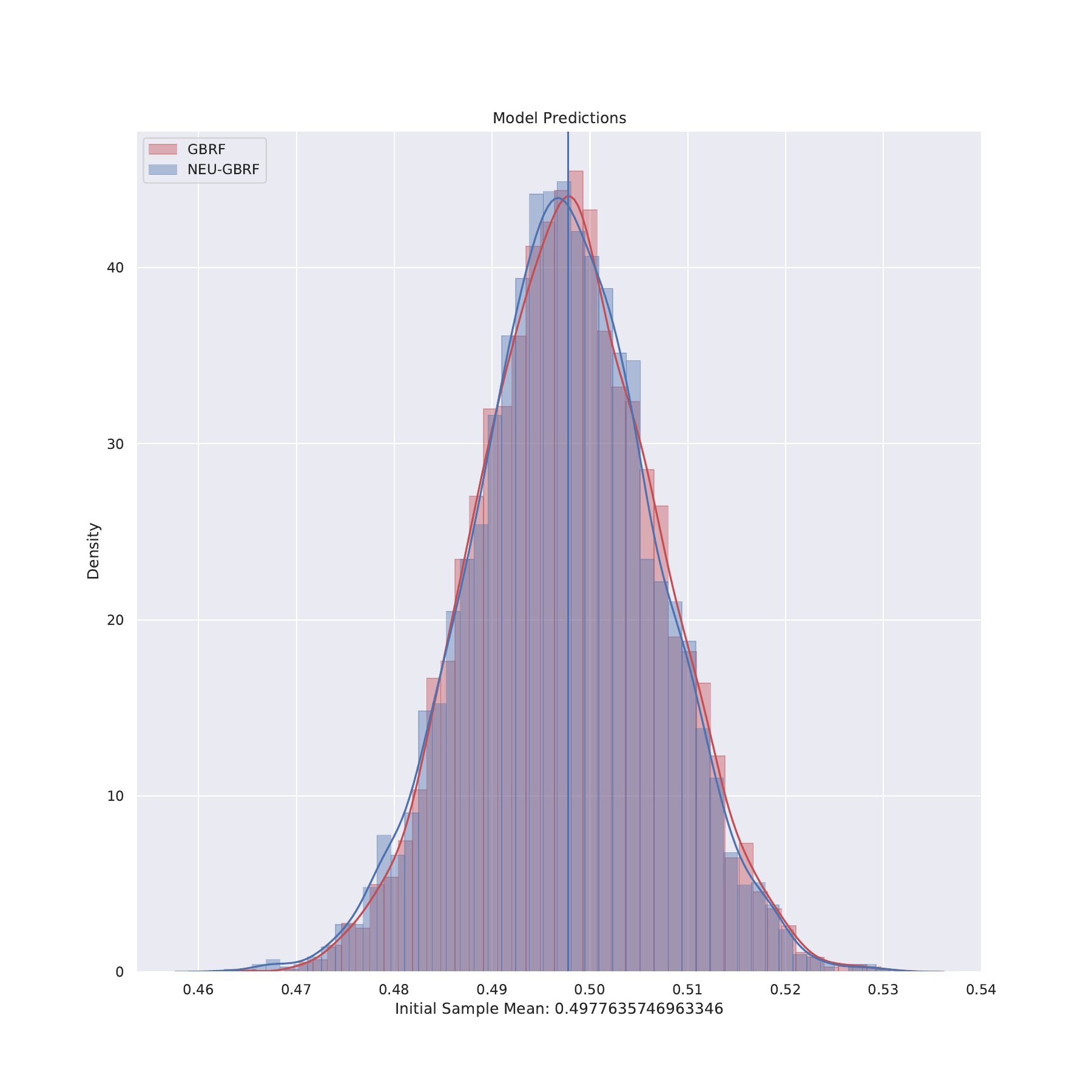}
			\caption{NEU-GBRF vs. GBRF}
		\end{subfigure}%
		
		\begin{subfigure}[b]{.5\textwidth}
			\includegraphics[width=1\textwidth]{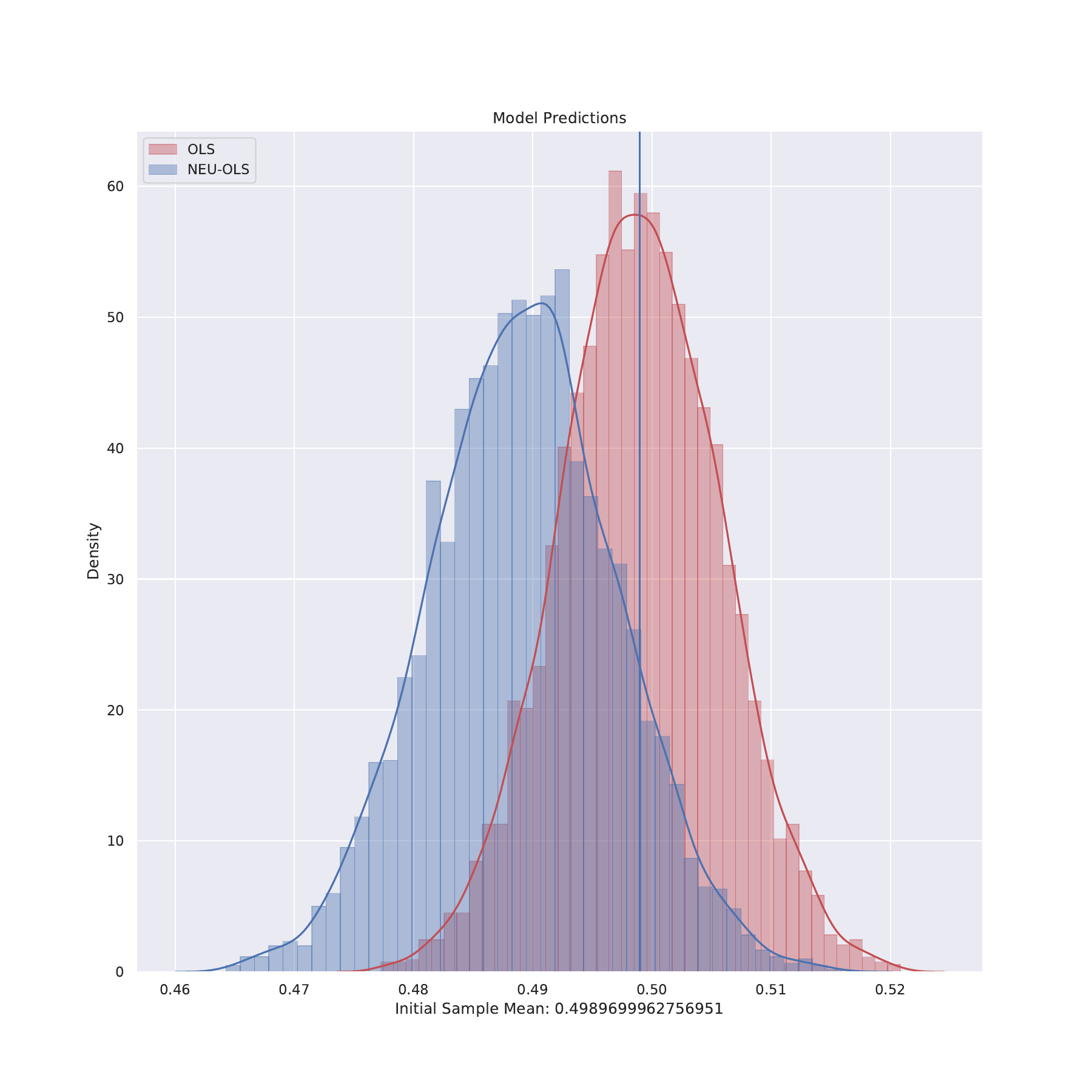}
			\caption{NEU-OLS vs. OLS}
		\end{subfigure}%
		\begin{subfigure}[b]{.5\textwidth}
			\includegraphics[width=1\textwidth]{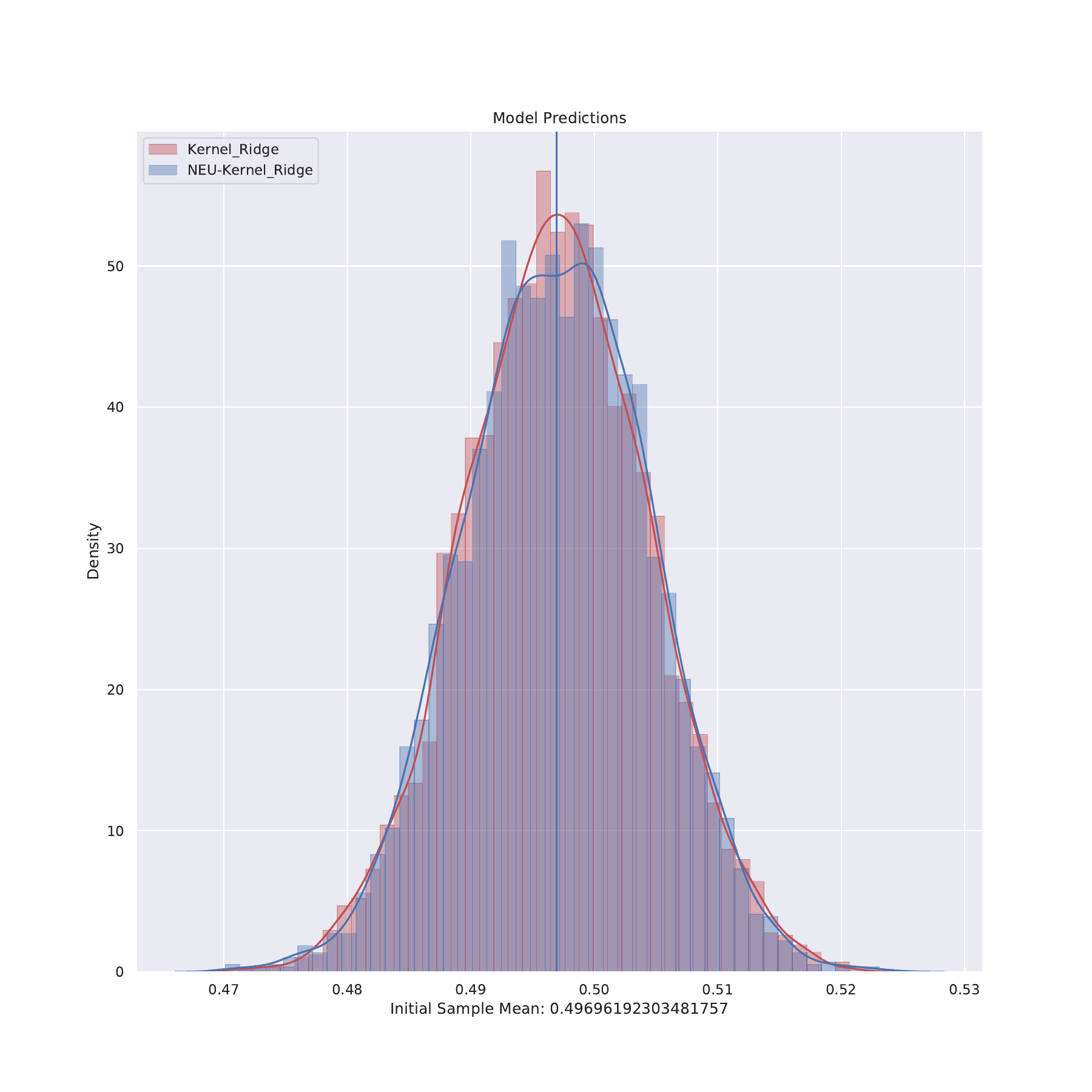}
			\caption{NEU-kRidge vs. Kernel}
		\end{subfigure}%
		\caption{Error Distribution Comparisons for:  $min(e^{\frac{-1}{(1+x)^2}},x+cos(x))$.  }
		\label{fig_Error_dists_jump_2}
	\end{figure}

	Each of the simulated pathological regression problems, illustrated in this section showed that NEU is not only capable of out-performing each of the benchmark models regardless of how badly behaved the unknown target function is. Moreover, this performance improvement was maintained in the face of high amounts of multiplicative and additive noise.  These results mirror our theoretical findings and support the hypothesis that the performance improvement observed by using NEU in the context of quantitative finance, was not a singular instance but rather part of a general theoretically-founded trend.  
	\section{Conclusion}\label{sec:conclude}
	This paper introduced the first generic algorithmic procedure for learning any feature map with the invariance properties (P-i) and (P-ii) while simultaneously guaranteeing the performance enhancement of properties (P-iii) and (P-v).  From the perspective of Kernel methods, NEU is a universal procedure for learning a low-dimensional generic feature map with many desirable properties.  From the standpoint of geometric deep learning, NEU also provides an answer to the recent research problem initiated in \cite{kratsios2020non} and in \cite{kratsios2021quantitative}, of how to generically learn optimal UAP-invariant feature maps from $\rrd$ in cases where a UAP-preserving feature map is not already explicitly provided.
	
	From the manifold learning perspective, reconfiguration networks are the first provably universal and computationally tractable class of topological embeddings.  As a meta-algorithm, NEU can generically learn the optimal linearizing pre-processing step to nearly any model class $\fff$, provided that $\fff$ at-least contains all linear maps.  NEU's theoretical properties were also supported experimentally.  
	
	NEU successfully introduced tools from geometric deep learning into financial data-analysis.  NEU was found to outperform the current leading machine learning methods for non-parametric dimension reduction of yield curves and produce a competitive performance in the non-parametric stock-returns replication problem.

	\acks{This research was supported by the ETH Z\"{u}rich Foundation and the Natural Sciences and Engineering Research Council of Canada (NSERC).  The authors thank Alina Stancu (Concordia University) for helpful discussions, Josef Teichmann and the entire working group at ETH for their helpful feedback, and Behnoosh Zamanlooy for the Python guidance.  }

	\bibliography{Updated_References}
	\printindex

	\section{Proofs}\label{ap:proofs}
	This appendix contains proofs of the paper's main results and some auxiliary technical lemmas.  We draw the reader's attention to the fact that many of the paper's results are heavily interdependent and that this sequential dependence is different from the order giving the cleanest presentation, which we chose for the paper's main body.  Accordingly, proofs will be presented in their logical order, even if it differs from the paper's main expos\'{e}.  
	
	
	\subsection{Technical Lemmas}\label{lem_tech_lemmas}
	This section contains some technical lemmas which we often refer to throughout the paper's proofs.  
	\begin{lemma}[Properties of Reconfiguration Units/Network]\label{lem_reconfig_is_a_reconfig}
		Every reconfiguration unit is a reconfiguration network.  Moreover, the following hold:
		\begin{enumerate}[(i)]
			\item Every $\phi \in \Phi_{\star:d}$ is a homeomorphism in $\hhh(\rrd)$,
			\item For every $\phi \in \Phi_{\star:d}$, there exist $(\phi_{\alpha})_{\alpha \in [0,1]}\subset \hhh(\rrd)$ such that 
			for each $x \in \rrd$:
			\begin{equation}
			\begin{aligned}
			&\phi_0(x)=x, \qquad
			& \xi_{\alpha} \in \Phi_{\star:d},
			\qquad & \phi_1(x) = \phi(x)
			,
			\end{aligned}
			\label{defn_reconfiguration_isotopy_condition}
			\end{equation}
			for every $\alpha \in [0,1]$.  
			Moreover, for every $x \in \rrd$, the map $\alpha \mapsto \phi_{\alpha}(x)$ is continuous.  
			\item For every $x,y \in \rrd$ and every $\epsilon>0$ there exists a reconfiguration \textit{unit} $\phi$ such that
			$$
			\phi(x)= y \mbox{ and } \phi(z)=z
			,
			$$
			for every $z \in \rrd$ satisfying $\|x-z\|>\epsilon+\|x-y\|$,
			\item If $\phi_1,\phi_2 \in \Phi_{\star:d}$, then $\phi_2\circ \phi_1\in \Phi_{\star:d}$.  
		\end{enumerate}
	\end{lemma}
	\begin{proof}[{Proof of Lemma~\ref{lem_reconfig_is_a_reconfig}}]
		First, we observe that (iv) holds by construction and the fact that any two reconfiguration units are composable since they map to and from $\rrd$.

		By (iv) and the fact that the composition of homeomorphisms is again a homeomorphism, then it is enough to establish (ii) on a single map of the form
		\begin{equation}
		\sigma_{\alpha_n}\left(A(x)(x-c) \right) +b
		\label{eq_proof_of_prop1}
		.
		\end{equation}
		First, observe that for every $\alpha \in [0,1]$, the map $x\mapsto \sigma_{\alpha}(x)$ is monotonically increasing, continuous, and surjective; thus, by \citep{hoffmann2015continuity} implies that $\sigma_{\alpha}$ is a homeomorphism from $\rr$ onto itself.  Since the $d$-fold Cartesian products of homeomorphisms is again a homeomorphism, then the map $x\mapsto \sigma_{\alpha}\bullet x$ is a homeomorphism from $\rrd$ onto itself.  Moreover, for any $b \in \rrd$, the maps $x\mapsto x+b$ and $x\mapsto x-c$ are homeomorphisms; thus the map of~\eqref{eq_proof_of_prop1} is a homeomorphism if each reconfiguration unit $A(x)$ is a homeomorphism.  Hence, we show that $A \in \hhh(\rrd)$.  
		
		For notational simplicity, we let $F(x)\triangleq \Skw(f_0)(x)+ A_1\Skw(f_1)(x)L_{\eta}(x)$, $f_0,f_1 \in \NN[d,d(d-1)/2][\sigma_{\operatorname{ReLU}}]$ with $1$ hidden unit, and we observe that 
		the reconfiguration unit $A(x)$ may be written as $A(x)=\exp(F(\|x\|))x$.  Define the map $B(y)\triangleq \exp(-F(\|y\|))y$.  Since $\exp$ is continuous, matrix multiplication is continuous, and $F$ is continuous then both $A$ and $B$ are continuous.  Thus, if $B$ is a two-sided inverse of $A$ then $A$ is a homeomorphism; we show this now.  
		First, observe that for every $z \in \rr$, $F(z)$ is a $d\times d$ skew-symmetric matrix and therefore, by \citep[Section 4]{RohanSOnsurjectiveExp}, for every $x \in\rrd$, $\exp(F(\|x\|))$ is an orthogonal matrix with determinant $\det(\exp(F(\|x\|)))=1$.  Thus, $\exp(F(\|x\|))$ is an isometry fixing the origin; hence
		\begin{equation}
		\|\exp(F(\|x\|))x\|=\|x\|
		\label{eq_isometry_redux}
		.
		\end{equation}
		Therefore,~\eqref{eq_isometry_redux} implies that 
		$$
		\|y\|=\left\|\exp(F(\|x\|))x\right\|=\|x\|
		,
		$$ 
		where $y=F(\|x\|)x$.  Hence, we compute that
		\begin{equation}
		\begin{aligned}
		B(y) = & \exp\left(-F(\|y\|)\right)y 
		\\= & \exp\left(-F(\|x\|)\right)y \\
		= & \exp\left(-F(\|x\|)\right)
		\exp\left(F(\|x\|)\right)x\\
		= & 
		\exp\left(-F(\|x\|) + F(\|x\|)\right)x\\
		= & \exp(0)x\\
		= & I_d x \\
		= & x
		;
		\end{aligned}
		\label{eq_last_computation_in_long_story}
		\end{equation} 
		where $I_d$ is the $d\times d$ identity matrix.  
		Mutatis mundais, the computation of the right-inverse is analogous.  Therefore, $B$ is the two-sided inverse of $A$ and thus $A$ is a homeomorphism.  This gives (i).  
		
		For (ii), since the composition of identity maps is again the identity map then it is enough to demonstrate that any one reconfiguration unit may be parameterized to be the identity.  Indeed, setting 
		$$
		\phi_{\alpha}(x)\triangleq \sigma_{\alpha \alpha_n }\exp\left(
		\alpha\Skw(f_0)(\|x-c\|^2) + \alpha\operatorname{Skw}(f_1)(x)L_{\eta}(\|x-c\|^2)
		\right); \quad \alpha \in [0,1],
		$$
		this gives the result since $\exp(0)=I_d$.  
		
		Next, we complete the proof by showing (iii).  By (ii) and (iv) any reconfiguration unit $A(x)\in \Phi_{\star:d}$.  Let $x,y,z\in \rrd$ be such that $\|x-y\|<\|x-z\|$.  Then, the barycenter $c=2^{-1}(x+y)$ satisfies
		$
		\|x-c\|=\|y-c\|< \|c-z\|.
		$
		We set $\eta\triangleq 2^{-1}\left(\|c-z\|-\|c-x\|\right)$ and $f_0=0$.  
		
		Now, since the subset of $\NN[d,d(d-1)/2][\sigma_{\operatorname{ReLU}}]$ with one hidden unit, contains all constant functions, $\Skw$ is a bijection from $\so{d}$ onto $\rrflex{d(d-1)/2}$, and since any matrix $B \in \exp(\so{d})$ satisfies $B^{\top}B=I_d$ and $\det(B)=1$ then for any $B \in \so{d}$ the constant function $f_1^B\triangleq \Skw^{-1}(\log(B)) $ belongs to $\NN[d,d(d-1)/2][\sigma_{\operatorname{ReLU}}]$. Moreover, the reconfiguration unit 
		$ 
		A^{c,B,\eta}(u)\triangleq \exp\left(
		f_1^B L_{\eta}(\|u-c\|^2)
		\right)(u-c)
		,
		$ 
		belongs to $\Phi_{\star:d}$ and by construction $A^{c,B,\eta}(z)=z$ since $L_{\eta}(\|z-c\|^2)=0$.  Therefore $\exp(f_1^BL_{\eta}(\|z-c\|^2))z=\exp(0)z=z$.  Furthermore, $A^{c,B,\eta}(x)\neq x$ and $A^{c,B,\eta}(y)\neq y$ whenever $B\neq 0$.  
		
		At this point, we seek a matrix $B$ such that such that $Bx=y$, $B^{\top}B=I_d$, and $\det(B)=1$.  This matrix is explicitly computed to be
		\begin{equation}
		\begin{aligned}
		B &= \left(\frac{y-c}{\|x-c\|}-\frac{x-c}{\|x-c\|}\right)\frac{x-c}{\|x-c\|}^{\top}
		+\left(
		\tilde{y}
		-
		\tilde{x}\right)\tilde{x}^{\top}
		,\\
		\tilde{x} & \triangleq \frac{y- \left(y^{\top}x\right)x}{\left\|y- \left(y^{\top}x\right)x\right\|},%
		\quad \tilde{y}  \triangleq \frac{-x+ \left(x^{\top}y\right)y}{\left\|x- \left(x^{\top}y\right)y\right\|},\quad
		c \triangleq \frac{x+y}{2}
		.
		\end{aligned}
		\label{eq_explicit_B}
		\end{equation}
		Since $L_{\eta}(u)>0$ while $|u|<\eta$ then $L_{\eta}(\|x-c\|)^{-1}$ is a well-defined strictly positive number.  Setting $f_1^B\triangleq L_{\eta}(\|x-c\|)^{-1}\log(B)$, where $B$ is given by~\eqref{eq_explicit_B}, gives the conclusion.  
	\end{proof}

	\begin{proof}[{Proof of Proposition~\ref{prop_instability_theorem}}]
		We argue analogously to the proof of \citep[Lemma A.1]{kratsios2020overcoming}.  If $\sigma$ is analytic, then so is the function $\Sigma:\rrd\rightarrow \rrd$ obtained by component-wise application of $\sigma$.  Since the composition of analytic functions is again analytic, and since every affine function is analytic, then every $f \in \NN[d,d]$ is analytic.  Since the difference of two analytic functions is analytic, then for every $f \in \NN[d,d]$, the function 
		$F(x)\triangleq f(x)-x$ is analytic from $\rrd$ to $\rrd$.  
		
		Suppose that property (P-iv) (from page 3) holds.  Denote the zero-set of $F$ by $Z_F\triangleq \left\{x \in \rrd:\, \phi(x)=x\right\}$ and note that by hypothesis we have that $\mu(Z_F)>0$.  Since $Z_F$ is a set of positive Lebesgue measure then it must have a accumulation point and therefore it is identically $0$ by the Principle of Permanence\footnote{The Principle of Permanence is sometimes called the \textit{Uniqueness Theorem of Analytic Functions}}.  Hence, $F$ must be identically $0$ on all of $\rrd$ and therefore $f(x)=x$ for every $x \in \rrd$.  However, this contradicts the hypothesis that $f(x_1)=y_1$ for some $x_1\neq y_1$; both in $\rrd$.  Thus property (P-iv) never holds if $\sigma$ is analytic.
	\end{proof}

	The proof of Proposition~\ref{ex_ReLU_type_Example} uses the following observation.  
	
	\begin{lemma}\label{lem_injectivity_requirement}
		Let $h:X\rightarrow Y$ and $g:Y\rightarrow Z$, where $X,Y,Z$ are non-empty sets.  Then, $g\circ h$ is injective only if $h$ is injective.  
	\end{lemma}
	\begin{proof}[{Proof of Lemma~\ref{lem_injectivity_requirement}}]
		Let $g\circ h:\rightarrow Z$ be injective.  For a contradiction, assume that $h$ is injective.  Then there would exist distinct $x_1,x_2 \in X$ for which $h(x_1)=h(x_2)$.  Therefore, $g\circ h(x_1)=g(f(x_1))=g(f(x_2))=g\circ h(x_1)$ which contradicts the assumed injectivity of $g\circ h$.  Hence, $h$ must be injective.  
	\end{proof}

	\begin{proof}[{Proof of Proposition~\ref{ex_ReLU_type_Example}}]
		Let $f \in \NN[d,D]$.  Clearly, if $d=0$ then $f$ is injective since there is only one point in its domain.  Assume that $d\geq 1$.  Let $\hat{f} \in \NN[d,D][\sigma_{r-\operatorname{ReLU}}]$ be UAP-preserving and therefore by \cite[Theorem 3.4]{kratsios2020non} is must be injective.  
		
		Define the function $g_W\in C(\rrd,\rrd)$ by
		$
		g_W\triangleq \sigma_{r-\operatorname{ReLU}}\bullet W,
		$ where $W$ is an affine function from $\rrd$ to itself.  By
		Lemma~\ref{lem_injectivity_requirement}, $\hat{f}\circ g_W$ is injective only if $g_W$ is injective.  
		
		We show that this is never the case.  To see this, we note that $\sigma_{r-\operatorname{ReLU}} \bullet W$ is injective only if $Im(W)=[0,\infty)^{d}$; where $\rrflex{d}$ is the co-domain of $W$.  By Lemma~\ref{lem_injectivity_requirement}, $\sigma_{r-\operatorname{ReLU}} \bullet W_1$ is injective only if $W$ is also injective.  
		
		We show that these two conditions cannot hold simultaneously.  Suppose that $W$ is injective and takes values in $[0,\infty)^{d}$.  Since $W$ is affine then $W(x)=Ax +b$ for some $d\times d$-dimensional matrix $A$ and some $b\in \rrd$.  Since $W$ is injective then $A\neq 0$.  Let $e_i\in \rrd$ be the vector with entry $1$ in its $i^{th}$ component and zero otherwise; that is, $\{e_i\}_{i=1}^d$ is the standard orthonormal basis of $\rrd$.
		Therefore, $W((-b_j-A_{i,j})e_i)_j=-1$ for every $1\leq j\leq d$ and every $1\leq i\leq d$.  Thus, $Im(W)\cap (-\infty,0)^{d}\neq \emptyset $, which is a contradiction.  In turn, this contradicts the assumption that $\hat{f}\circ \sigma_{r-\operatorname{ReLU}}\bullet W$ is not injective.  By  \cite[Theorem 3.4]{kratsios2020non}, $\hat{f}\circ \sigma_{r-\operatorname{ReLU}}\circ W$ is not UAP-preserving.  Moreover, by definition it is not an embedding on $[0,1]^D$, since all topological embeddings are injective.  Therefore, if $\hat{f} \in \NN[d,D][\sigma_{r-\operatorname{ReLU}}]$ is UAP-preserving, then for every affine map $W:\rrd\rightarrow \rrd$, the DNN $\hat{f}\circ \sigma_{r-\operatorname{ReLU}}\bullet W$ is not; nor is it a topological embedding on $[0,1]^d$.  
	\end{proof}
	
	\subsection{Proof of Theorem~\ref{thrm_memory_capacity_w_guessing}}\label{ss_AProof_Mem_capacity}
	The proof of Theorem~\ref{thrm_memory_capacity_w_guessing} depends on the following lemma.  Recall the length of a piece-wise smooth curve $\gamma:[0,1]\rightarrow \rrd$, denoted by $Len(\gamma)$, is defined by:
	$$
	Len(\gamma) =
	\int_{t \in D} \left\|
	\dot{\gamma}
	(t)
	\right\|dt
	=
	\int_0^1 \left\|
	\dot{\gamma}
	(t)
	\right\|dt
	,
	$$
	where $D$ is a dense subset of $[0,1]$ for which $\mu(D)=1$ and on which $\gamma$ is differentiable, and where $\dot{\gamma}$ is the derivative of $\gamma$ on $D$.  
	\begin{lemma}\label{lem_thrm_memory_capacity_w_guessing_two_point_version}
		Let $d,N\in \nn_+$ with $d>1$ and \textbf{$x_1,\dots,x_N,x,z$} be distinct points in $\rrd$.  Fix $0<\delta < \Delta$ where $\Delta$ is given by:
		$$
		\Delta 
		\triangleq 
		\frac{1}{2}
		\min\left\{2,
		\min_{i,j=1,\dots,N,\, i\neq j} \|x_i-x_j\|
		,
		\min_{i=1,\dots,N} \|x_i-z\|
		,
		\min_{i=1,\dots,N} \|x_i-x\|
		\right\}
		.
		$$
		There exists a piece-wise smooth curve $\gamma_{[x:z]}:[0,1]\rightarrow \rrd$ satisfying
		\begin{enumerate}[(a)]
			\item $\underset{{\underset{t \in [0,1]}{j=1,\dots,n}}}{\min}\,\left\|
			\gamma_{[x:z]}(t) - x_j
			\right\|>0$,
			\item $Len\left(\gamma_{[x:z]}\right) \leq \frac{\pi}{2\Delta}
			,$
			\item $\gamma_{[x:z]}(0)=x$ and $\gamma_{[x:z]}(1)=z$.  
		\end{enumerate}
		In particular, there exists reconfiguration units $A_1,\dots,A_J$ such that
		\begin{enumerate}[(i)]
			\item $A_J\circ \dots \circ A_1(x)=z$,
			\item $A_J\circ \dots \circ A_1(x_i)=x_i$,
			\item There exits a compact subset $K\subseteq \rrd$ for which 
			$A_J\circ \dots \circ A_1(x)=x$ for every $x \not\in K$.
		\end{enumerate}
		Furthermore, the following estimates hold:
		$$
		\begin{aligned}
		&J 
		\leq 
		\left\lceil
		\frac{\pi}{2(\min\{2 \delta,1\})}
		\right\rceil
		,\,
		& N_{\delta}(K)\leq J
		,\,
		&\mbox{ and }
		\mu\left(
		K
		\right)
		\leq 
		\frac{J\delta^2 \pi^\frac{d}{2}}{\Gamma\left(\frac{d}{2} + 1\right)}
		\end{aligned}
		$$
	\end{lemma}
	\begin{proof}[{Proof of Lemma~\ref{lem_thrm_memory_capacity_w_guessing_two_point_version}}]
		Let $n\in \nn$, and $x_1,\dots,x_n,x,z$ be distinct points $\rrd$.  For any $y \in \mathbb{R}^d$ and $\epsilon>0$ let
		$$
		B_{\epsilon}(y)\triangleq \left\{
		p \in \rrd :\,
		\|p-y\|<\epsilon
		\right\}
		.
		$$  We denote the boundary of $B_{\epsilon}(y)$ by $\partial B_{\epsilon}(y)$.  
		We proceed by induction.  If $N=0$ then $\delta>0$ can be chosen arbitrarily.  Let $l_{[x:z]}:[0,1]\rightarrow \rrd$ be the straight line joining $x$ to $z$.  Since this function is piece-wise smooth its length, which we denote $\operatorname{Len}\left(l_{[x:z]}\right)$, is given by
		\begin{equation}
		\operatorname{Len}\left(
		l_{[x:z]}
		\right) = \int_0^1 \left\|
		\dot{
			l}_{[x:z]}
		(t)
		\right\|dt= \|x-z\| 
		\label{eq_length_line}
		.
		\end{equation}
		
		Since $l_{[x:z]}$ and $B_{\delta}(x_i)$ is a convex body in $\rrd$ then $l_{[x:z]}$ can each $\partial B_{\delta}(x_i)$ at-most two points.  Without loss of generality, if $l_{[x:z]}([0,1])\cap B_{\delta}(x_i)=\emptyset$ for some $i=1,\dots,N$ then $l_{[x:z]}$ does not need to be modified to avoid $x_i$; thus we may assume that $l_{[x:z]}([0,1])\cap B_{\delta}(x_i) \neq \emptyset$ for each $i=1,\dots,N$.   For each $i=1,\dots,N$ let $t^i_I<  t^i_O$ be the respective first and final times where $l_{[x:z]}$ intersects $\partial B_{\delta}(x_i)$; these are given by
		$$
		t^i_I\triangleq \min\left\{
		t \in [0,1]:\,
		l_{[x:z]}(t) \in \partial B_{\delta}(x_i)
		\right\}
		\mbox{ and }
		t^i_O\triangleq \min\left\{
		t \in [t^i_I,1]:\,
		l_{[x:z]}(t) \in \partial B_{\delta}(x_i)
		\right\}
		$$
		Accordingly define the points $p^i_I\triangleq l_{[x:z]}(t^i_I)$ and $p^i_O\triangleq l_{[x:z]}(t^i_O)$.  We modify $l_{[x:z]}$ to circumvent $B_{\delta}(x_i)$ and connect $p^i_I$ to $p^i_O$ about a minimal length curve on $\partial B_{\delta}(x_i)$.  From \cite{fletcher2011geodesic} one finds that this is given by a great circle on the sphere $\partial B_{\delta}(x_i)$ given by the curve $\gamma^i:[0,1]\to \partial B_{\delta}(x_i)$%
		$$
		\gamma^i(t)\triangleq \delta^2\left(
		p_I^i \cos\left(
		t \left\|
		v^i
		\right\|
		\right)
		+
		\frac{
			\sin\left(
			t\|v^i\|
			\right)}{
			\|v^i\|
		}v^i
		\right)
		,
		$$
		where $v^i\triangleq \arccos(\langle p_I^i,p_O^i\rangle)
		(p_O^i - \langle p_I^i,p_O^i\rangle p_I^i)
		$ (here we have adjusted the formula in \cite{fletcher2011geodesic} to the case where $\delta \leq 1$ by using basic trigonometry).  Accordingly, we modify $\ell_{[x:z]}$ to the following curve
		\begin{equation}
		\gamma_{[x:z]}(t)\triangleq 
		\begin{cases}
		\gamma^i\left(
		\frac{(t-t_I^i)t_O^i}{(t_O^i-t_I^i)} 
		+
		\frac{(t-t_O^i)t_I^i}{(t_I^i-t_O^i)} 
		\right) & : t \in [t^i_I,t^i_O]
		\\
		l_{[x:z]}(t) &: \, \mbox{ else}			
		\end{cases}
		,
		\label{eq_definition_of_modified_curve_joining_x_to_z}
		\end{equation}
		this is indeed well-defined since $B_{\delta}(x_i)\cap B_{\delta}(x_j)=\emptyset$ for every $i\neq j$ by definition of $\Delta$.  
		
		Since the length of $\gamma^i$ is given by the geodesic distance on the sphere $\partial B_{\delta}(x^i)$ which was shown in \cite{fletcher2011geodesic} to be equal to 
		\begin{equation}
		\delta^2 \arccos\left(\langle p_I^i,p_O^i\rangle \right) \leq \frac{\delta ^2\pi}{2}
		\label{eq_length_sphere_geodesic}
		.
		\end{equation}
		Then, since $\gamma^i$ is piece-wise smooth then, by~\eqref{eq_length_line} and~\eqref{eq_length_sphere_geodesic}, it's length is computed to be
		\begin{equation}
		Len\left(
		\gamma_{[x:z]}
		\right) = \int_0^1 \left\|
		\dot{\gamma}(t)
		\right\|
		dt
		\leq \frac{\pi}{2}\|x-z\|
		.
		\label{eq_length_estimate}
		\end{equation}
		Moreover, by construction $\gamma_{[x:z]}$ satisfies the bound
		\begin{equation}
		\delta
		< 
		\inf_{t\in [0,1],x_1,\dots,x_N} \|
		\gamma_{[x:z]}(t) - x_{i}
		\|
		\label{eq_bounding_curve_away_points}
		.
		\end{equation}
		Thus, (a)-(c) hold.  
		
		Since $Len(\gamma_{[x:z]})\leq \frac{\pi}{2\Delta}<\infty$ then we may pick $1\leq t_1<\dots<t_{\tilde{J}}\leq 1$ such that the length of the segment of $\gamma_{[x:z]}$ between $[t_i,t_{i+1}]$ is at-most $2\delta$; i.e.:
		$$
		\int_{t_i}^{t_{i+1}} 
		\left\|
		\dot{\gamma}(t)
		\right\|
		dt \leq 2 \delta \qquad (\forall i =1,\dots,\tilde{J})
		,
		$$
		where $\tilde{J}\triangleq \left\lceil
		\frac{Len(\gamma_{[x:z]})}{\min{2\delta,1}}
		\right\rceil$ balls of radius $\delta$.  In particular, by~\eqref{eq_length_estimate} we may take 
		$$
		\tilde{J}
		\leq 
		\left\lceil
		\frac{Len(\gamma_{[x:z]})}{\min{2\delta,1}}
		\right\rceil
		\leq
		\left\lceil
		\frac{\pi}{2(\min\{2 \delta,1\})}
		\right\rceil
		.
		$$
		Furthermore, combining with~\eqref{eq_bounding_curve_away_points}, we observe that the collection of closed balls $\left\{
		\overline{
			B_{\delta}\left(
			t_{j}
			\right)
		}
		\right\}_{j=1}^{\tilde{J}}
		$
		satisfies: 
		$$
		\gamma_{[x:z]}([0,1])\subset \bigcup_{j=1}^{\tilde{J}} B_{\delta}(t_j) 
		\mbox{ and } B_{\delta}(t_j) \cap \left\{x_i\right\}_{i=1}^n = \emptyset
		.
		$$	
		By Lemma~\ref{lem_tech_lemmas} (iv) there exists reconfiguration units $A_1,\dots,A_J$ such that for each $j=1,\dots,\tilde{J}-1$ 
		\begin{equation}
		A_j\left(
		\gamma_{[x:z]}(t_j)	
		\right)
		= 
		\gamma_{[x:z]}(t_{j+1})	
		\mbox{ and } 
		A_j(p)=p 
		;\qquad (\forall p \in \rrd - \overline{B_{\delta}(\gamma_{[x:z]}(t_j))})
		\label{eq_conclusion_local_transience}
		.
		\end{equation}
		Let $\tilde{K}\triangleq \bigcup_{j=1}^{\tilde{J}} 
		\overline{B_{\delta}(\gamma_{[x:z]}(t_j))}
		$, by the Heine-Borel theorem each $\overline{B_{\delta}(\gamma_{[x:z]}(t_j))}$ is compact and since $\tilde{J}$ is finite then $\tilde{K}$ is compact.  
		
		Lastly, by the sub-additive of measure and by \citep[Equation 5.19 (iii)]{OnlineFormulaBankShpereVolume}, we have the following estimate of $\mu(\tilde{K})$
		$$
		\mu\left(
		\tilde{K}
		\right)\leq \sum_{j=1}^J
		\mu\left(
		\overline{B_{\delta}(\gamma_{[x:z]}(t_j))}
		\right)
		= 
		\sum_{j=1}^J
		\frac{\delta^2 \pi^\frac{d}{2}}{\Gamma\left(\frac{d}{2} + 1\right)}
		= 
		\frac{J\delta^2 \pi^\frac{d}{2}}{\Gamma\left(\frac{d}{2} + 1\right)}
		.
		$$
		We have defined an open cover of $K$ by the balls $\left\{B_{\delta}\left(
		\gamma_{[x:z]}(t_j)
		\right)\right\}_{j=1}^{\tilde{J}}$.  Therefore $N_{\delta}\left(\tilde{K}\right)\leq \tilde{J}$.  
	\end{proof}
	We will obtain the result now follows from Theorem~\ref{thrm_memory_capacity_w_guessing} by repeatedly applying Lemma~\ref{lem_thrm_memory_capacity_w_guessing_two_point_version}.  
	\begin{proof}[{Proof of Theorem~\ref{thrm_memory_capacity_w_guessing}}]
		We proceed by induction on $N \in \nn_+$.  Suppose that $N=1$, set $x=x_1$, $z=z_1$, and let $\{y_k\}_{k=1}^K$ be given.  Since 
		$$
		\begin{aligned}
		\Delta %
		\leq  &
		\frac1{2}\min\left\{ 2,
		\min{\|u-v\|:\, u\neq v,\, u,v \in \{x_i,z_i\}_{i=1}^N\cup\{y_k\}_{k=1}^K}
		\right\}
		\\
		\leq &
		\frac{1}{2}
		\min\left\{2,
		\min_{i,j=1,\dots,N,\, i\neq j} \|x_i-x_j\|
		,
		\min_{i=1,\dots,N} \|x_i-z\|
		,
		\min_{i=1,\dots,N} \|x_i-x\|
		\right\},
		\end{aligned}
		$$
		then Lemma~\ref{lem_thrm_memory_capacity_w_guessing_two_point_version} applies; hence, there exists some sequence of reconfiguration units 
		$A_{1,1},\dots,A_{1,N_1}$ satisfying Theorem~\ref{thrm_memory_capacity_w_guessing}.  This yields the base case of the induction.  
		
		For the induction hypothesis, suppose that for any $N\in \nn_+$ with $N\geq 1$, there exists some sequence 
		$A_{1,N},\dots, A_{J_N,N}$ and
		some non-empty compact subset $K_{N_n} \subseteq \rrd$ satisfying the conclusion of Theorem~\ref{thrm_memory_capacity_w_guessing} for the given set $\{y_k\}_{k=1}^K$.  Set $x= x_{N+1}$ and $z=z_{N+1}$.  Our requirements on $\Delta$ imply that Lemma~\ref{lem_thrm_memory_capacity_w_guessing_two_point_version} applies; whence, there exists some sequence of reconfiguration units 
		$\{A_{s,N+1}\}_{s=1}^{J_{N+1}}$ and some non-empty compact subset $K_{N+1}\subseteq \rrd$ such that
		\begin{align}
		\label{proof_Theorem_thrm_memory_capacity_w_guessing_first_induction_bound_moving}
		A_{J_{N+1}}\circ \dots \circ A_1(x)&=z\\
		\label{proof_Theorem_thrm_memory_capacity_w_guessing_first_induction_bound_fixing}
		A_{J_{N+1}}\circ \dots \circ A_1(\tilde{x})&=\tilde{x}
		\qquad 
		\forall \tilde{x} \in K_{N+1}\cup \left[
		\{y_m\}_{m=1}^M\cup \{x_n,z_n\}_{n=1}^{N}
		\right]
		\\
		\label{proof_Theorem_thrm_memory_capacity_w_guessing_first_induction_bound_on_J}
		J_{N+1} &\leq \left\lceil
		\frac{\pi}{2(\min\{2 \delta,1\})}
		\right\rceil
		\\
		\label{proof_Theorem_thrm_memory_capacity_w_guessing_first_induction_bound_on_compact}
		\mu\left(
		K_{N+1}
		\right) & \leq 
		\frac{J_{N+1}\delta^2 \pi^\frac{d}{2}}{\Gamma\left(\frac{d}{2} + 1\right)},\\
		\label{proof_Theorem_thrm_memory_capacity_w_guessing_first_induction_bound_covering_number}
		N_{\delta}(K_{N+1}) & \leq J_{N+1}
		.
		\end{align}
		Consider the sequence of reconfiguration units
		$
		A_{1,N},\dots,A_{J_N,N},A_{1,N+1},\dots,A_{J_{N+1},N+1}
		$ 
		and the set $K\triangleq K_N \cup K_{N+1}$.  Note that this sequence is of length $J\triangleq J_{N+1} + J_N$.  Note that the set $K$ is compact since the finite union of compact subsets of $\rrd$ is again compact, $K_N$ is compact by induction hypothesis, and $K_{N+1}$ is compact by Lemma~\ref{lem_thrm_memory_capacity_w_guessing_two_point_version}.  
		
		Consider the reconfiguration network 
		$
		\Phi\triangleq
		A_{J_{N+1},N+1}\circ \dots
		\circ 
		A_{1,N+1}\circ \dots
		\circ 
		A_{J_N,N}\circ \dots 
		\circ 
		A_{1,N}
		\dots \circ A_{1,N}
		$.  For notational simplicity let 
		$\Phi_1\triangleq 
		A_{J_N,N}\circ \dots \circ A_{1,N},
		$ 
		and 
		$
		\Phi_2\triangleq 
		A_{J_{N+1},N+1}\circ \dots \circ A_{1,N+1}
		.$    
		By the induction hypothesis, $\Phi_1(x)=x$ and $\Phi_1(z)=z$.  Therefore,~\eqref{proof_Theorem_thrm_memory_capacity_w_guessing_first_induction_bound_moving} implies that
		$$
		\Phi(x_{N+1})=\Phi(x)= \Phi_2\circ \Phi_1(x)= \Phi_2(x)= z=z_{N+1}.
		$$
		Again, by the induction hypothesis, we have that $\Phi_1(x_n)=z_n$ for each $n\leq N$.  Hence,~\eqref{proof_Theorem_thrm_memory_capacity_w_guessing_first_induction_bound_fixing} implies that
		$
		\Phi(x_{n})=\Phi_2\circ \Phi_1(x)= \Phi_2(z_n)= z_n,
		$ 
		for each $n\leq N$.  Thus (i) holds.  
		
		By the induction hypothesis, $\Phi_1(y_m)=y_m$ for each $m\leq M$.  Thus,~\eqref{proof_Theorem_thrm_memory_capacity_w_guessing_first_induction_bound_fixing} yields
		$$
		\Phi(y_m) = \Phi_2\circ \Phi_1(y_m)= \Phi_2(y_m)= y_m
		,
		$$
		for each $m\leq M$.  Hence, (ii) holds. 
		
		Since, $K_N,K_{N+1}\subseteq K$ then $\rrd-K \subseteq \left[ \rrd - K_N \right]\cap \left[\rrd -K_{N+1}\right]$.   Thus, the induction hypothesis implies that $\Phi_1|_{\rrd - K_N}(p)=p$ for every $p \in \rrd-K$.  Likewise, by definition of $K_{N+1}$, $\Phi_2|_{\rrd - K_N}(p)=p$ for every $p \in \rr-K_{N+1}$.  Therefore, for every $p \in \rrd-K$ we have that
		$ 	
		\Phi(p)=\Phi_2\circ \Phi_1(p)= \Phi_2(p)=p
		.
		$  Therefore, (iii) holds.  
		
		Next, by the induction hypothesis, the definition of $J$, and by~\eqref{proof_Theorem_thrm_memory_capacity_w_guessing_first_induction_bound_on_J} we have that
		$$
		\begin{aligned}
		J &=  J_N + J_{N+1} %
		\leq 
		\left\lceil
		\frac{N\pi}{2(\min\{2 \delta,1\})}
		\right\rceil
		+
		\left\lceil
		\frac{\pi}{2(\min\{2 \delta,1\})}
		\right\rceil
		\leq \left\lceil
		\frac{(N+1)\pi}{2(\min\{2 \delta,1\})}
		\right\rceil
		.
		\end{aligned}
		$$
		This gives the bound on $J$.  
		Lastly, combining the induction hypothesis on the covering number of $K_{N}$ with the bound of~\eqref{proof_Theorem_thrm_memory_capacity_w_guessing_first_induction_bound_covering_number} we make the following computation
		$$
		\begin{aligned}
		N_{\delta}(K)\leq & N_{\delta}(K_N) + N_{\delta}(K_{N+1})
		\leq J_N + J_{N+1} = J.
		\end{aligned}
		$$
		This completes the induction hypothesis and therefore the proof.  
	\end{proof}
	\subsection{{Proof of Theorem~\ref{thrm_uniform_bound}}}\label{a_proof_thrm_3}
	\begin{proof}[{Proof of Theorem~\ref{thrm_uniform_bound}}]
		Let $\phi \in \hhh_{M,\omega}^{\delta}(\rrd)$, $M,\delta>0$, and $d \in \nn_+$.  By definition, there exists $\{x_n\}_{n\leq N}$ in $\rrd$ and $\{\varphi_n\}_{n\leq N}$ in $\hhh(\rrd)$ such that $[-M,M]^d\subseteq\cup_{n\leq N}B(x_n,\delta)$,~\eqref{eq_fragmentation_property_description}, and~\eqref{eq_definition_HMdeltaomega_1}-\eqref{eq_definition_HMdeltaomega_4} hold.  
		
		\textbf{Step 1 - Upper-bound $N$:} 
		Let us begin by upper-bounding $N$.  Observe that if $x \in [-M,M]^d$ then $\|x\|\leq 2\sqrt{d}M$.  Therefore,~\citep[page 337]{shalev2014understanding} implies that
		\begin{equation}
		\begin{aligned}
		N_{2^{-1}\delta}([-M,M]^d)\leq & \sup_{x \in [-M,M]} \left(\frac{2\|x\|\sqrt{d}}{2^{-1}\delta}\right)^d\\
		= & 
		\left(\frac{8\sqrt{d}M\sqrt{d}}{\delta}\right)^d
		.
		\end{aligned}
		\label{eq_proof_thrm_uniform_bound}
		\end{equation}
		Next, $[-M,M]^d\subseteq\cup_{n\leq N}B(x_n,\delta)$ and~\eqref{eq_definition_HMdeltaomega_3}, together with the bound on the packing number $N_{2^{-1}\delta}^{pack}([-M,M]^d)$ in \citep[Lemma 27.3]{shalev2014understanding} imply that
		\begin{equation}
		\begin{aligned}
		N\leq N_{\delta}([-M,M]^d)\leq & N_{2^{-1}\delta}^{pack}([-M,M]^d)
		\leq 
		N_{2^{-1}\delta}([-M,M]^d)
		.
		\end{aligned}
		\label{eq_proof_thrm_uniform_bound_1}
		\end{equation}
		Combining~\eqref{eq_proof_thrm_uniform_bound} with~\eqref{eq_proof_thrm_uniform_bound_1} yields the following upper-bound on $N$
		\begin{equation}
		N\leq N_{2^{-1}\delta}([-M,M]^d)\leq 
		\left(\frac{8dM}{\delta}\right)^d
		\label{eq_proof_thrm_uniform_bound_2}
		.
		\end{equation}
		\textbf{Step 2 - Representation of each $\varphi_n$:} 
		Next, let us determine the form of any $\phi \in \hhh_{M,\omega}^{\delta}(\rrd)$.  By the fragmentation representation~\eqref{eq_fragmentation_property_description}, we only need to describe a single $\varphi_n$ for $n\leq N$.  
		In what follows, for $x \in \rrd$ and $r\geq 0$, we use $S^{d-1}(x,r)$ to denote the $d-1$ dimensional sphere in $\rrd$ with center $x$ and radius $r$; defined by $S^{d-1}(x,r)\triangleq \left\{z \in \rrd:\, \|x-z\|=r\right\}$.  Together, properties~\eqref{eq_definition_HMdeltaomega_1} and~\eqref{eq_definition_HMdeltaomega_2} imply that $\varphi_n$ maps $S^{d-1}(x_n,\delta)$ to itself and that $\varphi_n|_{S^{d-1}(x_n,r)}:S^{d-1}(x_n,\delta) \rightarrow S^{d-1}(x_n,r)$ is an isometry, for every $r\geq 0$. Since any isometry on $\rrd$ belongs to the Euclidean group\footnote{The Euclidean group is set of all isometries on $\rrd$.  A map $f \in C(\rrd,\rrd)$ belongs to the Euclidean group if and only if it is an affine map of the form $f(x)=Ax+b$ where $A$ is an orthogonal matrix and $b\in \rrd$.  } then it is of the form $\varphi_n|_{S^{d-1}(x_n,r)} = A_{r}x+b_r$ for some orthogonal matrix $A_r$ and some $b_r \in \rrd$.  However, since $\varphi_n$ maps each $S^{d-1}(x_n,r)$ into itself then it must fix $x_n$ and therefore $b_n = -A_rx_n$.  Next, since $\varphi_n \in \hhh(\rrd)$ then it is orientation-preserving and therefore $A_r \in SO(d)$.  Thus, for every $n\leq N$ and every $r\geq 0$
		\begin{equation}
		\varphi_n(x)
		= A_{\|x\|}(x-x_n)
		\label{eq_representation_1}
		.
		\end{equation}
		The map $\exp:\so{d}\mapsto SO(d)$ is a continuous surjection with continuous right-inverse $\log:SO(d)\mapsto \so{d}$ (see \citep[Section 4]{RohanSOnsurjectiveExp}).  Hence, the map $r \mapsto \tilde{R}_r\triangleq \log\left(A_{r}\right)$ from $[0,\infty)$ to $\so{d}$ must be continuous and by~\eqref{eq_representation_1} it satisfies
		\begin{equation}
		\varphi_n(x)
		= \exp\left(\tilde{R}_{\|x-x_n\|}\right)(x-x_n)
		\label{eq_representation_2a}
		.
		\end{equation}
		since $\Skw:\rrflex{d(d-1)/2}\rightarrow\so{d}$ is a bijective isometry then there exists some function $R \in C([0,\infty),\rrflex{d(d-1)/2})$ satisfying 
		$
		\Skw\circ R = \tilde{R}.
		$
		Thus,~\eqref{eq_representation_2a} simplifies to
		\begin{equation}
		\varphi_n(x)
		= \exp\left(\Skw{R}_{\|x-x_n\|}\right)(x-x_n)
		\label{eq_representation_2}
		.
		\end{equation}
		Since $\varphi_n\in \hhh(\rrd)$ then representation~\eqref{eq_representation_2}, the continuity of $\log$ on $SO(d)$, $\Skw^{-1}$ on $\so{d}$, the continuity of the Euclidean norm, the continuity of affine transformations, and the fact that the composition of continuous functions is again continuous all together imply that the map $x\mapsto \exp(\Skw(R_{\|x-x_n\|}))(x-x_n)$ must be also be continuous.  
		
		\textbf{Step 3 - Representation of Reconfiguration Network's Layers:}
		For any given $\tilde{N},n\in \nn_+$ with $n\leq N$, whenever we set
		$\alpha_k=0$, $\eta_k=0$ for $1\leq k\leq \tilde{N}_n$, $c_k=b_k=0$ for $1<k\leq \tilde{N}_n$, and $-c_1=b_1=x_n$, we can represent any such reconfiguration network $\phi^{(n)} \triangleq \phi_{\tilde{N}_n,n}\circ \dots \circ \phi_{1,n}$, with $\phi_{k,n}$ for the form:
		\begin{equation}
		\phi_{k,n}(x) = 
		\prod_{k\leq \tilde{N}_n}
		\exp\left(
		\Skw(W_{2,k:n}\circ\sigma_{\operatorname{ReLU}}\bullet W_{1,k:n})(\|x-x_n\|^2)
		\right)(x - x_n) 
		.
		\label{eq_segment_representation_1}
		\end{equation}
		Using the fact that $\exp(A+B)=\exp(A)\exp(B)$ for any $d\times d$ matrices $A$ and $B$ and the linearity of $\Skw$, we rewrite~\eqref{eq_segment_representation_1} as
		\begin{equation}
		\phi_{k,n}(x) = 
		\exp\left(
		\Skw\left(
		\sum_{k\leq \tilde{N}_n}
		W_{2,k:n}\circ\sigma_{\operatorname{ReLU}}\bullet W_{1,k:n}(\|x-x_n\|^2)
		\right)
		\right)(x - x_n) 
		.
		\label{eq_segment_representation_reconfiguration_network}
		\end{equation}
		Finally, by construction each $\phi_{k,n}$ maps each $S^{d-1}(x_n,r)$, for $r\geq 0$, isometrically into itself we have that $\|\phi_{k,n}(x)-x_n\|=\|x-x_n\|$, for every $x \in \rrd$ and every $k\leq \tilde{N}$, and therefore 
		\begin{equation}
		\phi^{(n)}(x) = 
		\phi_{\tilde{N}_n,n}\circ \dots \circ \phi_{1,n}
		=
		\exp\left(
		\Skw\left(
		f_n(\|x-x_n\|^2)
		\right)
		\right)(x - x_n) 
		,
		\label{eq_segment_representation_reconfiguration_network_2}
		\end{equation}
		where $f_n \in \NN[1,d(d-1)/2][\sigma_{\operatorname{ReLU}}]$ with $1$ hidden layer and of width $\nu_{\tilde{N}_n}\triangleq \frac{\tilde{N}_n d(d-1)}{2}$.  
		
		Thus, in the next step, we consider reconfiguration networks of the form $\hat{\phi}(x)= \phi^{(N)}\circ \dots \circ \phi^{(1)}$; where each $\phi^{(n)}$ is represented via~\eqref{eq_segment_representation_reconfiguration_network_2}.  Hence, $\phi$ will always be a reconfiguration network of depth $\sum_{n\leq N} \nu_{\tilde{N}_n}$.  
		
		\textbf{Step 4 - Upper-bounding the Modulus of Continuity of $r\mapsto R_r$: }
		We conclude this portion of the proof by computing an upper-bound of the modulus of continuity of $R$.  Indeed, since we are estimating on $[-M,M]^d$ and since the Euclidean norm maps $[-M,M]^d$ to the compact interval $[0,2\sqrt{d}M]$ then we only need to concern ourselves with approximating the modulus of continuity of $r\mapsto R_r$ on $[0,2\sqrt{d}M]$.  Since $r\mapsto R_r$ is continuous and since $[0,2\sqrt{d}M]$ is compact, then by the Heine-Cantor Theorem (\citep[Theorem 27.6]{munkres2014topology}) every continuous function on a compact set is uniformly continuous.  Hence, $R$ genuinely admits an optimal modulus of continuity, which we denote by $\omega_R$.  
		
		If $f,g$ are composable with respective optimal moduli of continuity $\omega_f$ and $\omega_g$ then the optimal modulus of continuity of $g\circ f$, denoted $\omega_{g\circ f}$, satisfies the bound
		\begin{equation}
		\omega_{g\circ f}(t)\leq \omega_{g}\circ \omega_f(t)
		\label{eq_bound_composition_uniform_bound}
		,
		\end{equation}
		for every $t \in [0,\infty)$.  
		Since the Riemannian exponential map on $SO(d)$ coincides with its Lie exponential at the identify and therefore by \citep[Equation 1.11]{sternberg2004lie}, we know that for every pairs of $d\times d$ skew symmetric matrices $X$ and $Y$ we have that:
		\begin{equation}
		\begin{aligned}
		d\exp(Y) & = \exp(I_d)\int_0^1 \exp(-sX)Y\exp(sX)ds\\
		& = \int_0^1 Yds\\
		& = Y;
		\end{aligned}
		\label{eq_babycompute}
		\end{equation}
		where we have use the fact that if $X\in \so{d}$, and similarly for $sX$ for any $s \in [0,1]$ since $\so{d}$ is a vector space, then $\exp(X)$ is an isometry.  Since $\exp$ is smooth then the mean-value theorem and~\eqref{eq_babycompute} imply that $\exp$ is $1$-Lipschitz from $\so{d}$ with the Fr\"{o}benius norm $\|\cdot\|_F$ to $SO(d)$ with the its Riemannian distance and, in particular, so for its Euclidean norm.  Furthermore, the generalized right-inverse of $\omega_{\exp}$, in the sense of \cite{embrechts2013note}, is defined by $$\omega_{\exp}^{-1}(t)\triangleq 
		\inf\{s \in \rr:\omega_{\exp}(s)\geq t\}
		=t
		$$
		and it satisfies $t\leq \omega_{\exp}\circ \omega_{\exp}^{-1}(t)$ for every $t \in [0,\infty)$.  
		These considerations, together with \citep[Proposition 2.3]{embrechts2013note} and~\eqref{eq_definition_HMdeltaomega_4} imply that
		\begin{equation}
		\begin{aligned}
		\omega_{R}(t)\leq \omega_{R}\circ \omega_{\exp} \circ \omega_{\exp}^{-1}(t) =& \omega_{\exp\circ R} \circ \omega_{\exp}^{-1}(t) 
		\\
		= &
		\omega_{\phi_n}\circ \omega_{\exp}^{-1}(t)\leq \omega \circ \omega_{\exp}^{-1}(t)\\
		=& \omega (t)
		,
		\end{aligned}
		\label{eq_first_bound_on_modulus_of_R}
		\end{equation}
		for every $t \in [0,\infty)$.  
		
		\textbf{Step 5 - Computing Bounds (Single block Case): }
		By \citep[Excersize 106]{DeniseMatrices2002}, the map $\exp:\so{d}\rightarrow SO(d)$ is $1$-Lipschitz when $\so{d}$ is equipped with the operator norm $\|\cdot\|_{op}$.  Combining the representations for $\phi_n$ and of $\hat{\phi}^{(n)}$ is steps $2$ and $3$, respectively, we compute the following estimate.
		\begin{equation*}
		\begin{aligned}
		\left\|
		\phi_n(x) - \phi^{(n)}(x)
		\right\| = &
		\left\|
		\exp\left(\Skw\circ R_{\|x-x_n\|}\right)(x-x_n) - 
		\exp\left(
		\Skw\left(
		f_n(\|x-x_n\|^2)
		\right)
		\right)(x - x_n)
		\right\|\\
		\leq &
		\left\|
		\exp\left(\Skw\circ R_{\|x-x_n\|}\right) - 
		\exp\left(
		\Skw\left(
		f_n(\|x-x_n\|^2)
		\right)
		\right)
		\right\|_{op}\|(x - x_n)\|\\
		\leq &
		\left\|
		\exp\left(\Skw\circ R_{\|x-x_n\|}\right)- 
		\exp\left(
		\Skw\left(
		f_n(\|x-x_n\|^2)
		\right)
		\right)
		\right\|_{op} 2\sqrt{d}M\\
		\leq &
		\left\|
		\Skw\circ R_{\|x-x_n\|} - 
		\Skw\circ f_n(\|x-x_n\|^2)
		\right\|_{op} 2\sqrt{d}M
		\\
		= & \sqrt{d}\left\|
		\Skw\circ R_{\|x-x_n\|} - 
		\Skw\circ f_n(\|x-x_n\|^2)
		\right\|_{F} 2\sqrt{d}M
		\\
		= & 
		\left\|
		R_{\|x-x_n\|} - 
		f_n(\|x-x_n\|^2)
		\right\| 2dM
		\label{eq_computation_bound}
		;
		\end{aligned}
		\end{equation*}
		where appealed the bound $\|\cdot\|_{op}\leq \sqrt{d}\|\cdot\|_F$ and the fact that $\Skw$ is an isometry between $\so{d}$ with the Fr\"{o}benius norm, and $\rrflex{d(d-1)/2}$ with the Euclidean norm $\|\cdot\|$.  
		Since~\eqref{eq_first_bound_on_modulus_of_R} gives us a uniform upper-bound of the modulus of continuity of $R$, and since $f_n \in \NN[1,\frac{d(d-1)}{2}][\sigma_{\operatorname{ReLU}}]$ width (breadth) $\frac{\nu_{\tilde{N}_n}d(d-1)}{2}$ and one hidden layer, then we may apply \citep[Proposition 1]{pmlrv75yarotsky18a} to the left-hand side of~\eqref{eq_computation_bound} to obtain the following bound
		\begin{equation}
		\left\|
		\phi_n(x) - \phi^{(n)}(x)
		\right\| 
		=
		\left\|
		R_{\|x-x_n\|} - 
		f_n(\|x-x_n\|^2)
		\right\| 2dM
		\leq 2dM C
		\omega\left(
		\left(\frac{\nu_{\tilde{N}_n}d(d-1)}{2}\right)^{\frac{-1}{d}}
		\right)
		\label{eq_computation_bound_done}
		,
		\end{equation}
		where $C>0$ is a constant independent $\nu$.  
		
		\textbf{Step 5 - Computing Bounds (General Case): }
		Suppose that $N$ is known exactly.  We make the following abbreviations 
		$ 
		\omega_{N-1}\triangleq 
		\sup_{\|x\|\leq M} \left\|
		\phi^{(N-1)}\circ \dots \circ \phi^{(1)}(x) - \varphi_{N-1}\circ \dots \circ \varphi_1(x)
		\right\|
		$, $\tilde{\varphi}_{N-1}\triangleq \varphi_{N-1}\circ \dots \circ \varphi_1$ and $\tilde{\phi}^{N-1}\triangleq \phi^{(N-1)}\circ \dots \circ \phi^{(1)}$.  
		Then, using~\eqref{eq_definition_HMdeltaomega_4}, the definition of the modulus of continuity of $\phi_{N}$, and Step 4 we compute the following recursive-bound for any $x \in [-M,M]^d$.  
		\begin{equation}
		\begin{aligned}
		\left\|
		\phi_n\circ \tilde{\phi}^{(n-1)}(x) - \varphi_{n}\circ \tilde{\varphi}_{n-1}(x)
		\right\| 
		= &
		\left\|
		\phi_n\circ \tilde{\phi}^{(n-1)}(x) 
		- 
		\varphi_{n}\circ \tilde{\phi}^{(n-1)}(x)
		\right.
		\\
		& 
		\left.
		- 
		\varphi_{n}\circ \tilde{\varphi}_{n-1}(x)
		+
		\varphi_{n}\circ \tilde{\phi}^{(n-1)}(x)
		\right\| 
		\\
		\leq 
		&
		\left\|
		\phi_n\circ \tilde{\phi}^{(n-1)}(x) 
		- 
		\varphi_{n}\circ \tilde{\phi}^{(n-1)}(x)
		\right\|
		\\+ & 
		\left\|
		\varphi_{n}\circ \tilde{\phi}^{(n-1)}(x)
		-
		\varphi_{n}\circ \tilde{\varphi}_{n-1}(x)
		\right\| \\
		\leq &
		\left\|
		\phi_n\circ \tilde{\phi}^{(n-1)}(x) 
		- 
		\varphi_{n}\circ \tilde{\phi}^{(n-1)}(x)
		\right\|\\
		+ & 
		\omega\left(
		\left\|
		\tilde{\phi}^{(n-1)}(x)
		-
		\tilde{\varphi}_{n-1}(x)
		\right\| 
		\right)
		\\
		\leq &
		2dM C
		\omega\left(
		\left(\frac{\nu_{\tilde{N}_n}d(d-1)}{2}\right)^{\frac{-1}{d}}
		\right)
		\\
		+ &
		\omega\left(
		\omega_{n-1}
		\right)
		\end{aligned}
		\label{eq_upper_Bound}
		\end{equation}
		Set $\nu_{\tilde{N}_n}\triangleq \nu \left\lceil
		\frac{\delta^d}{(8dM)^d}
		\right\rceil$.  
		By the bound~\eqref{eq_proof_thrm_uniform_bound_2}, we know that $
		\nu_{\tilde{N}_n}\triangleq \nu \left\lceil
		\frac{\delta^d}{(8dM)^d}
		\right\rceil \leq \frac{\nu}{N}$ and by the monotonicity of modulus of continuity $\omega$ we have that for every $k\geq 0$
		\begin{equation}
		\omega\left(
		\nu_{\tilde{N}_n} k
		\right)
		\leq 
		\omega\left(
		\frac{\nu}{N}k
		\right)
		\label{break_down_bound}
		.
		\end{equation}
		Therefore, we have the following approximation bound
		\begin{equation*}
		\sup_{\|x\|\leq M}
		\,
		\left\|
		\phi_n\circ \tilde{\phi}^{N-1}-\hat{\phi}^{(N)}\circ \tilde{\phi}^{(N-1)}(x)
		\right\|
		\leq 
		\omega_{N} 
		,
		\end{equation*}
		where $ \omega_{N}$ is determined by the recurrence relation:
		\begin{equation}
		\omega_{n+1} = 2d MC \omega\left(
		\left(\frac{\nu_{\tilde{N}_n}d(d-1)}{2}\right)^{\frac{-1}{d}}
		\right) + \omega\left(\omega_{n}\right),
		\qquad
		\omega_0=0
		.
		\label{eq_thrm_uniform_bound_formal_statement_recurrence_relation_inside_proof}
		\end{equation}
		Since $\omega$ takes values in the non-negative integers; then, the sequence $\{\omega_n\}_{n\in \nn}$ is monotonically increasing.  Whence, $\omega_N\leq \omega_{\left\lceil 
			\frac{(8dM)^d}{\delta^d}
			\right\rceil}$.  
		This gives us the desired bound.  
	\end{proof}
	
	\subsection{Proof of Theorem~\ref{cor_GR_UAT}}\label{ss_AProof_UAT_General}
	The proof of Theorem~\ref{cor_GR_UAT} relies on the observation that given any $(d+D)\times d$ matrix $A$, any $f \in C(\rrd,\rrD)$ can be expressed as $f = P\circ \Phi_f(x,Ax)$ where $P:\rrflex{d+D}\rightarrow \rrD$ is the (linear) orthogonal projection mapping $(x,y)\in \rrd\times \rrD$ to $y\in \rrD$ and $\Phi_f \in \hhh(\rrflex{d+D})$ is given by
	$$
	\Phi_{f,A}(x,y) = \left(x,y+f(x)-Ax\right)
	.
	$$
	Since $P$ is $1$-Lipschitz, $A$ is arbitrary, then we only need to approximate $\Phi_{f,A}$; which we know can always be done using Theorem~\ref{thrm_UAH}.  
	\begin{proof}[{Proof of Theorem~\ref{cor_GR_UAT}}]
		By Theorem~\ref{thrm_UAH} since $\Phi_{f,A}\in\hhh(\rrflex{d+D})$ there exists some $\hat{\phi}\in \NN[\star:d+D][]$ satisfying 
		\begin{equation}
		\sup_{(x,y)\in \rrflex{d+D};\, \|(x,y)\|\leq M}
		\left\|
		\Phi_{f,A}(x,y) - \hat{\phi}(x,y)
		\right\|<\epsilon
		\label{proof_cor_GR_UAT_1}
		.
		\end{equation}
		Since the orthogonal projection $P(x,y)\triangleq y$ is $1$-Lipschitz, then~\eqref{proof_cor_GR_UAT_1} yields (iv) via
		\begin{equation}
		\begin{aligned}
		\sup_{\|x\|\leq M}
		\left\|
		f(x) - P\hat{\phi}(x,Ax)
		\right\|
		= &
		\sup_{\|x\|\leq M}
		\left\|
		P\circ \Phi_{f,A}(x,Ax) - P\hat{\phi}(x,Ax)
		\right\|\\
		= &
		\sup_{\|x\|\leq M}
		1
		\left\|
		\Phi_{f,A}(x,Ax) - \hat{\phi}(x,Ax)
		\right\|\\
		<&\epsilon
		.
		\end{aligned}
		\label{proof_cor_GR_UAT_2}
		\end{equation}
		Setting $B\triangleq P$, yields (iii).

		Since $x\mapsto I_d\oplus Ax$ is a linear map, it is continuous.  By construction, $I_d\oplus A$ is of fullrank; whence the associated linear map (obtained by matrix-multiplication) is injective and since $\hat{\phi} \in \hhh(\rrflex{d+D})$, then it is also continuous and injective.  Hence, $\psi(x)\triangleq \hat{\phi}(I_d\oplus Ax)$ is a continuous injective map.  Since $\psi$ is continuous then the pre-image of any compact subset $K\subseteq \rrflex{d+D}$ is closed in $\rrd$ and $\psi^{-1}[K]\cap [-M,M]^d$ is a closed subset of the compact-set $[-M,M]^d$.  Hence, $\psi^{-1}[K]\cap [-M,M]^d$ is compact in $[-M,M]^d$.  Therefore, $\psi$ is a continuous, injective, and by \citep[Exercises 26.6]{munkres2014topology} it is a closed map also.  Thus, it is a topological embedding; i.e. a homeomorphism onto its image with respect to the subspace topology.  
		This gives (ii) and the statement that $\psi$ is an isometry in (i).  
		
		Moreover, since $\psi$ is a homeomorphism then it is a bijection with inverse $\phi^{-1}\circ A^{\dagger}$ where $I_d\oplus A^{\dagger}$ is the left-inverse of $I_d\oplus A$ given in the statement of (i).  Since $\phi$ is a bijection then the metric 
		$$
		d_{\hat{\phi},M,\epsilon}(z_1,z_2)\triangleq \|\psi^{-1}(z_1)-\phi^{-1}(z_2)
		$$ is well-defined on $\mmm_{\hat{\phi},M,\epsilon}$ and by construction $\psi$ is an isometry from $[-M,M]^d$ to $\mmm_{\hat{\phi},M,\epsilon}$.  This completes the proof of (i).  
	\end{proof}
	
	\subsection{Proof of Theorem~\ref{thrm_UAH}}\label{ss_AProofs_UAH}
	\subsubsection{{Comments on the method of proof for Theorem~\ref{thrm_UAH}}}\label{ss_comments_method_of_proof}
	We begin by noting that the space $\hhh(\rrd)$ is highly non-linear.  To see this, observe that the identity map $1_{\rrd}$ belongs to $\hhh(\rrd)$, however, $1_{\rrd}-k1_{\rrd} \not\in \hhh(\rrd)$ for any $k>0$.  
	
	This non-linearity unfortunately renders most classical tools used to establish universality, such as the Stone-Weirestra{\ss} theorem (used in \cite{hornik1989multilayer}) from classical approximation theory, the Hahn-Banach theorem (used in \cite{MicchelliUniversalFeature}) from functional analysis, the Wiener-Tauberian Theorem (used in \cite{Cybenko}) from harmonic analysis, or hypercylicity results (used in \cite{kratsios2020characterizing}) from linear dynamical systems, useless.
	
	Nevertheless, we still have access to the less standard tools from infinite-dimensional topology; which were specifically built to handle limits of homeomorphisms in $\hhh(\rrd)$.  Specifically, we appeal to \textit{the inductive convergence criterion}, a less-known result which provides conditions on a sequence of homeomorphisms $(\Phi_k)_{k \in \nn_+}$ in $\hhh(\rrd)$ such that the limit of $\Phi_k\circ \dots\Phi_1$ in the topology of $\hhh(\rrd)$ exists.  For a self-contained exposition, we briefly discuss state and discuss the result here before applying it.  
	
	The result relies on some notation which we now develop.  Let $\epsilon>0$, $K\subseteq \rrd$ be non-empty and compact.  For any $\Phi\in \hhh(K)$ define the quantity:
	$$
	\alpha(h,\epsilon)\triangleq \inf\left\{
	\delta : \exists x,y \in K,\, \|x-y\|>\epsilon \mbox{ and }
	\|\Phi(x)-\Phi(y)\|=\delta
	\right\}.
	$$
	As discussed on \citep[page 367]{AndersonOnTopInfDeficiency}, if $d\in \nn_+$ then for any $\Phi\in \hhh(K)$ and any $\epsilon>0$ we have that $\alpha(\Phi,\epsilon)>0$.  Several formulations of this lemma exist, see \cite{VanMillInductiveLimitCriterion,CountabledensehomogeneityoffunctionspacesRodrigo}; however, we prefer to use of the phrasing presented in \cite{AndersonOnTopInfDeficiency}.  
	\begin{lemma}[{Inductive Convergence Criterion \citep[Lemma 2.1]{AndersonOnTopInfDeficiency}}]\label{lem_ICC}\hfill\\
		If $\left(\phi_k\right)_{k \in \nn_+}$ is a sequence in $\hhh(K)$ and if
		$$
		\sup_{x \in K} \left\|
		\phi_k(x) - x
		\right\| < 3^{-k}\min\left\{
		1
		,
		\alpha\left(\phi_{k-1}\circ \dots\circ \phi_1, 2^{-k}\right)
		\right\},
		$$
		for each $k>1$ then $\left\{\phi_k\circ \dots \circ \phi_1\right\}_{k \in \nn_+}$ converges (to a homeomorphism) in $\hhh(K)$.  
	\end{lemma}
	\subsubsection{{Proof for Theorem~\ref{thrm_UAH}}}\label{sss_comments_method_of_proof}
	\begin{proof}[{Proof of Theorem~\ref{thrm_UAH}}]
		Let $d\in \nn_+$ with $d\geq 2$, $K\subseteq \rrd$ be compact, and let $\phi\in\hhh(\rrd)$.  We deduce some simplifications on $K$, on $\phi$, and on their relationship before moving onto the main part of the proof.  
		Since $\phi\in \hhh(\rrd)$ then it is isotopic to the identity; thus, the second Corollary\footnote{The results are not numbered in \cite{KirbyStableHomeos1969}.} to \citep[Theorem 2]{KirbyStableHomeos1969} implies that $\phi$ can be fragmented, i.e.: for any open cover $\mathcal{U}\triangleq \{U_n\}_{n=1}^N$ of $\rrd$ there exists $\phi_{\mathcal{U}:1},\dots,\phi_{\mathcal{U}:N} \in \hhh(\rrd)$ such that
		\begin{equation}
		\begin{aligned}
		\phi &= \phi_{\mathcal{U}:N}\circ \dots \circ \phi_{\mathcal{U}:1} &\\
		\phi_{\mathcal{U}:n}(x) & = x &\,(\forall x \not\in U_n)
		\label{eq_fragmentation_property}
		.
		\end{aligned}
		\end{equation}
		In particular,~\eqref{eq_fragmentation_property} implies that $\phi_{\mathcal{U}:n}(U_n)= U_n$, for every $n\leq N$, since the value of $\phi_n$ is determined to be the identity outside of $U_n$.  In particular, fix some $\delta>0$ and define $K_{\delta}'\triangleq \left\{z \in \rrd:\, (\exists x \in K)\, \|z-x\|< \delta\right\}$; then for the open cover $\left\{K_{\delta}',\rrd-K\right\}$ of $\rr$, the fragmentation property of $\phi$ implies that there are $\{\phi_{\mathcal{U}:i}\}_{i=1}^2 \subset \hhh(\rrd)$ satisfying
		\begin{equation}
		\begin{aligned}
		\phi  &= \phi_{\mathcal{U}:2}\circ \phi_{\mathcal{U}:1} &\\
		\phi_{\mathcal{U}:2}(x) &= x &\qquad (\forall x \in K)\\
		\phi_{\mathcal{U}:1}\left(K\right) &=K_{\delta}'. &
		\end{aligned}
		\label{eq_fragmentation_property_redux}
		\end{equation}
		In particular,~\eqref{eq_fragmentation_property_redux} implies that for each $x \in int(K)$ we have that
		$
		\phi(x) = \phi_{\mathcal{U}:2}\circ \phi_{\mathcal{U}:1}(x) = \phi_{\mathcal{U}:1}(x)
		.
		$
		Therefore, for every $\hat{\phi} \in \Phi_{\star:d}$, the following holds
		\begin{equation}
		\sup_{x \in K_{\delta}'} \left\|
		\hat{\phi}(x)-\phi(x)
		\right\| 
		= 
		\sup_{x \in K_{\delta}'} \left\|
		\hat{\phi}(x)-\phi_1(x)
		\right\| 
		.
		\label{eq_estimate_redux}
		\end{equation}
		Since $K_{\delta}'$ is, by definition, dense in its closure $\overline{K_{\delta}'}$ and since continuous functions are determined by their value on dense sets then
		\begin{equation}
		\max_{x \in K} \left\|
		\hat{\phi}(x)-\phi(x)
		\right\| 
		\leq 
		\sup_{x \in K_{\delta}'} \left\|
		\hat{\phi}(x)-\phi(x)
		\right\| 
		= 
		\max_{x \in \overline{K_{\delta}'}} \left\|
		\hat{\phi}(x)-\phi_1(x)
		\right\| 
		.
		\label{eq_estimate_redux_improved}
		\end{equation}
		Therefore, we only need to approximate $\phi_1$ on $\overline{K_{\delta}'}$ in order to approximate $\phi$ on $K\subset \overline{K_{\delta}'}$.  Thus, without loss of generality, we may assume that $K=\overline{int(K)}$ and that $\phi:K\rightarrow K$.

		With these key simplifications, we proceed to the main part of the proof.  
		Since $\rrd$ is separable then so is $K$.  Therefore, there exists a sequence $\{x_n\}_{n \in \nn}$ contained in $K$ such that $\{x_n\}_{n\in\nn}$ is a dense subset of $K$.  Since the value of any continuous function is determined by its value on any dense subset, and since $\phi$ is continuous, then we only need to approximate $\phi$ on $\{x_n\}_{n \in \nn}$.  
		
		Define the sequence $\{z_n\}_{n \in\nn} \in \phi(K)$ by $z_n\triangleq \phi(x_n)$, for each $n \in \nn$.  For any $N\in \nn_+$, Theorem~\ref{thrm_memory_capacity_w_guessing} implies that there exists some $\hat{\phi} \in \Phi_{\star:d}$ satisfying $\hat{\phi}(x_n)=\phi(x_n)$ for every $n\leq N$.  Set $\hat{\phi}_0\triangleq \hat{\phi}$.  We recursively generate a sequence of reconfiguration networks in $\Phi_{\star:d}$ converging to $\phi$ on $\{x_n\}_{n \in \nn}$ and therefore on $K$.  
		
		By Lemma~\ref{lem_thrm_memory_capacity_w_guessing_two_point_version}, for every $n> N$ there exists some continuous curve with finite length $\gamma_{x_n:z_n}:[0,1]\rightarrow \rrd$ satisfying $\gamma_{x_n:z_n}(0)=x_n$, $\gamma_{x_n:z_n}(1)=z_n$, and for which there is some $\Delta_n>0$ satisfying
		$$
		\Delta_n\triangleq \min_{k<n,t\in [0,1]}\|\gamma_{[x_n:z_n]}(t)- z_k\|>0.
		$$
		.  
		For any decreasing of positive numbers $\{\epsilon_{n:k}\}_{k \in \nn}$ in $(0,\Delta_n)$
		Since $\gamma_{[x_n:z_n]}$ has finite length then one can choose an increasing sequence $\{t_{n:k}\}_{k \in \nn}$ in $[0,1]$ such that $t_{n:0}=0$ and 
		$
		\gamma_{[x_n:z_n]}([0,1])\subseteq \bigcup_{k\in \nn} \operatorname{Ball}\left(\gamma_{[x_n:z_n]}(t_{n:k}),\epsilon_{n:k}\right) 
		.
		$  
		Since $\gamma_{[x_n:z_n]}$ is continuous and $[0,1]$ is compact, then \citep[Theorem 26.5]{munkres2014topology} implies that $\gamma_{[x_n:z_n]}([0,1])$ is compact and therefore there must exist a finite sub-collection $\{(\epsilon_{n:k_j},t_{n:k_j})\}_{j=1}^{J_n}$ such that
		\begin{equation}
		\gamma_{[x_n:z_n]}([0,1])\subseteq \bigcup_{j=1}^{J_n} \operatorname{Ball}\left(\gamma_{[x_n:z_n]}(t_{n:k_j}),\epsilon_{n:k_j}\right) 
		\label{eq_another_convering_lemma}
		.
		\end{equation}
		By Lemma~\ref{lem_tech_lemmas} (ii), we choose 
		reconfigurations networks 
		$\hat{\phi}_{n:1},\dots,\hat{\phi}_{n:J_n-1}\in \Phi_{\star:d}$ so that 
		\begin{align}
		\hat{\phi}_{n:j}\left(\gamma_{[x_n:z_n]}(t_{n:k_j})\right)
		&= 
		\gamma_{[x_n:z_n]}(t_{n:k_{j+1}}),
		\label{eq_second_bound_transport}
		\\
		\sup_{u \in \rrd,\, \|u-\gamma_{[x_n:z_n]}(t_{n:k_j})\|\geq \epsilon_{n:k_j}} 
		\|\hat{\phi}_{n:j}(u)-u\| & = 0
		\label{eq_second_bound_itself}
		\end{align}
		for $k\in \{1,\dots,J_n-1\}$.  In particular, $\sup_{u \in \rrd}
		\|\hat{\phi}_{n:j}(u)-u\| \leq 2 \epsilon_{n:k_j}.
		$

		Let us describe how to choose $\{\epsilon_{n:k}\}_{k \in \nn}$.  We define this sequence recursively according to
		$$
		\epsilon_{n,k}\triangleq \min\left\{3^{-{N(n,k)}},
		3^{-{N(n,k)}}
		\alpha\left(\hat{\phi}_{{n:k-1}}\circ \dots
		\hat{\phi}_{{n:1}}\circ \dots \hat{\phi}_{{1:1}},\frac1{2^{N(n,k)}}\right)
		\right\}
		,
		$$ 
		where $N(n,k)$ is the number of elements in $\epsilon_{1,k_1},\dots,\epsilon_{1,k_{J_1}},\dots,\epsilon_{n,1},\dots,\epsilon_{n,k_{j-1}}$.  
		Note that, as discussed in \citep[page 367 preceding Lemma 2.1]{AndersonOnTopInfDeficiency}, the quantity 
		$$
		\alpha\left(\hat{\phi}_{{n:k-1}}\circ \dots
		\hat{\phi}_{{n:1}}\circ \dots \hat{\phi}_{{1:1}},\frac1{2^{{N(n,k)}}}\right)
		$$ 
		is positive by the compactness of $K$.  Hence, the sequence $\{\epsilon_{n,k}\}_{k \in \nn}$ is decreasing and non-negative.  
		Thus, by the argument preceding~\eqref{eq_second_bound_transport}, there exist well-defined sequence $\hat{\phi}_{1,1},\dots,\hat{\phi}_{1,J_1}, \dots,$ $\hat{\phi}_{n,1},\dots$, $\hat{\phi}_{n,J_n},\dots$ satisfying
		\begin{align}
		\hat{\phi}_{n,J_n}\circ \dots \hat{\phi}_{n,1}\circ\dots \circ \hat{\phi}_{1,J_1}\circ \dots \circ \hat{\phi}_{1,1}(x_n)
		= & \phi(x_n)
		\label{eq_approximation_criterion}
		\\
		\sup_{u \in \rrd} 
		\|\hat{\phi}_{{n:j}}(u)-u\| & < \epsilon_{n,k}
		.
		\label{eq_second_bound_itself_done}
		\end{align}
		Hence,~\eqref{eq_second_bound_itself_done} and the Inductive Convergence Criterion imply that the limit 
		$$
		\lim\limits_{n \uparrow \infty} 
		\hat{\phi}_{n,J_n}\circ \dots \circ 
		\hat{\phi}_{n,1}\circ \dots \circ
		\hat{\phi}_{1,J_1}\circ \dots \circ 
		\hat{\phi}_{1,1}
		$$ converges in $\hhh(\rrd)$ (with respect to the topology of uniform convergence on compacts).  Denote its limit by $\hat{\phi}_{\infty}$.  Next,~\eqref{eq_approximation_criterion} implies that $\hat{\phi}_{\infty}(x_n)=\phi(x_n)$ for every $n \in \nn$.  Moreover, \citep[Theorem 46.5]{munkres2014topology} guarantees that the uniform limit of continuous functions is again continuous, therefore $\hat{\phi}_{\infty}$ and $\phi$ are both continuous.  Since continuous functions agreeing on dense subsets are equal, then we conclude that $\hat{\phi}_{\infty}=\phi$.  In particular, for the given value of $\epsilon>0$, there must be some integer ${N(n,k)}\in \nn_+$ such that for every $N\in \nn$ with $N\geq N(n,k)$ we have that:
		$$
		\sup_{x \in K}
		\left\|
		(\hat{\phi}_{N,J_N}\circ \dots \circ
		\hat{\phi}_{N,1}\circ \dots \circ 
		\hat{\phi}_{1,J_1}\circ \dots \circ 
		\hat{\phi}_{1,1})(x)
		- \phi(x)
		\right\|
		<
		\epsilon
		.
		$$
		This concludes the proof.
	\end{proof}

	\subsection{{Proof of Corollary~\ref{cor_NEU_1}}}\label{ss_Appendix_proof_NEU_1}
	\begin{lemma}\label{lem_existence_of_optimizer_to_L}
		Let $\{x_n\}_{n=1}^N$ be points in $X$, where $X$ is a compact subset of $\rrd$, and suppose that $L$ and $P$ satisfy Assumption~\ref{ass_optimizability}.  Then: 
		\begin{enumerate}[(i)]
			\item The map $g\mapsto \sum_{n=1}^N L\left(
			f(x_n),g(x_n),x_n
			\right) + P(g)$ is continuous from $C(X,\rrD)$ to $\rr$,
			\item There exists some $f^{\star} \in C(\rrd,\rrD)$ satisfying
			\begin{equation}
			f^{\star}\in 
			\underset{g\in C(X,\rrD)}{\operatorname{argmin}}\,
			\sum_{n=1}^N L\left(
			f(x_n),g(x_n),x_n
			\right) + P(g)
			\label{eq_lem_existence_of_optimizer_to_L}
			.
			\end{equation}
		\end{enumerate}
	\end{lemma}
	\begin{proof}[{Proof of Lemma~\ref{lem_existence_of_optimizer_to_L}}]
		Since $\rrD$ and $\rr$ are metric spaces then \citep[Theorem 46.8]{munkres2014topology} implies that the topology of uniform convergence and the compact-open\footnote{
			See \citep[page 285]{munkres2014topology} for a definition of the compact-open topology; however, its definition is never explicitly needed in our argument; but rather only its properties.  
		} topologies coincide on $C(X,\rrD)$ and on $C(X,\rrd)$.  Since $L$ and $f$ are continuous and since $X$ is compact, then \citep[Theorem 46.10]{munkres2014topology} implies that, for every $n\in \{1,\dots,N\}$ the evaluation map:
		$$
		g\mapsto L(f(x_n),g(x_n),x_n) ,
		$$
		is continuous from $C(X,\rrD)$ to $C(X,\rr)$.  Since the sum of (finitely many) continuous functions is continuous then the map
		\begin{equation}
		g\mapsto \sum_{n=1}^N L(f(x_n),g(x_n),x_n) + P(g)
		\label{eq_key_quantity_proof_of_lem_existence_of_optimizer_to_L}
		,
		\end{equation}
		is continuous on $C(X,\rrD)$.  This gives (i).  
		
		Continuing our argument, for every $g \in C(\rrd,\rrD)$ the quantity $\sum_{n=1}^N L(f(x_n),g(x_n),x_n) + P(g)$ is finite since $L$ and $P$ were assumed to be finite-valued.  Since $L$ was assumed to be bounded-below, then in particular:
		$$
		\min_{n=1,\dots,N,g\in C(X,\rrD)} L(f(x_n),g(x_n),x_n)\geq \min_{y_1,y_2\in \rrD,x \in \rrd} L(f(x_n),g(x_n),x_n)>-\infty.
		$$
		Now since $P$ was also assumed to be bounded-below on $C(\rrd,\rrD)$ then it particular it is bounded-below on $C(X,\rrD)$; hence the functional of~\eqref{eq_key_quantity_proof_of_lem_existence_of_optimizer_to_L} is continuous, bounded-below, and finite-valued on $C(X,\rrD)$.  Therefore, \citep[Theorem 1.15]{DalMasoGamma1993} applies therefore there exists some $f^{\star} \in C(X,\rrD)$ satisfying~\eqref{eq_lem_existence_of_optimizer_to_L}.  This gives (ii).  
	\end{proof}

	\begin{proof}[{Proof of Corollary~\ref{cor_NEU_1}}]
		By Lemma~\ref{lem_existence_of_optimizer_to_L} (ii) there exists some $f^{\star}\in C(X,\rrD)$ satisfying~\eqref{eq_lem_existence_of_optimizer_to_L}.  By Theorem~\ref{sss_UAP_cnt}, for every full-rank matrix $A \in \operatorname{Mat}_{d,d+D}$ and every $\delta>0$, there exists a matrix $B^{\delta}\in \operatorname{Mat}_{d+D,D}$ and some $\hat{\phi}^{\delta}\in \Phi_{\star:d+D}$ such that
		\begin{equation}
		\sup_{x \in X} \left\|
		f^{\star}(x) - B^{\delta}\hat{\phi}^{\delta}((I_d\oplus A)x)
		\right\|<\delta
		.
		\label{eq_proof_cor_NEU_1_estimate_1}    
		\end{equation}
		By Lemma~\ref{lem_existence_of_optimizer_to_L} (i) the map $F:g\mapsto \sum_{n=1}^N L\left(  f(x_n),g(x_n),x_n\right) + P(g)$ is continuous and therefore~\eqref{eq_proof_cor_NEU_1_estimate_1} implies that for every $\epsilon>0$ we may pick some $\delta>0$ such that:
		$$
		F\left(
		B^{\delta}\hat{\phi}^{\delta}((I_d\oplus A)\cdot)
		\right)
		\leq
		F(f^{\star}) + \epsilon.
		$$
		Relabeling, yields the conclusion.  
	\end{proof}
	
	\subsection{{Proof of Corollary~\ref{cor_NEU_2}}}\label{ss_Appendix_proof_NEU_2}
	\begin{proof}[{Proof of Corollary~\ref{cor_NEU_2}}]
		The existence of $f^{\star}$ is guaranteed by Lemma~\ref{lem_existence_of_optimizer_to_L}.  Define the matrices $B\in \operatorname{Mat}_{D\times d+D}$ and $A\in \operatorname{Mat}_{d+D,d}$, respectively, by:
		\begin{align}
		A\triangleq (I_d|0_D) \mbox{ and }
		B\triangleq (0_d|I_D),
		\end{align}
		where $0_D$ and $0_d$ are respectively the respective zero matrices of $\operatorname{Mat}_{D\times D}$ and of $\operatorname{Mat}_{d\times d}$.  Set $\hat{f}(z)\triangleq Bz$ and observe that $\hat{f}\in \fff$ since $\fff$ contains all linear maps.  
		
		For every $n\leq N$ set
		$$
		\begin{aligned}
		z_n &\triangleq (0,\dots,0,f^{\star}(x_n)) \in \rrflex{d+D}\\
		x_n' & \triangleq (x_n,0,\dots,0) \in \rrflex{d+D}.
		\end{aligned}
		$$
		Since $f^{\star}(x_n)\neq f^{\star}(x_m)$ for every $n\neq m$, with $n,m\leq N$, then for any $n,m\leq N$ if $n\neq m$ then $z_n\neq z_m$ and $x_n'\neq x_m'$ (the latter observation follows from the fact that if $x_n'$ did equal to $x_m'$ then $x_n$ would equal to $x_m$ and then $f(x_n)$ would equal to $f(x_m)$ which is a contradiction of our assumption).  Since $x_n\neq 0$ for any $n\leq N$ then $x_n'\neq z_m$ for any $n,m\leq N$ since these must differ their first coordinate.  Hence, $\{z_n\}_{n=1}^N\cup\{x_n'\}_{n=1}^N$ is a set of $2N$ distinct points in $\rrflex{d+D}$ and $d\geq 1$, $D\geq 1$, thus $d+D\geq 2$.  Since $\Delta$ of Theorem~\ref{thrm_memory_capacity_w_guessing} is at-most equal to the right-hand side of~\eqref{eq_condition_delta_NEU} then Theorem~\ref{thrm_memory_capacity_w_guessing} applies; hence, there exists some $\hat{\phi}\in \Phi_{\star:d+D}$ satisfying
		\begin{equation}
		\hat{\phi}(x_n') = z_n 
		,
		\label{eq_proof_cor_NEU_2_computation_1}
		\end{equation}
		for every $n\leq N$.  
		In particular,~\eqref{eq_proof_cor_NEU_2_computation_1} and the definitions of $A$ and $B$ imply that for every $n\leq N$
		\begin{equation}
		\hat{f}\circ \hat{\phi}^{\delta}(Ax_n) = 
		B\hat{\phi}^{\delta}(Ax_n) = B\hat{\phi}^{\delta}(x_n') = Bz_n =f^{\star}(x_n)
		.
		\label{eq_proof_cor_NEU_2_computation_2}
		\end{equation}
		Thus, (i) holds.  
		
		Moreover, Theorem~\ref{thrm_memory_capacity_w_guessing} implies that $\hat{\phi}^{\delta}$ has depth at-most
		$\left\lceil
		\frac{N\pi}{2(\min\{2 \delta,1\})}
		\right\rceil$; depending on our choice of $\delta>0$; sufficiently small; and in particular, as specified by condition~\eqref{eq_condition_delta_NEU}.  This gives (iii).  
		
		By construction we have the containment:
		\begin{equation}
		\begin{aligned}
		&
		\left\{
		(x,\hat{f}(Ax))\in \rrflex{d+D}:\, (x,\hat{f}(Ax))\neq \hat{\phi}^{\delta}(x,\hat{f}(Ax))
		\right\}
		\\
		\subseteq &
		\left\{
		(x,y)\in \rrflex{d+D}:\, (x,y)\neq \hat{\phi}^{\delta}(x,y)
		\right\}
		\end{aligned}
		\label{eq_eq_proof_cor_NEU_2_computation_3},
		\end{equation}
		and the bound on the external covering number of the right-hand side of~\eqref{eq_eq_proof_cor_NEU_2_computation_3} is also guaranteed by Theorem~\ref{thrm_memory_capacity_w_guessing}.  This gives (ii) and completes the proof.
	\end{proof}

	\subsection{Proof of Theorem~\ref{thrm_Robustification}}\label{ss_AProof_robustification}
	\begin{proof}[{Proof of Theorem~\ref{thrm_Robustification}}]
		Observe that the term $\underset{x \in K}{\sup}
		\, 
		L(f(x_n),\hat{f}(x_n),x_n)$ does not depend on $\{w_n\}_{n=1}^N \in (0,1)^n$; hence:
		\begin{equation}
		\begin{aligned}
		&
		\underset{\underset{\{w_n\}_{n=1}^N w_n =1}{w_n\in (0,1)}}{\operatorname{argmin}}
		\underset{x \in K}{\sup}
		\, 
		L(f(x_n),\hat{f}(x_n),x_n)
		-
		\left[
		\sum_{n\leq N} w_n L(f(x_n),\hat{f}(x_n),x_n) 
		-
		\lambda \sum_{n\leq N} \frac1{N} \log\left(\frac{w_n}{N}
		\right)
		\right]
		\\	
		=& 
		\underset{\underset{\{w_n\}_{n=1}^N w_n =1}{w_n\in (0,1)}}{\operatorname{argmin}}
		-
		\left[
		\sum_{n\leq N} w_n L(f(x_n),\hat{f}(x_n),x_n) 
		-
		\lambda \sum_{n\leq N} \frac1{N} \log\left(\frac{w_n}{N}
		\right)
		\right]	
		.
		\end{aligned}
		\label{eq_proof_thrm_Robustification_part_1}
		\end{equation}
		Notice that the set $\left\{
		\{w_n\}_{n=1}^N:\, w_n \in (0,1),\, \sum_{n\leq N} w_n =1
		\right\}$ 
		describes the collection of probability measures $\qq$ which are equivalent to the uniform probability measure $\pp$ on $\{x_n\}_{n=1}^N$.  Thus, $\sum_{n\leq N} w_n L(f(x_n),\hat{f}(x_n),x_n)$ is the expectation of $L(f(x_n),\hat{f}(x_n),x_n)$ for a random variable $X$ taking values on $\{x_n\}_{n=1}^N$ and $\sum_{n\leq N} \frac1{N} \log\left(\frac{w_n}{N}\right)$ is the relative entropy of the measure $\qq$ with respect to 
		$\pp$, in the sense of \citep[Definition 3]{wang2020entropic}.  Therefore \citep[Proposition 1]{wang2020entropic} applies, thus, the measure $\pp^{\star}$ defined by
		$$
		\pp^{\star}(X=x_i)=
		\frac{e^{\lambda^{-1}L(f(x_n),\hat{f}(x_n),x_n)}}{\sum_{n\leq N} e^{\lambda^{-1}L(f(x_n),\hat{f}(x_n),x_n)}}
		$$
		optimizes the right-hand side of~\eqref{eq_proof_thrm_Robustification_part_1}. 
	\end{proof}
	\begin{proof}[{Proof of Corollary~\ref{cor_gain_quantification}}]
		Fix $\hat{f}\in C(\rrd,\rrD)$ and let $\{w_{n}^{\lambda,\hat{f}}\}_{n=1}^N$ be as in Theorem~\ref{thrm_Robustification} and use $\qq^{\{w_{n}^{\lambda,\hat{f}}\}_{n=1}^N}$ to denote the measure on $\{x_n\}_{n=1}^N$ taking values with probabilities $\{w_n^{\lambda,\hat{f}}\}_{n=1}^N$.  Then by Theorem~\ref{thrm_Robustification}
		\begin{equation}
		\begin{aligned}
		\Gen[\hat{f}][\{w_{n}^{\lambda,\hat{f}}\}_{n=1}^N] + \lambda
		\Dkl[\pp][
		\qq^{\{w_{n}^{\lambda,\hat{f}}\}_{n=1}^N}
		]
		\leq &
		\Gen[\hat{f}][\left\{\frac1{N}\right\}_{n=1}^N]  + \lambda \Dkl[\pp][
		\pp
		].
		\end{aligned}
		\label{eq_proof_cor_gain_quantification_1}
		\end{equation}
		By Gibbs' inequality, see \citep[Exercise 2.26]{LearningMacKay}, we have that $\Dkl[\pp][\qq^{\{w_{n}^{\lambda,\hat{f}}\}_{n=1}^N}]\geq 0$ and $\Dkl[\pp][\pp]=0$.  Thus, since $\lambda>0$ then~\eqref{eq_proof_cor_gain_quantification_1} reduces to
		\begin{equation}
		\begin{aligned}
		\Gen[\hat{f}][\{w_{n}^{\lambda,\hat{f}}\}_{n=1}^N] 
		\leq &
		\Gen[\hat{f}][\{w_{n}^{\lambda,\hat{f}}\}_{n=1}^N] + \lambda
		\Dkl[\pp][
		\qq^{\{w_{n}^{\lambda,\hat{f}}\}_{n=1}^N}
		]
		\\
		\leq &
		\Gen[\hat{f}][\left\{\frac1{N}\right\}_{n=1}^N]  + \lambda \Dkl[\pp][
		\pp
		]
		\\
		=&  \Gen[\hat{f}][\left\{\frac1{N}\right\}_{n=1}^N],
		\end{aligned}
		\label{eq_proof_cor_gain_quantification_2}
		\end{equation}
		with equality in the left-most inequality if and only if $\qq^{\{w_{n}^{\lambda,\hat{f}}\}_{n=1}^N} = \pp$.  
		Hence, from~\eqref{eq_proof_cor_gain_quantification_2} we have that
		\begin{equation}
		0\leq  
		\Gen[\hat{f}][\left\{\frac1{N}\right\}_{n=1}^N] 
		- 
		\Gen[\hat{f}][\{w_{n}^{\lambda,\hat{f}}\}_{n=1}^N] 
		\label{eq_proof_cor_gain_quantification_3}
		,
		\end{equation}
		with equality if and only if $\qq^{\{w_{n}^{\lambda,\hat{f}}\}_{n=1}^N}=\pp$.  
		Applying Theorem~\ref{thrm_Robustification} we have an explicit form for $\qq^{\{w_{n}^{\lambda,\hat{f}}\}_{n=1}^N}$ and therefore, we compute the right-hand side of~\eqref{eq_proof_cor_gain_quantification_3} as follows
		\begin{equation}
		\begin{aligned}
		0\leq  & 
		\Gen[\hat{f}][\left\{\frac1{N}\right\}_{n=1}^N] 
		- 
		\Gen[\hat{f}][\{w_{n}^{\lambda,\hat{f}}\}_{n=1}^N] \\
		= & \left[
		\sup_{x \in K} L(f(x_n),\hat{f}(x_n),x_n) - \frac1{N}\sum_{n\leq N}L(f(x_n),\hat{f}(x_n),x_n)
		\right]
		\\
		& -
		\left[
		\sup_{x \in K} L(f(x_n),\hat{f}(x_n),x_n) - \sum_{n\leq N} w_{n}^{\lambda,\hat{f}}L(f(x_n),\hat{f}(x_n),x_n)
		\right]
		\\
		= & 
		\sum_{n\leq N} w_{n}^{\lambda,\hat{f}}L(f(x_n),\hat{f}(x_n),x_n)
		-
		\frac1{N}\sum_{n\leq N}L(f(x_n),\hat{f}(x_n),x_n)
		\\ 
		= &
		\sum_{n\leq N} 
		\left(
		\frac{
			e^{-\lambda^{-1}L(\theta, x_n)}
		}{
			\sum_{n\leq N} e^{-\lambda^{-1}L(\theta, x_n)}
		}
		-
		\frac1{N}
		\right)
		L(f(x_n),\hat{f}(x_n),x_n)
		.
		\end{aligned}
		\label{eq_proof_cor_gain_quantification_4}
		\end{equation}
		Lastly, note that $\qq^{\{w_{n}^{\lambda,\hat{f}}\}_{n=1}^N}=\pp$ if and only if for each $1\leq n\leq N$ we have
		\begin{equation}
		\frac{
			e^{\lambda^{-1}L(f(x_n),\hat{f}(x_n),x_n)}
		}{
			\sum_{n\leq N} e^{\lambda^{-1}L(f(x_n),\hat{f}(x_n),x_n)}
		} = \frac1{N}
		\label{eq_proof_cor_gain_quantification_b1_equality_condition}
		.
		\end{equation}
		Thus,~\eqref{eq_proof_cor_gain_quantification_b1_equality_condition} holds if and only if for every $1\leq n,m\leq N$
		\begin{equation}
		\frac{
			e^{\lambda^{-1}L(f(x_n),\hat{f}(x_n),x_n)}
		}{
			\sum_{n\leq N} e^{\lambda^{-1}L(f(x_n),\hat{f}(x_n),x_n)}
		} = \frac1{N}
		= 
		\frac{
			e^{\lambda^{-1}L(\theta,x_m)}
		}{
			\sum_{n\leq N} e^{\lambda^{-1}L(f(x_n),\hat{f}(x_n),x_n)}
		}
		\label{eq_proof_cor_gain_quantification_b2_equality_condition}
		.
		\end{equation}
	\end{proof}
	\newpage

\end{document}